\documentclass{article}

\PassOptionsToPackage{numbers, compress}{natbib}
\usepackage[final]{neurips_2021}

\usepackage[utf8]{inputenc} % allow utf-8 input
\usepackage[T1]{fontenc}    % use 8-bit T1 fonts
\usepackage{hyperref}       % hyperlinks
\usepackage{url}            % simple URL typesetting
\usepackage{booktabs}       % professional-quality tables
\usepackage{amsfonts}       % blackboard math symbols
\usepackage{nicefrac}       % compact symbols for 1/2, etc.
\usepackage{microtype}      % microtypography
\usepackage{xcolor}         % colors
\usepackage{amsmath,amssymb,amsfonts}
\DeclareMathOperator*{\argmax}{argmax} % thin space, limits underneath in displays
\DeclareMathOperator{\EX}{\mathbb{E}}% expected value

\usepackage{algorithmic}
\usepackage{graphicx}
\usepackage{textcomp}
\usepackage{xcolor}
\usepackage{tabularx,booktabs}
\usepackage{tikz}
\usepackage[toc,page]{appendix}

\usepackage[utf8]{inputenc}
\usepackage[english]{babel}

\usepackage{wrapfig}

\hypersetup{
	colorlinks = true,
    urlcolor  = cyan,
	citecolor = blue,
	linkcolor = blue}

\usetikzlibrary{positioning}
\usetikzlibrary{calc}

\title{A Robust Unsupervised Ensemble of Feature-Based Explanations using Restricted Boltzmann Machines}

\newcommand*\samethanks[1][\value{footnote}]{\footnotemark[#1]}

\author{%
  Vadim Borisov\thanks{Corresponding author: \texttt{vadim.borisov@uni-tuebingen.de}} \\
  University of Tübingen\\
  \And
  Johannes Meier\thanks{Equal contribution}\\
  University of Tübingen\\
  \And
  Johan van den Heuvel\samethanks[2]{}\\
  University of Tübingen\\
  \And
  Hamed Jalali\\
  University of Tübingen\\
  \And
  Gjergji Kasneci\\
  University of Tübingen\\
}
\begin{document}

\maketitle 

\begin{abstract}
Understanding the results of deep neural networks is an essential step towards wider acceptance of deep learning algorithms. Many approaches address the issue of interpreting artificial neural networks, but often provide divergent explanations. Moreover, different hyperparameters of an explanatory method can lead to conflicting interpretations. 
In this paper, we propose a technique for aggregating the feature attributions of different explanatory algorithms using Restricted Boltzmann Machines (RBMs) to achieve a more reliable and robust interpretation of deep neural networks. 
Several challenging experiments on real-world datasets show that the proposed RBM method outperforms popular feature attribution methods and basic ensemble techniques. 
\end{abstract}

\section{Introduction}
\label{sec:intro}

As the applications of deep neural networks (DNNs) continue to grow, the black-box nature of DNNs creates potential trust issues~\cite{zhang2020survey}. Moreover, numerous life-critical (such as medical, automotive, or financial) applications utilize DNNs for various estimation tasks. In such applications, and especially for the long-term acceptance of artificial intelligence (AI) solutions, a deeper understanding and trust in the produced results is crucial. Furthermore, feature attribution methods are important tools for deep model debugging and diagnosis \cite{setzu2021glocalx}. 

Explaining how the input influences the output for a given DNN is one form to interpret the black-box nature of the DNN and bring trust to a system. These so-called feature-based explanation methods received a lot of attention in recent years~\cite{zhang2020survey, shrikumar2017learning, gilpin2018explaining,  adebayo2018sanity}. They can be grouped into three broad categories, (1) approaches based on gradient information~\cite{sundararajan2017axiomatic, kasneci2016licon}, (2) perturbation-based approaches~\cite{ribeiro2016should, lundberg2017unified}, and (3) attribution-based approaches~\cite{shrikumar2017learning, montavon2019layer}. Interestingly, different feature-based explanation approaches regularly produce mixed views on the main attributes (areas of an image or variables), and in the absence of the ground truth, it is still a challenge to verify which explanation method is the most trustworthy. Moreover, in the AI community, there are no yet accepted quality measures for feature-based explanations. 
\textit{All these difficulties resulted in a large number of different explanation methods and in a lack of consensus on which techniques are most reliable.}

Within the machine learning (ML) community, there is much work on the combination of methods that do not always agree with each other, i.e. \textit{ensemble learning}~\cite{zhang2014learning, zhou2019ensemble}. Normally ensemble models outperform the non-ensemble models and turn out to be more robust to outliers. The main idea is that if multiple methods make mistakes in different areas, combining them in an intelligent way improves performance and reduces the effect of outliers as compared to the single method.  Moreover, from statistical learning theory and practical applications, it is well understood that ensemble learning is the path of choice towards a more robust machine learning system~\cite{kuncheva2003measures}, even in unsupervised learning scenarios where the target is not available~\cite{jaffe2016unsupervised, shaham2016deep}. 

In this work, utilizing ideas from~\cite{shaham2016deep, zhang2014learning} and \cite{kasneci2011cobayes,kasneci2010bayesian}, we introduce a novel approach for the unsupervised ensemble learning of reliable and robust feature-based explanations for deep neural networks. To this end, we propose using a model based on Restricted Boltzmann Machines (RBMs), which achieves this goal by aggregating the results (saliency maps) of different feature-based explanation methods in a principled probabilistic fashion. Also, it has been shown that an RBM can be used in the truth discovery setting~\cite{broelemann2017restricted, DBLP:journals/corr/abs-1809-09703}, which is analogous to our task of finding a reliable feature importance map from different importance maps.

The main contributions of this work are: 
\begin{itemize}
    \item We introduce a novel method for a robust and reliable feature-based explanation using ensemble learning. 
    \item We empirically and visually show the superior performance of the proposed method in comparison to state-of-the-art feature attribution baselines.
    \item We open-source our code and make it publicly available, as an RBM ensemble framework: (\href{https://github.com/JohanvandenHeuvel/AggregationOfLocalExplanations}{https://github.com/JohanvandenHeuvel/AggregationOfLocalExplanations}), Besides, we also developed a single Python package with various evaluation metrics for feature attribution methods metrics: \href{https://github.com/meier-johannes94/ExplainableAIImageMeasures}{https://github.com/meier-johannes94/ExplainableAIImageMeasures}
\end{itemize}

The paper is organized as follows: In Section~\ref{sec:related_work} we discuss the related work and provide essential background information. Section~\ref{sec:rbm} presents the proposed ensemble method for local feature-based explanations using an RBM. In Section~\ref{sec:experiments}, we present the results of various experiments. Section~\ref{sec:discussion} discusses limitations of our work and ways to address them in the future. Section~\ref{sec:conclusion} concludes our work with a short summary.

%%%%%%%%%%%%%%%%%%%%%%%%%%%%%%%%%%%%%
%%%%%%%%%%%%%%%%%%%%%%%%%%%%%%%%%%%%%
%%%%%%%%%%%%%%%%%%%%%%%%%%%%%%%%%%%%%
%%%%%%%%%%%%%%%%%%%%%%%%%%%%%%%%%%%%%
%%%%%%%%%%%%%%%%%%%%%%%%%%%%%%%%%%%%%
%%%%%%%%%%%%%%%%%%%%%%%%%%%%%%%%%%%%%
%%%%%%%%%%%%%%%%%%%%%%%%%%%%%%%%%%%%%
%%%%%%%%%%%%%%%%%%%%%%%%%%%%%%%%%%%%%
%%%%%%%%%%%%%%%%%%%%%%%%%%%%%%%%%%%%%

\section{Background and Related Work}
\label{sec:related_work}

This part of the manuscript provides the needed background and discusses related approaches. First, we present the basic notation used in this work and proceed by presenting two ensemble techniques for aggregating feature importance maps. 

\subsection{Feature Attribution Function}

Formally, a feature attribution function can be seen as $\phi(f, \textbf{x}, c_x)$, where $f$ is a black box model and $\textbf{x}$ is an input data point from a corresponding class $c_x$. The output of $\phi$ is an explanation vector or matrix $\textbf{e}_{f(\textbf{x})}$, where each element of $\textbf{e}_{f(\textbf{x})}$ is an importance score for the corresponding feature value in $\textbf{x}$. A large positive or negative value in $\textbf{e}_{f(\textbf{x})}$ indicates that the corresponding feature (pixel) has a large influence on the outcome of the black-box model $f$.

\textbf{Assumption 2.1} In the following, we assume that a \underline{true} feature attribution $\bar{\textbf{e}}_f(\textbf{x})$ for a given model $f$ and input $\textbf{x}$ exists and can be constructed by adequately aggregating available attributions $\textbf{e}_{f(\textbf{x}), i}, i\in\{1,...,N\}$, where $N$ is the number of baseline explanations (from $N$ baseline methods).

For better readability and simplicity, from here we omit the index $f(\textbf{x})$.

The goal of any explanation method $\phi$ is to obtain an attribution $\textbf{e}$ that is as close as possible to $\bar{\textbf{e}}$. Note that our method naturally generalizes to probabilistic local explanation methods \cite{zhao2021baylime}.  Given the before-mentioned assumption, we can say that there is a joint probability distribution of the pair $(\textbf{e}, \bar{\textbf{e}})$ parametrized by $\theta$. 

\begin{equation*}
    p_\theta(\textbf{e}, \bar{\textbf{e}}) = p_\theta(\bar{\textbf{e}}) p_\theta(\textbf{e} | \bar{\textbf{e}}).
\end{equation*}

The joint distribution $p_\theta(\textbf{e}, \bar{\textbf{e}})$ is not known, and neither are the marginals $p_\theta(\textbf{e})$, $p_\theta(\bar{\textbf{e}})$.

For the following theoretical results we require that the explanation methods give independent explanations when conditioned on the true explanation. However, as with Naive Bayes methods, for practical purposes, this assumption can be violated without negatively impacting the aggregation quality \cite{shaham2016deep}. Also note that we do assume some consistency between the explanations, following the assumption that feature attributions reflect the underlying (but unknown) importance distributions of the feature values~\cite{sturmfels2020visualizing}.

Assuming conditional independence between the provided baseline explanations given the (unknown) true explanation, we have 

\begin{equation*}
    p_\theta(\textbf{e} | \bar{\textbf{e}}) = \prod_{n=1}^N p_\theta(\textbf{e}_n | \bar{\textbf{e}}),
\end{equation*}
where $\textbf{e}_n$ is a baseline explanation in the ensemble involving $N$ different baseline explanations.

\subsection{Ensemble Learning}

As we state in the introduction, ensemble learning is a well-studied approach for improving the performance of an ML system. One of the most basic ensemble methods employs the mean of results of base learners \cite{kuncheva2003measures}, where a \textit{base learner} is a single algorithm from the ensemble. 

\begin{equation}
    \textbf{e}_{mean} = \frac{1}{N} \sum_{n=1}^N\textbf{e}_{n}
    \label{eq:mean}.
\end{equation}

A significant drawback of the \emph{mean ensemble approach} is that it still is sensitive to outliers or noisy estimations. Furthermore, data scaling may strongly influence the aggregation. In Section \ref{sec:experiments}, these weaknesses of the \emph{mean ensemble approach} are also seen in the experimental evaluation.

To mitigate these weaknesses, the authors of \cite{rieger2020aggregating} propose to take the local uncertainty into account. To this end, they divide the mean by the local variance plus a constant $\epsilon$ for stability reasons, which results in the \emph{variance ensemble approach}:

\begin{equation*}
    \textbf{e}_{var} = \frac{1}{N} \sum_{n=1}^N \frac{\textbf{e}_{n}}{\sigma_*(\textbf{e}_{i \in \{1,...,N\}}) + \epsilon},
    \label{eq:var}
\end{equation*}

where $\sigma_*(\textbf{e}_{i \in \{1,...,N\}})$ is the point-wise standard deviation over all the available explanations $\textbf{e}_{i}, i\in\{1,...,N\}$. This method assigns less relevance to explanations that have high disagreement with the remaining explanations.

Also, the authors of \cite{bhatt2020evaluating} proposed a novel method to aggregate Shapley values through an explanation function that minimizes sensitivity. 

\section{Ensemble Learning using Restricted Boltzmann Machines}
\label{sec:rbm}

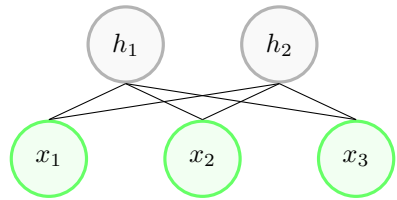
\begin{wrapfigure}{r}{5.5cm}
% \centering
\begin{tikzpicture}[
roundnode/.style={circle, draw=green!60, fill=green!5, very thick, minimum size=10mm},
roundnode2/.style={circle, draw=gray!60, fill=gray!5, very thick, minimum size=10mm}
]

%Nodes
\node[roundnode]    (x1)                                    {$x_1$};
\node[roundnode]    (x2)        [right=of x1]               {$x_2$};
\node[roundnode]    (x3)        [right=of x2]               {$x_3$};
\node[roundnode2]    (h1)        [above=of $(x1)!0.5!(x2)$]  {$h_1$};
\node[roundnode2]    (h2)        [above=of $(x2)!0.5!(x3)$]  {$h_2$};

%Lines
\draw[-] (h1.south) -- (x1.north);
\draw[-] (h1.south) -- (x2.north);
\draw[-] (h1.south) -- (x3.north);
\draw[-] (h2.south) -- (x1.north);
\draw[-] (h2.south) -- (x2.north);
\draw[-] (h2.south) -- (x3.north);

\end{tikzpicture}
\caption{An RBM with three visible and two hidden units. In our work, we use an RBM with a single hidden node.}
\label{fig:RBM} 
\vspace{-1.0cm}
\end{wrapfigure}

In this section, we present an unsupervised aggregation of feature attribution maps using a Restricted Boltzmann Machine (RBM). Similar aggregation techniques have been proposed in other contexts, e.g., in~\cite{shaham2016deep, broelemann2017restricted}.

\subsection{The Restricted Boltzmann Machine}

An RBM is an undirected bipartite graph that can be parametrized by a neural network. It is a variant of the Boltzmann Machine, with the additional property that there are no connections within both the group of visible nodes or the group of hidden nodes. The advantage of this property is that nodes in one group are conditionally independent of each other, given that we know the state of the nodes in the other group. One of the main characteristics of an RBM is that it can learn a probability distribution over its set of inputs. A graphical representation of an example RBM is shown in Figure \ref{fig:RBM}.

The formal definition of an RBM is as follows. There is a set $X$ of $n$ visible binary random variables and a set $H$ of $m$ hidden binary random variables. The RBM has parameters $\lambda = (\textbf{W}, \textbf{a}, \textbf{b})$. $\textbf{W}$ is the weight matrix of the connections between the nodes, $\textbf{a}$ is the bias of the visible layer and $\textbf{b}$ is the bias of the hidden layer.
Each possible state of the RBM, i.e. the particular values of $(X, H)$, is associated with the following energy function (in matrix notation):

\begin{equation*}
    E_\lambda(\textbf{x}, \textbf{h}) = -(\textbf{a}^T\textbf{x} + \textbf{b}^T\textbf{h} + \textbf{x}^T\textbf{W}\textbf{h}),
\end{equation*}

which then can also be used to define the joint probability distribution for the visible and hidden vectors is defined in terms of the energy function:

\begin{equation*}
    P_\lambda(\textbf{x}, \textbf{h}) = \frac{1}{Z} e^{-E(\textbf{x}, \textbf{h})},
\end{equation*}

where Z is the sum over $e^{-E(\textbf{x}, \textbf{h})}$ for all possible configurations $\textbf{x}, \textbf{h}$, which can be seen as a normalization constant to ensure that all probabilities sum to $1$, also known as the partition function.

The optimization objective of the RBM is to maximize the expected log probability of a training sample $\textbf{x}$:
\begin{equation}
\begin{split}
    &\argmax_\lambda \EX [\log P_\lambda(X = \textbf{x})] = \\ 
    &\argmax_\lambda \EX [\log \sum_\textbf{h} P_\lambda(X = \textbf{x}, H = \textbf{h})].
\end{split}
\end{equation}

To train an RBM, a gradient-based optimization can be applied using the contrastive divergence algorithm  \cite{hinton2006fast, bengio2009learning}.

\subsection{Aggregation of Local Explanations using an RBM}

Given an RBM with $N$ visible nodes and one hidden node, with input $\textbf{x}$, where $N$ is the number of baseline explanations in our ensemble, it can be shown that the true posterior probability of $y$ can be efficiently estimated (Lemma 4.1, Lemma 4.2 from \cite{shaham2016deep}). Furthermore, given the previously discussed mild assumptions on the input data (which are in line with those in \cite{shaham2016deep}), the maximum likelihood estimate $\bar{\lambda}_{MLE}$ for the parameters of the RBM, the RBM posterior probability $P_{\bar{\lambda}_{MLE}}(H = 1 | X = \textbf{x})$ converges to true posterior $P_{\theta}(Y = 1 | X = \textbf{x})$.

\begin{figure}
    \centering
    \includegraphics[width=13cm]{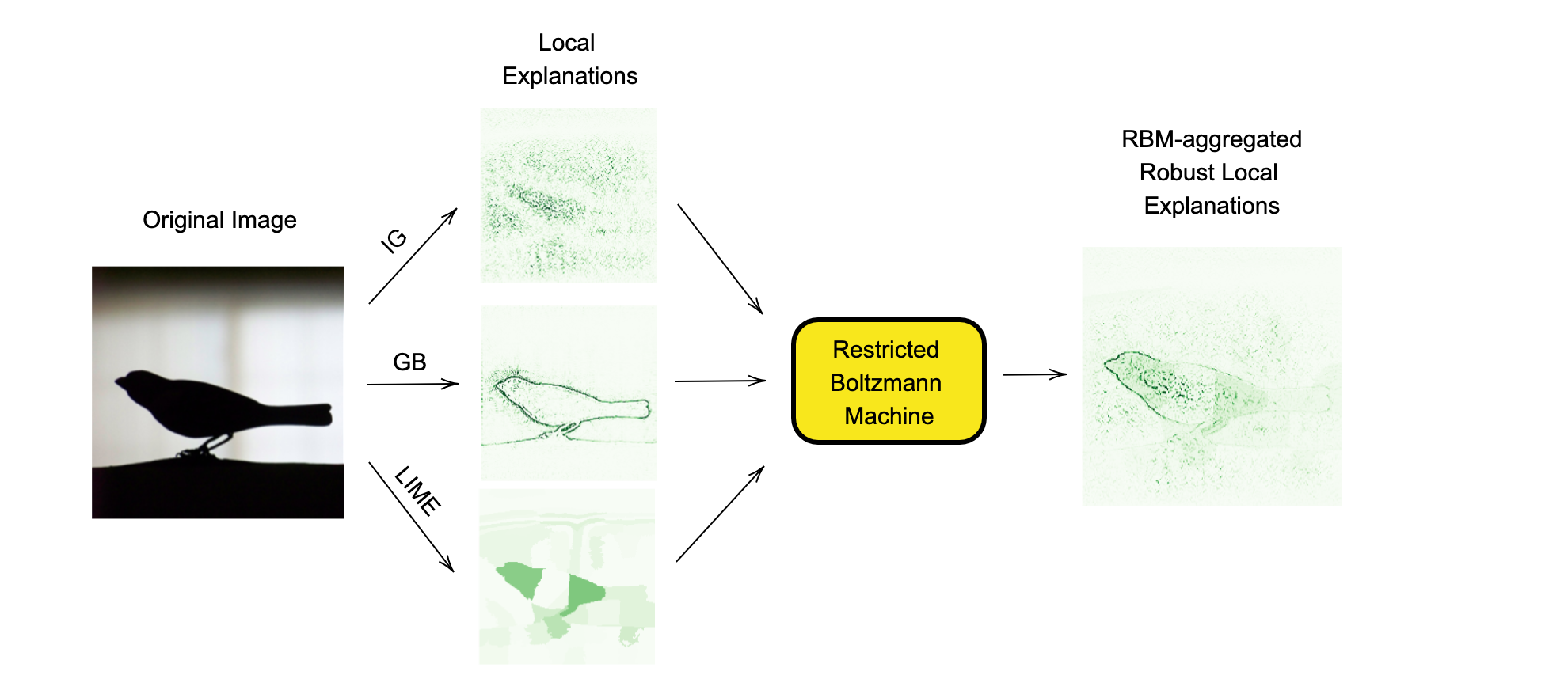}
    \caption{An overview on the ensemble of feature attribution maps from three different local explanation algorithms using an RBM for an image from the ImageNet dataset \cite{deng2009imagenet}.}
    \label{fig:overview}
    \vspace{-0.3cm}
\end{figure}

Hence, we are able to apply the RBM to the \textit{unsupervised aggregation} of $N$ available feature-based explanations. We assume a joint distribution $p_\theta( \textbf{e}, \bar{\textbf{e}})$, and that the $\textbf{e}_{i}$'s are conditionally independent from each other given $\bar{\textbf{e}}$. By fitting the RBM we learn the parameters $\theta$ and thus obtain the relationship between our known explanations $\textbf{e}_i$ and the true explanation $\bar{\textbf{e}}$. The ensemble pipeline of the proposed method is depicted in Fig. \ref{fig:overview}. 

In order to preserve the spatial information for visual data using the RBM-based ensemble, we do a pixel-wise aggregation. Therefore, for each pixel we train a Bernoulli RBM with a single hidden unit. 

A known limitation of an RBM is the so-called \textit{flipping issue} \cite{shaham2016deep, broelemann2017restricted, DBLP:journals/corr/abs-1809-09703}, which arises from the RBM parametrization symmetry. That is, the weights of the RBM can be flipped symmetrically without changing the behavior of the RBM. In order to avoid this unwanted effect, we propose two approaches: flip detection and metric optimization. 
The \textit{flip detection} algorithm extends the idea from Remark 4.3 in \cite{shaham2016deep}, by comparing the top 5 \% of most important and 5 \% of less important pixels to the mean baseline. The algorithm inverts the current important scores if there is a strong disagreement between the proposed approach and the mean baseline. The \textit{metric optimization} method utilizes the chosen metric to overcome the flipping issue. It compares two versions of the RBM ensemble results and selects the one with a better performance according to the selected metric. 

% In the rest of this report we denote $E_{mean}$ (eq. \ref{eq:mean}) and $E_{var}$ (eq. \ref{eq:var}) by MEAN-ENS and VAR-ENS respectively, and the RBM ensemble method by RBM-ENS.

% \textcolor{red}{Question: As you can see in \cite{zhang2014learning}, the main target in the ensemble learning is to aggregate some local classifiers $C_1,\ldots,C_J$ to obtain a final classifier $C^*$. I did not found a clear description in both the reference and this document about the $C^*$. How a RBM can return $C^*$. The observed nodes are the local predictions of J base classifiers but how hidden nodes can explain the ensemble? I think here you need to add a little information about this issue instead of only cite the related works in Line 135. It should be clear how you obtain the $C^*$ (or E in your definition in section 2). The dimension of X and h in the RBM is not known. If h layer show the aggregation, you only have one latent node. But if this layer shows the final predictions, it should have same number of nodes as X-layer.  }

% \textcolor{blue}{Thank you, for pointing this out. I agree, I will extend the subsection. Also, the base work for this paper is \cite{shaham2016deep}, since we do not have access to the true labels, we have to do find them in an unsupervised way.}

\section{Experiments}
\label{sec:experiments}

To demonstrate the effectiveness of the proposed ensemble algorithm we conduct various visual and quantitative experiments.  First, we present the visual inspection results on the MNIST \cite{lecun1998mnist} and ImageNet \cite{deng2009imagenet} datasets in two settings, with and without noisy explanation maps in our ensemble.

Despite a growing body of research focusing on explainable ML, the fair quantitative comparison of local explanation (or saliency-based) algorithms is still an open question, since the existing methods mostly utilize the pixel perturbation strategy (e.g., removing the most or least important pixels and reporting the change in recognition quality) \cite{ancona2018better, Hooker2018}. Also, such evaluations have a significant drawback, 
replacing image pixels with black or "mean" or any other pixel values may lead to artifacts affecting the data distribution~\cite{Hooker2018,DBLP:journals/corr/abs-2101-00905}. 
Nevertheless, since pixel perturbation analyses are employed in many related works, for our quantitative analysis we select the following approaches:  the pixel perturbation for insertion (IAUC) and deletion (DAUC). Furthermore, we utilize the iterative removal of features (IROF) analysis \cite{rieger2020irof}.  We explain each evaluation method in detail in the corresponding subsections. 

In our last experiment, we demonstrate that our ensemble approach can be also used within a singe feature attribution framework to achieve more robust and stable explanations. Since, it has been shown that hyperparameters choice can significantly affect the saliency maps \cite{bansal2020sam}.

\begin{table*}
%  \caption{Visual results of MNIST and ImageNet(VGG-19) in the noise case}
% \label{vis_results}
\begin{tabularx}{\textwidth}{l @{\extracolsep{\fill}} c @{\extracolsep{\fill}} c@{\extracolsep{\fill}}c@{\extracolsep{\fill}} c@{\extracolsep{\fill}}c@{\extracolsep{\fill}}c@{\extracolsep{\fill}} c@{\extracolsep{\fill}}c@{\extracolsep{\fill}}}
\toprule
$\,\,$ Original & LIME \cite{ribeiro2016should} & GB \cite{springenberg2015striving} & IG \cite{SundararajanTY17} & GS  \cite{lundberg2017unified} & SG \cite{smilkov2017smoothgrad} & Mean & Variance & \textbf{RBM} \\
 & & & & & & ensemble & ensemble &  \textbf{ensemble} \\
\midrule

\multicolumn{9}{c}{\textbf{Without} noisy feature attribution maps in the ensemble} \\

\includegraphics[width=14.5mm]{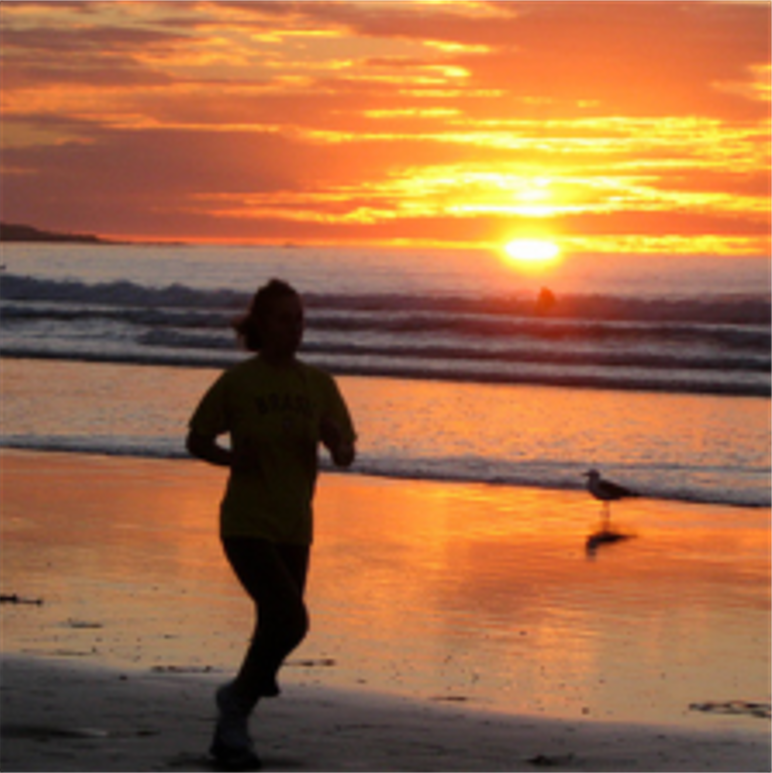} & 
\includegraphics[width=14.5mm]{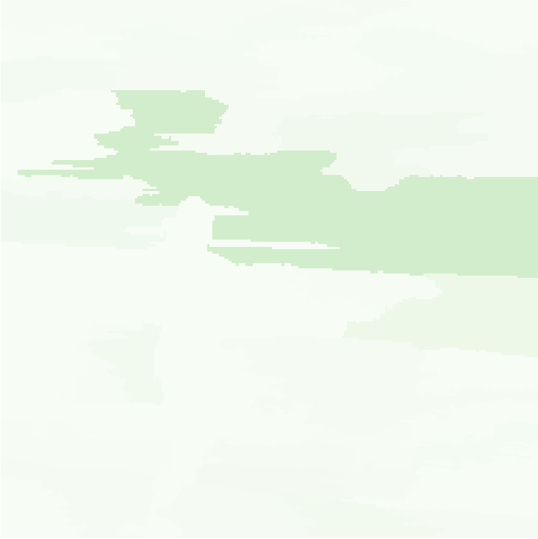} & 
\includegraphics[width=14.5mm]{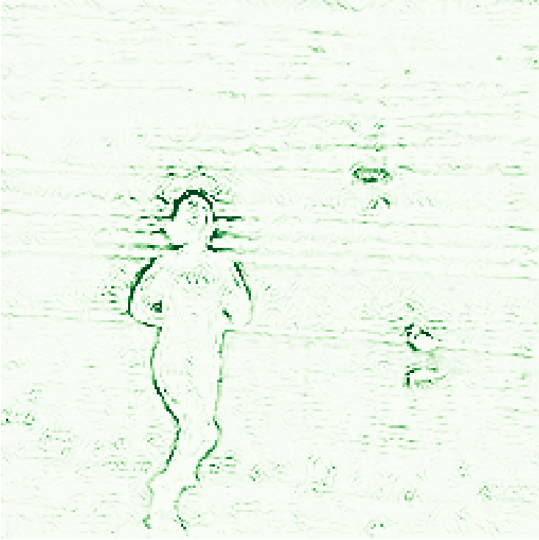} & 
\includegraphics[width=14.5mm]{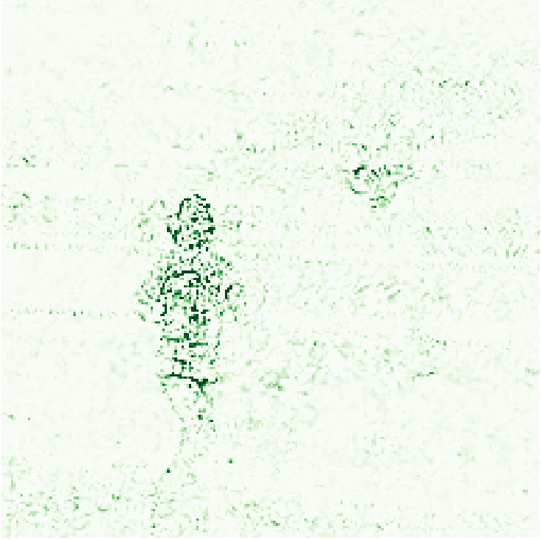} & 
\includegraphics[width=14.5mm]{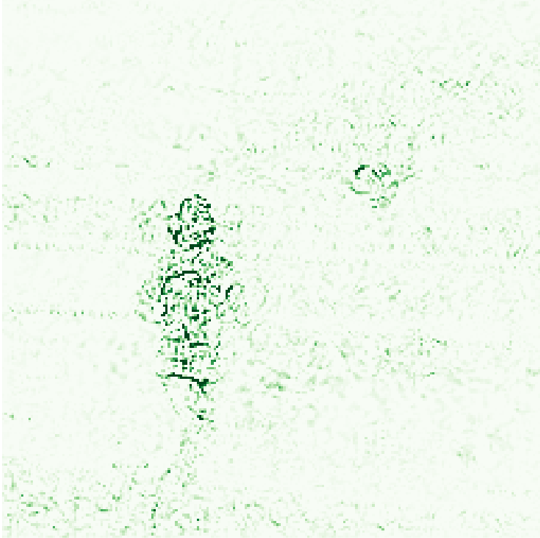} &
\includegraphics[width=14.5mm]{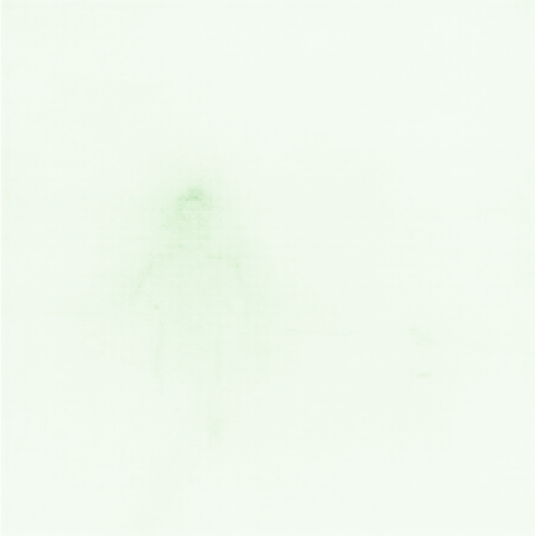} & 
\includegraphics[width=14.5mm]{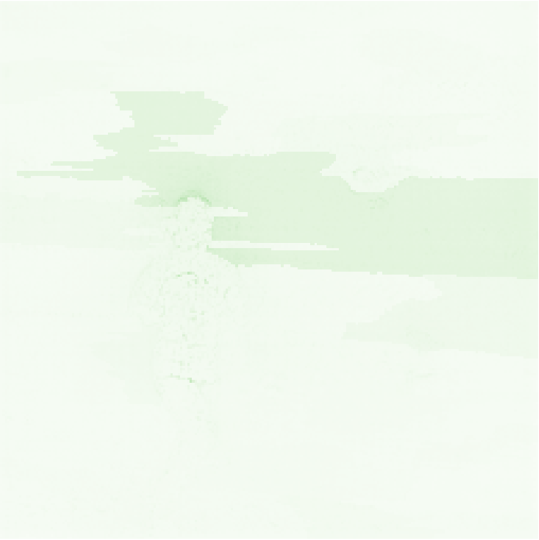} & 
\includegraphics[width=14.5mm]{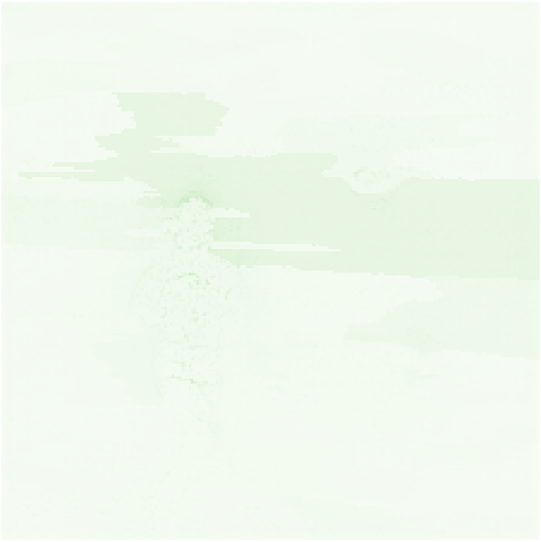} & 
\includegraphics[width=14.5mm]{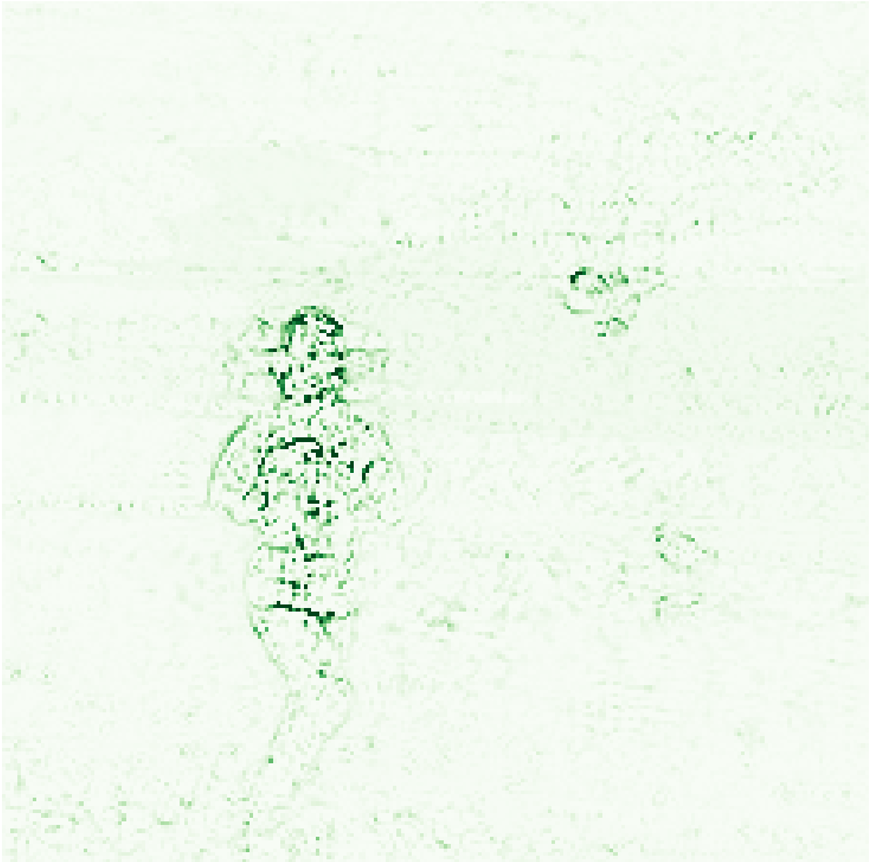} \\

\includegraphics[width=14.5mm]{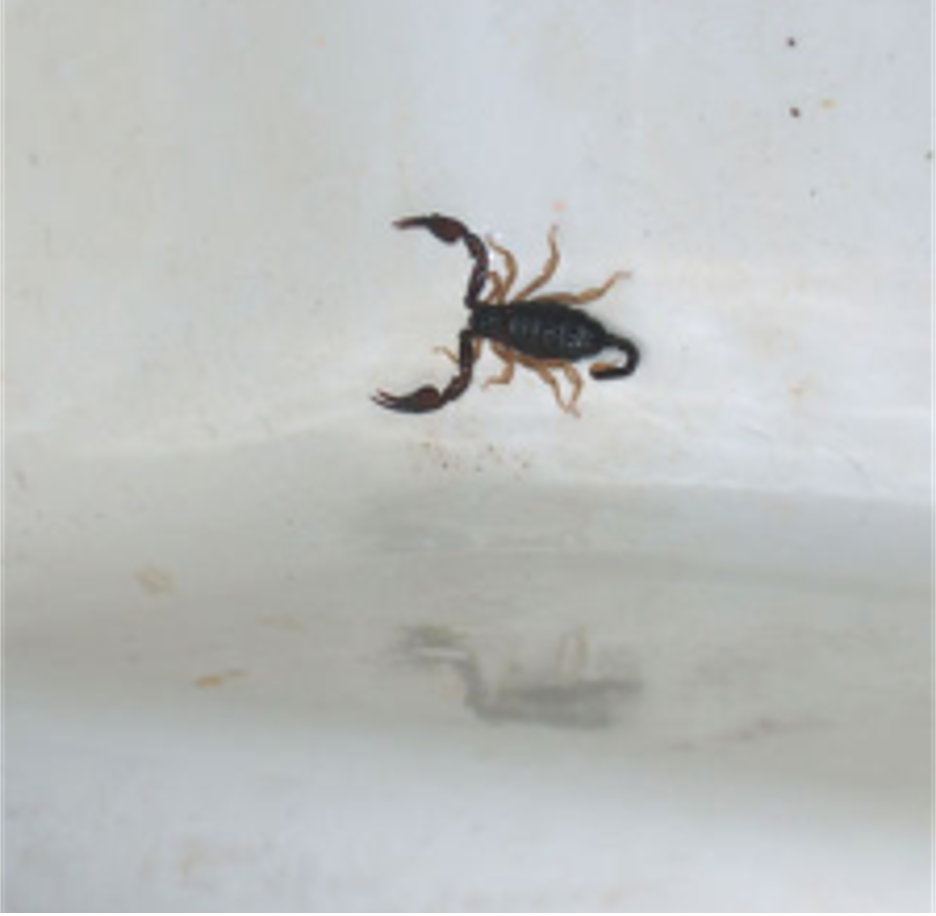} & 
\includegraphics[width=14.5mm]{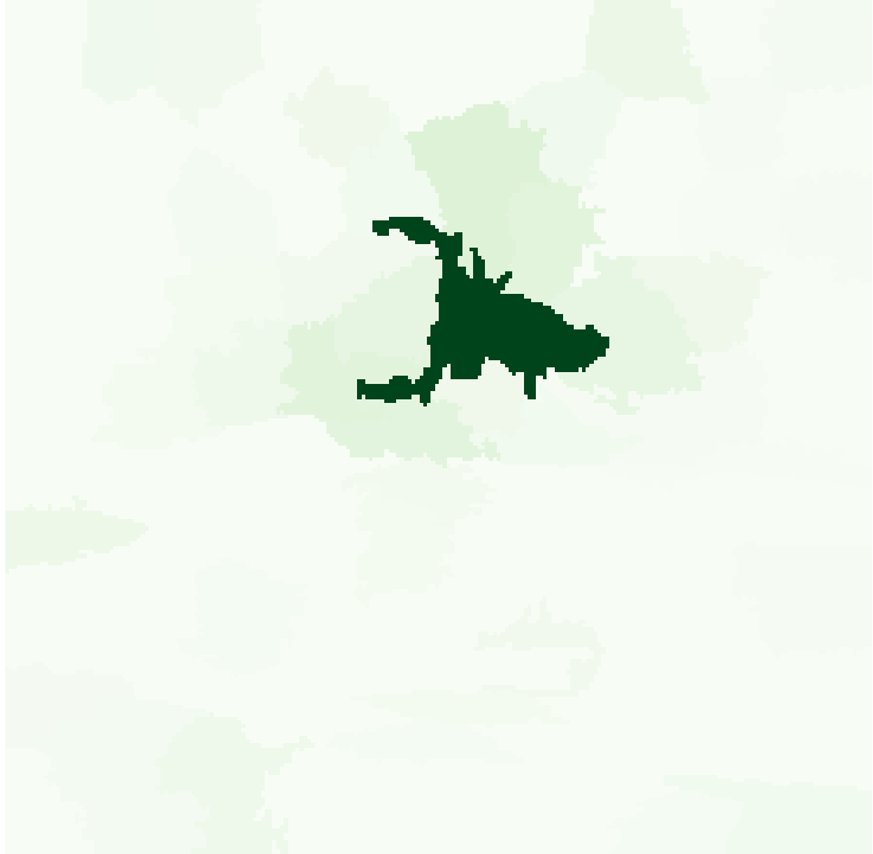} & 
\includegraphics[width=14.5mm]{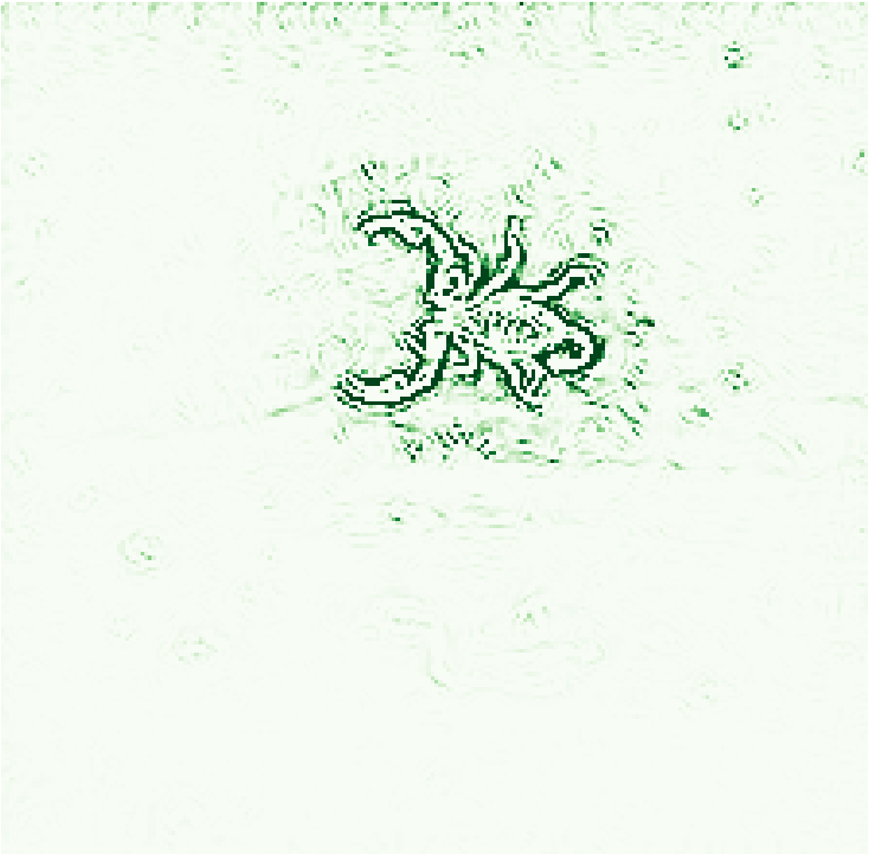} & 
\includegraphics[width=14.5mm]{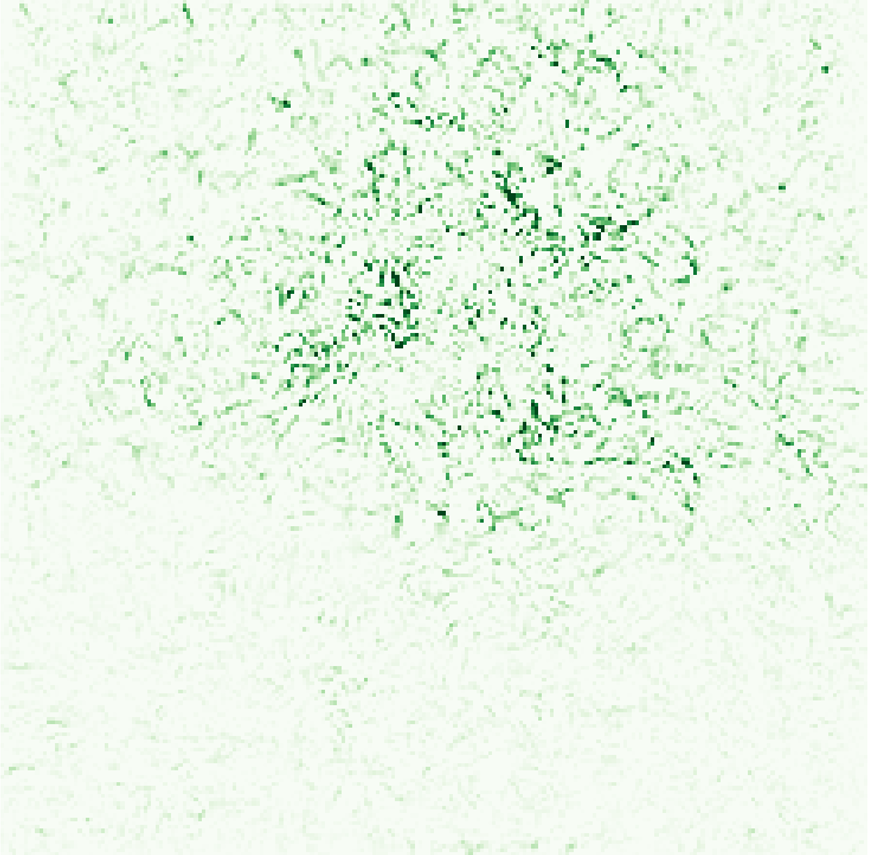} & 
\includegraphics[width=14.5mm]{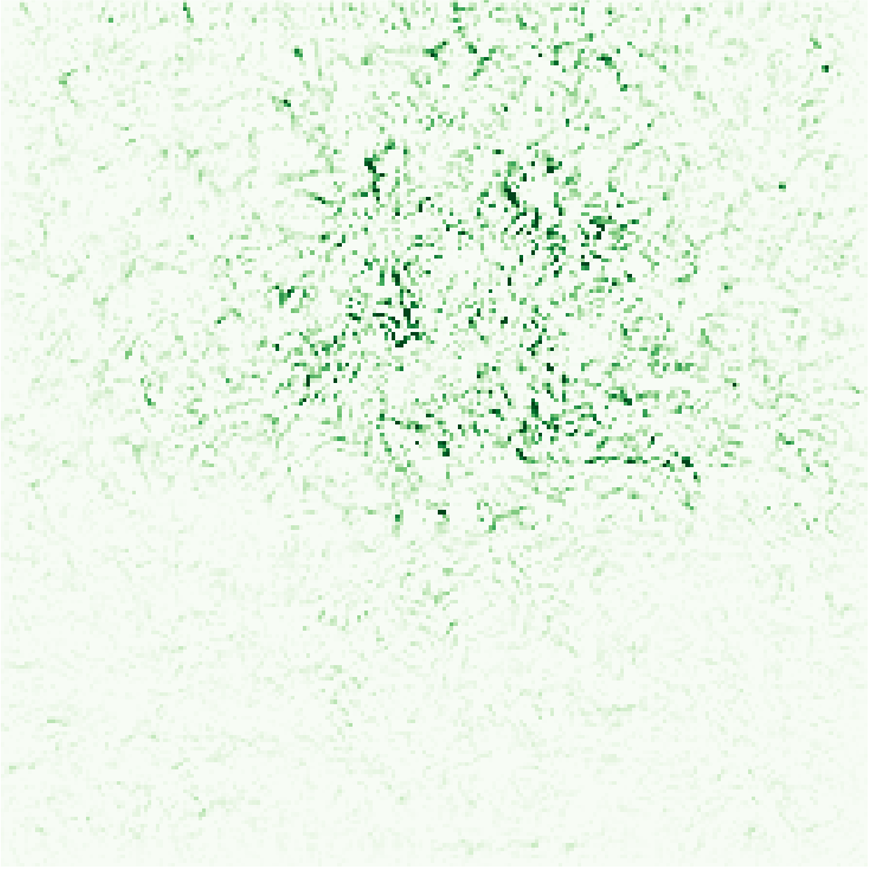} & 
\includegraphics[width=14.5mm]{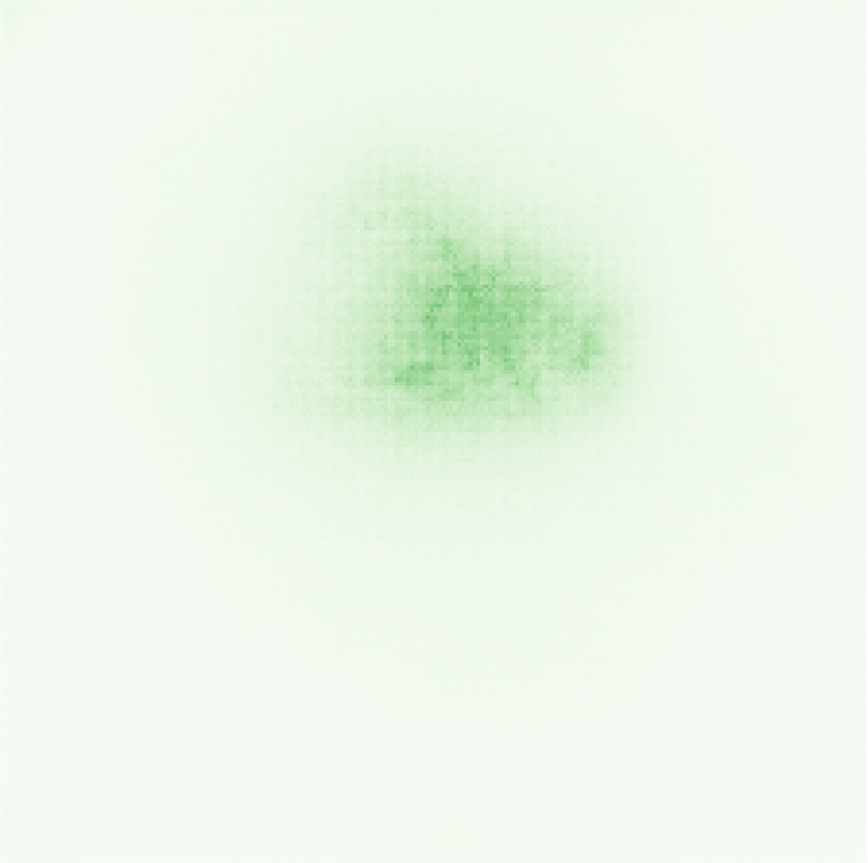} & 
\includegraphics[width=14.5mm]{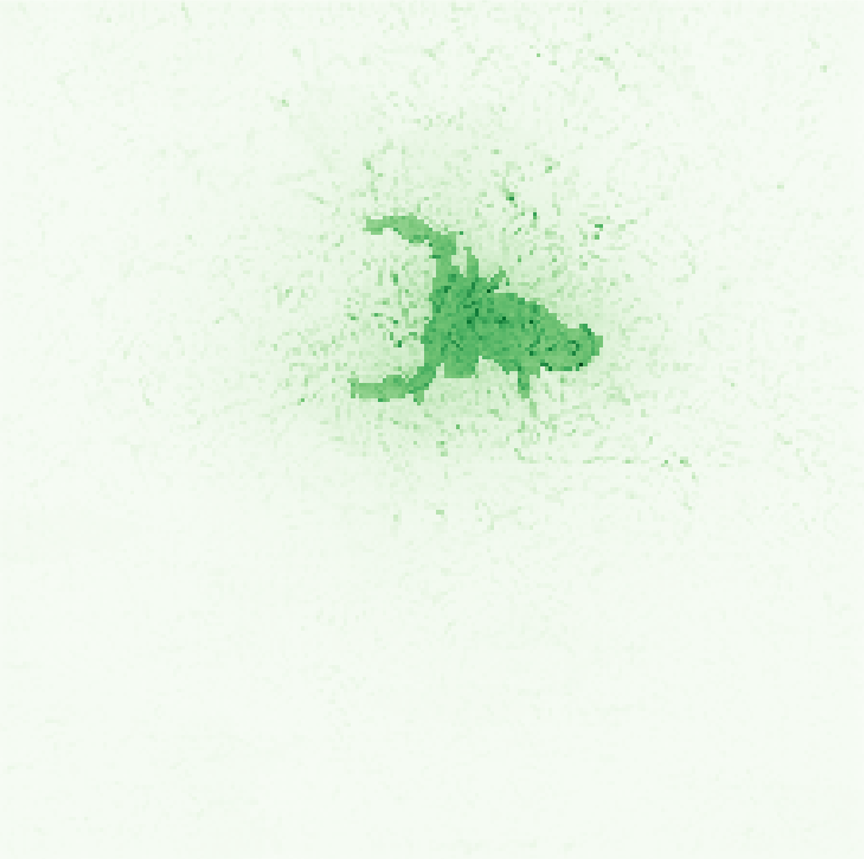} & 
\includegraphics[width=14.5mm]{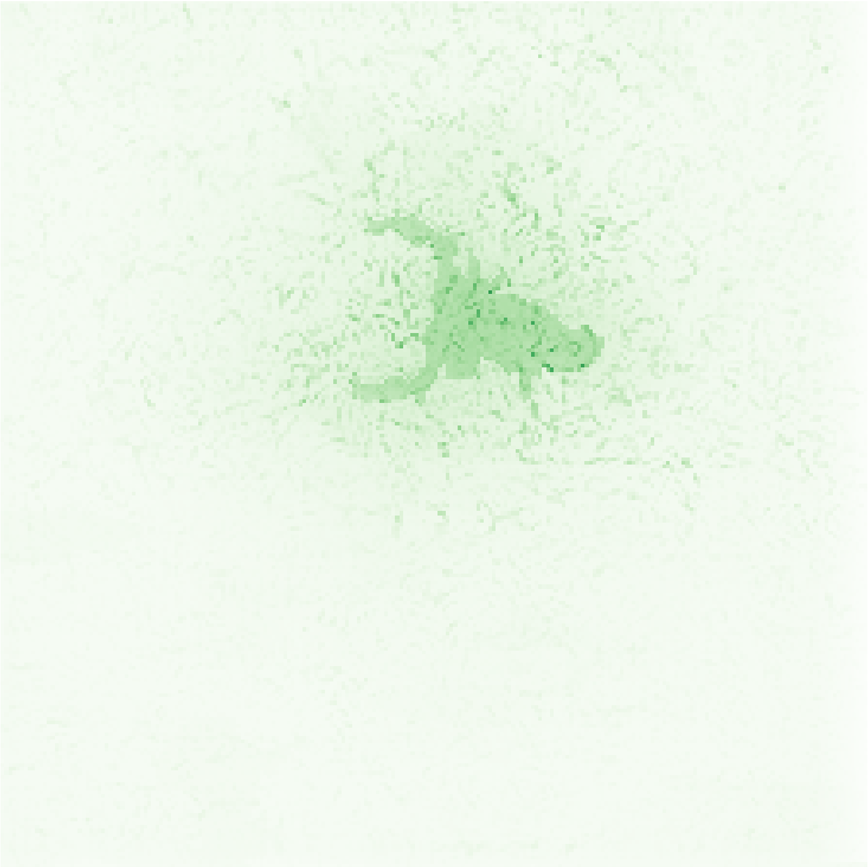} & 
\includegraphics[width=14.5mm]{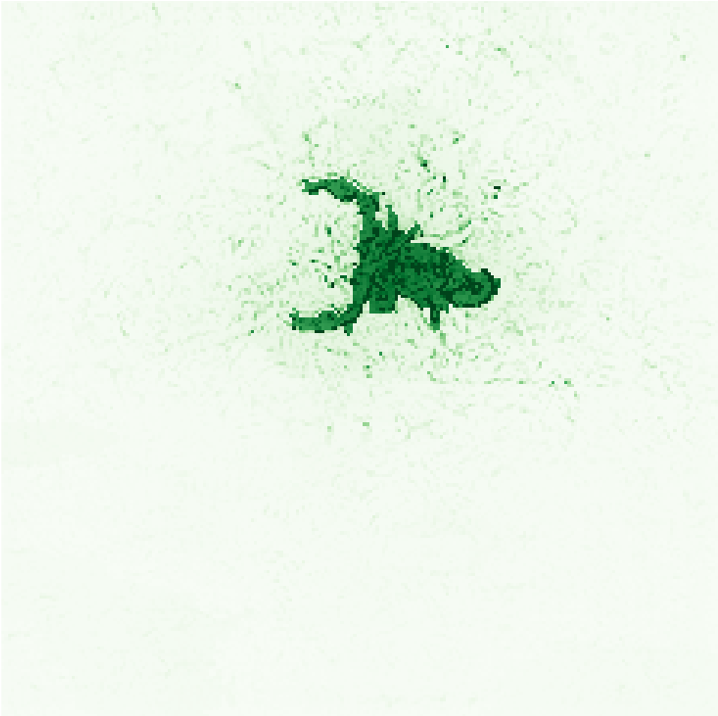} \\

\includegraphics[width=14.5mm]{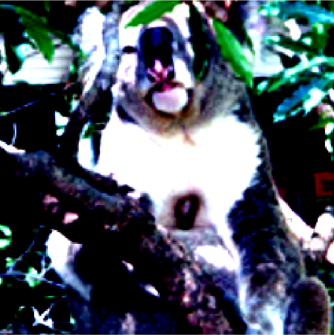} & 
\includegraphics[width=14.5mm]{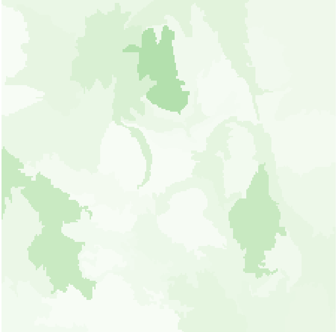} & 
\includegraphics[width=14.5mm]{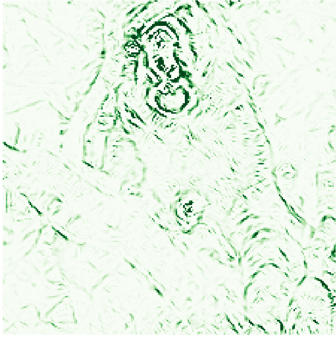} & 
\includegraphics[width=14.5mm]{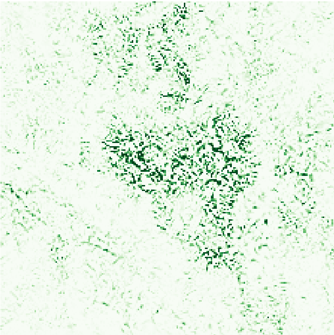} & 
\includegraphics[width=14.5mm]{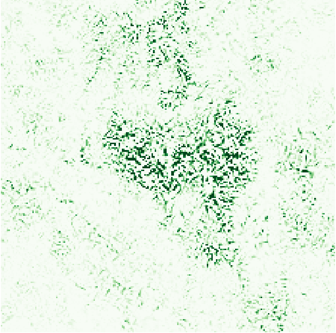} & 
\includegraphics[width=14.5mm]{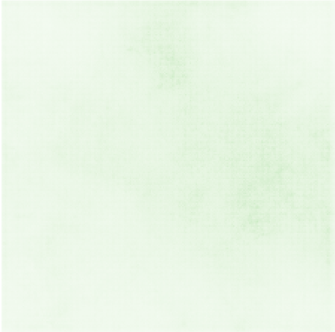} & 
\includegraphics[width=14.5mm]{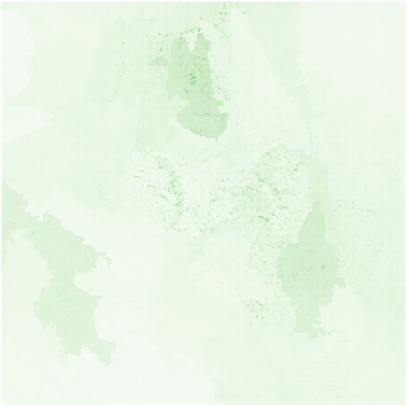} & 
\includegraphics[width=14.5mm]{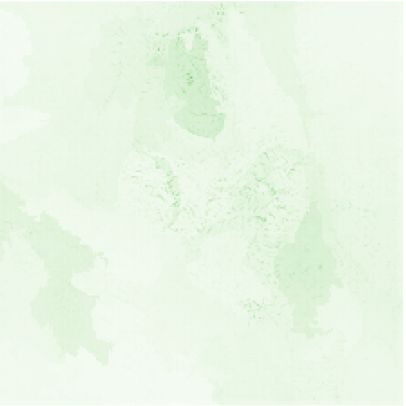} & 
\includegraphics[width=14.5mm]{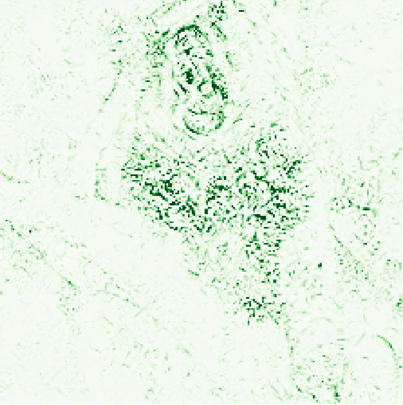} \\

\includegraphics[width=14.5mm]{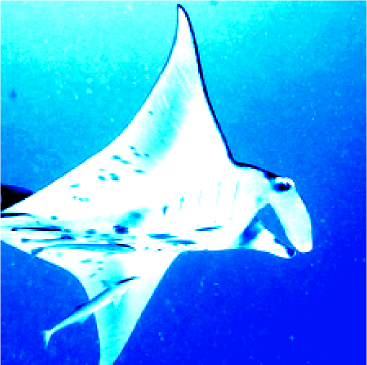} & 
\includegraphics[width=14.5mm]{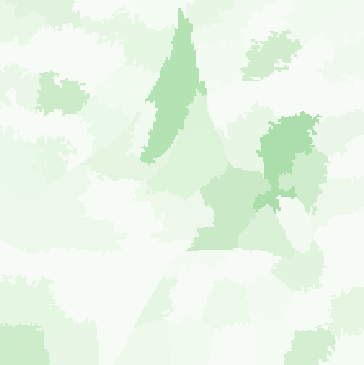} & 
\includegraphics[width=14.5mm]{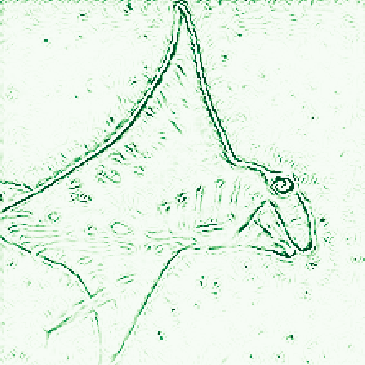} & 
\includegraphics[width=14.5mm]{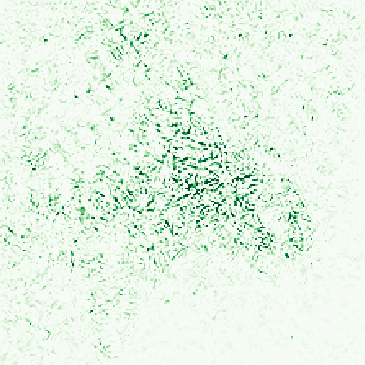} & 
\includegraphics[width=14.5mm]{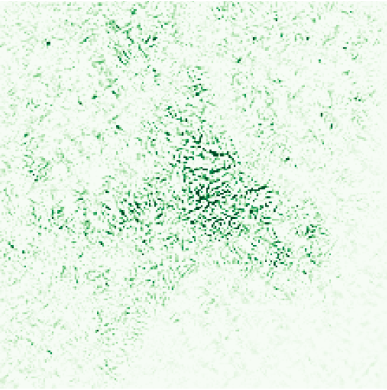} &\includegraphics[width=14.5mm]{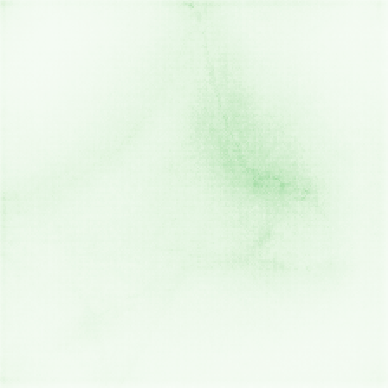} & 
\includegraphics[width=14.5mm]{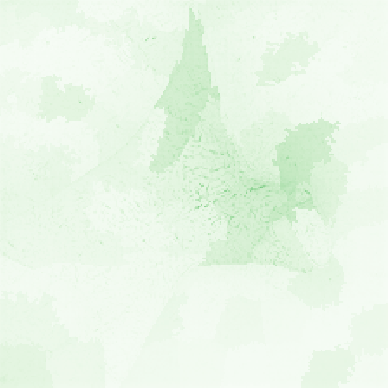} & 
\includegraphics[width=14.5mm]{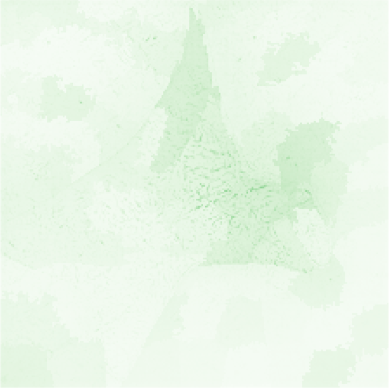} & 
\includegraphics[width=14.5mm]{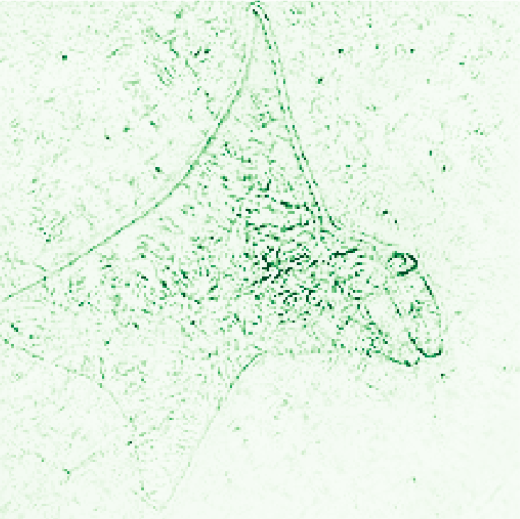} \\

\includegraphics[width=14.5mm]{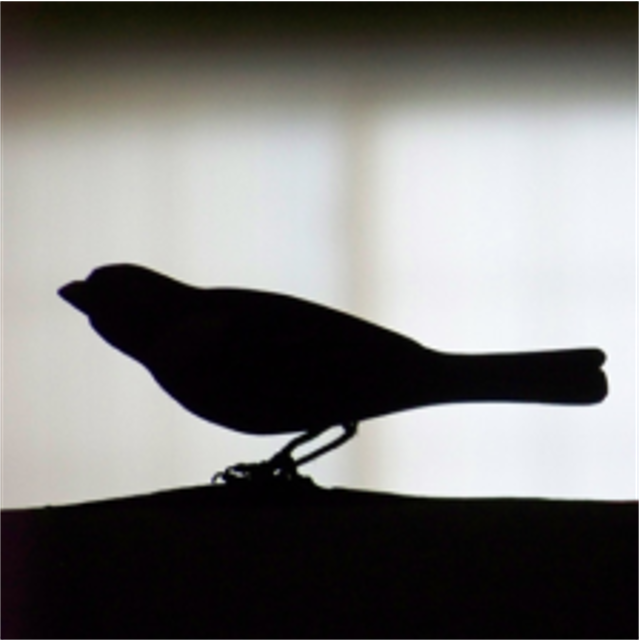} & 
\includegraphics[width=14.5mm]{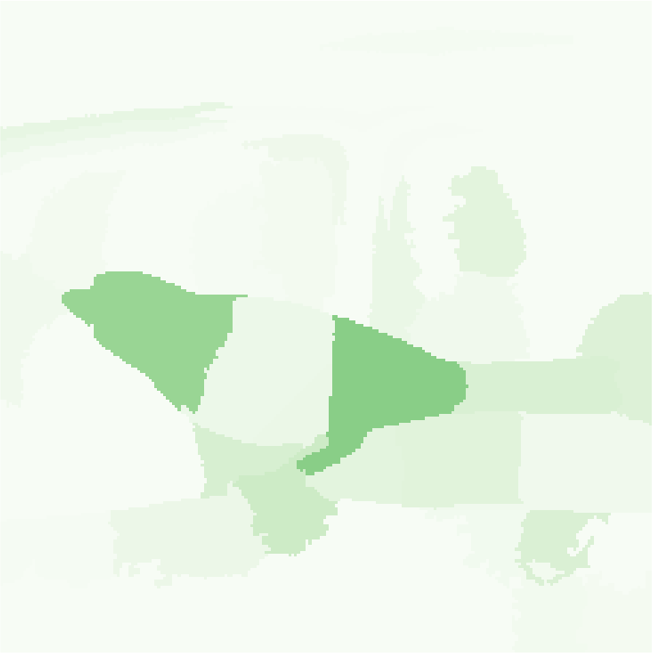} & 
\includegraphics[width=14.5mm]{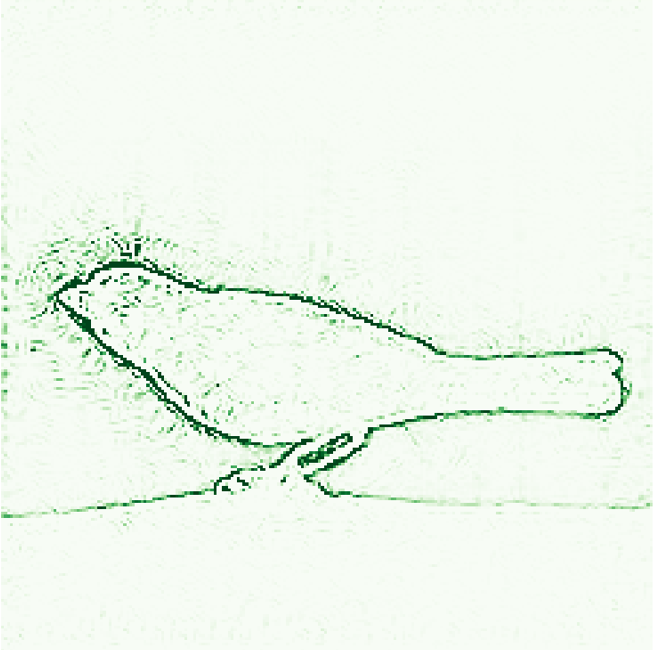} & 
\includegraphics[width=14.5mm]{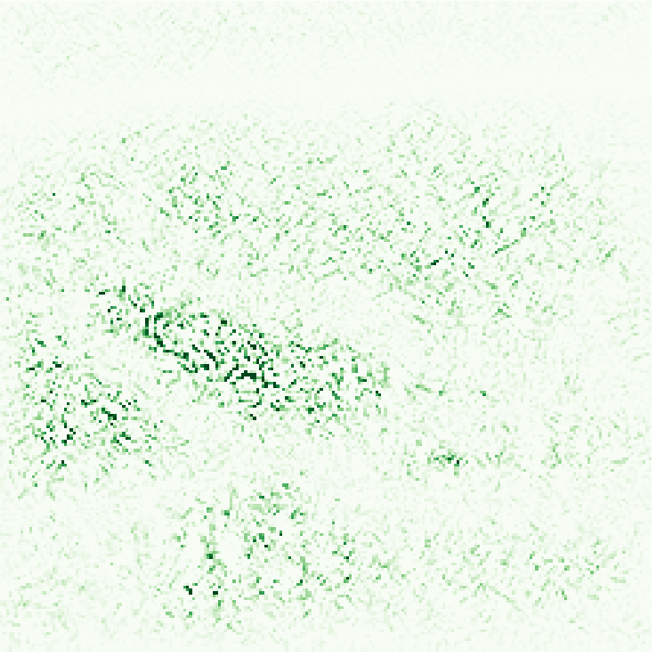} & 
\includegraphics[width=14.5mm]{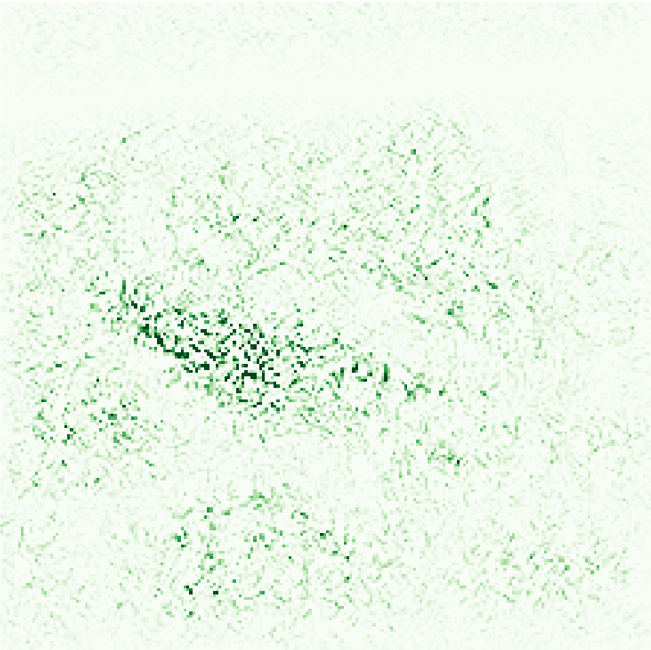} & 
\includegraphics[width=14.5mm]{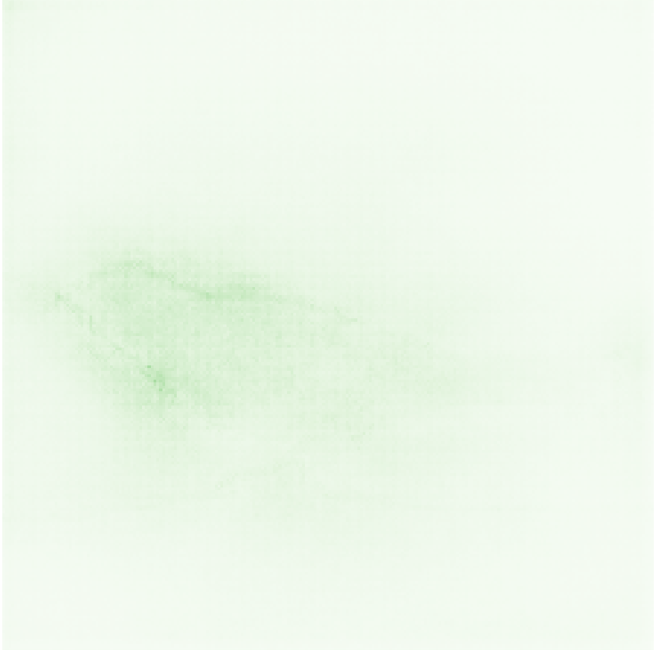} & 
\includegraphics[width=14.5mm]{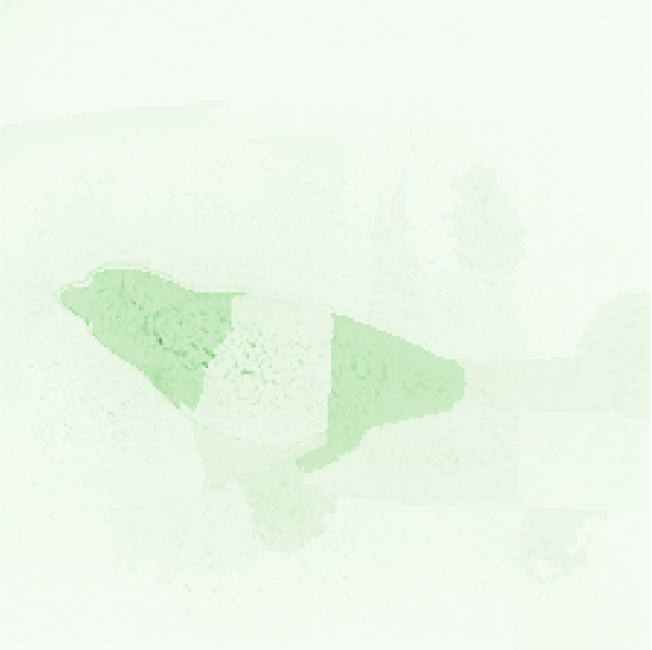} & 
\includegraphics[width=14.5mm]{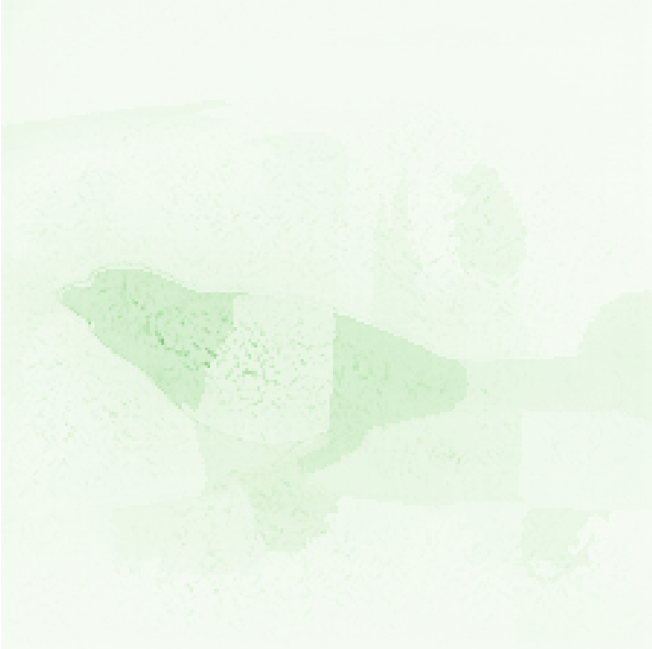} & 
\includegraphics[width=14.5mm]{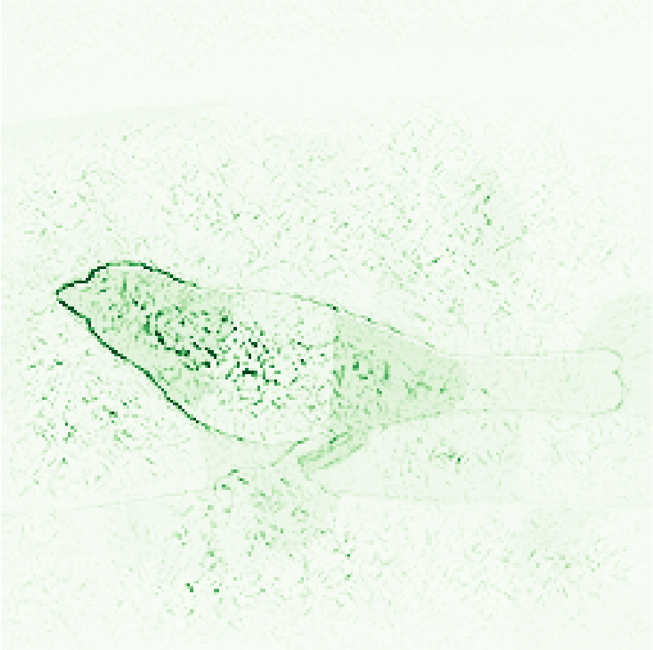} \\

\multicolumn{9}{c}{\textbf{With} noisy feature attribution maps in the ensemble} \\ 

\includegraphics[width=14.5mm]{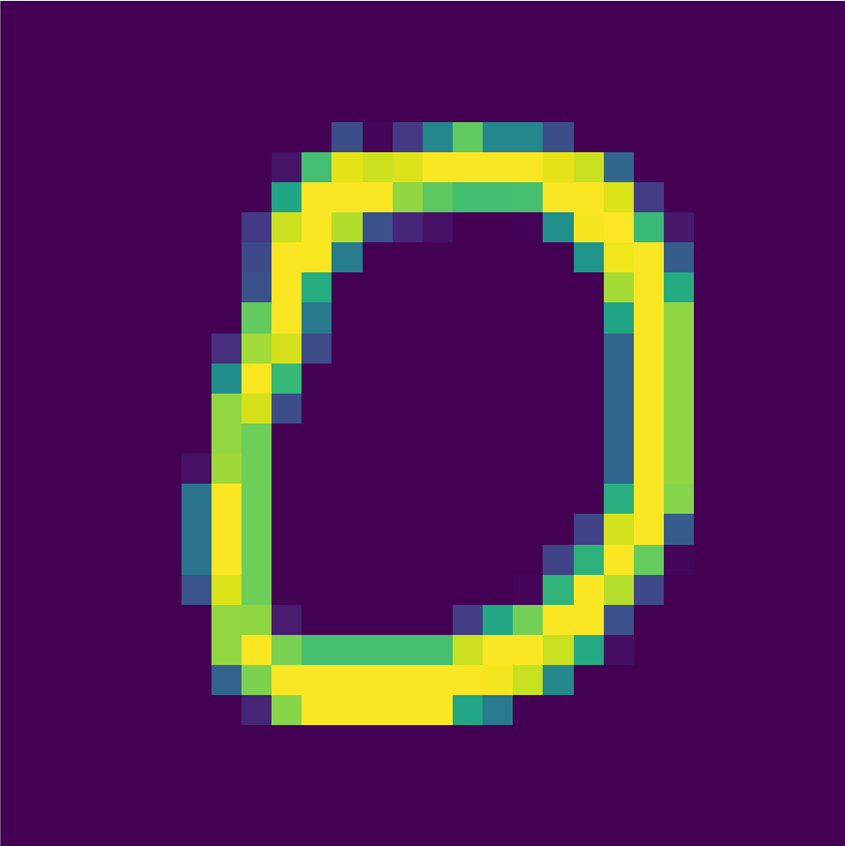} & 
\includegraphics[width=14.5mm]{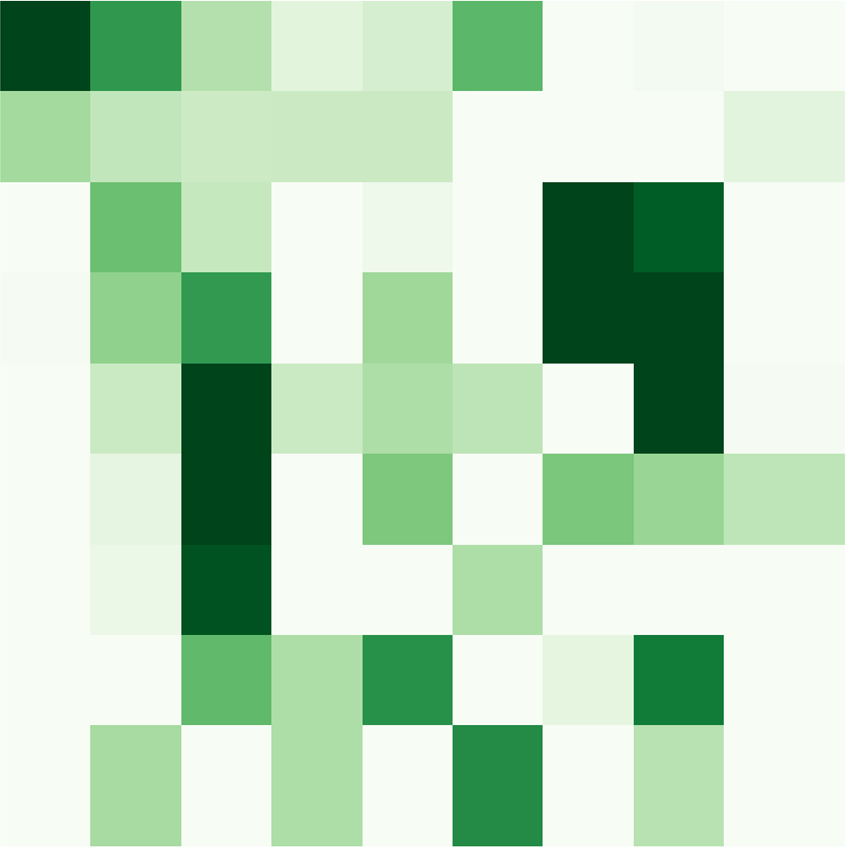} & 
\includegraphics[width=14.5mm]{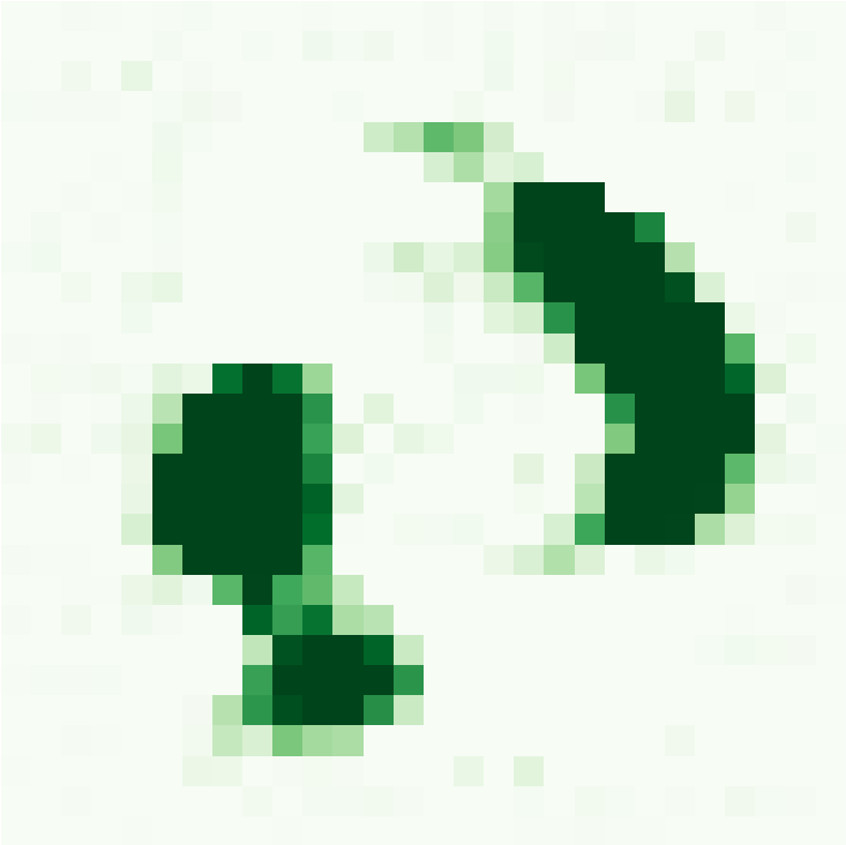} & 
\includegraphics[width=14.5mm]{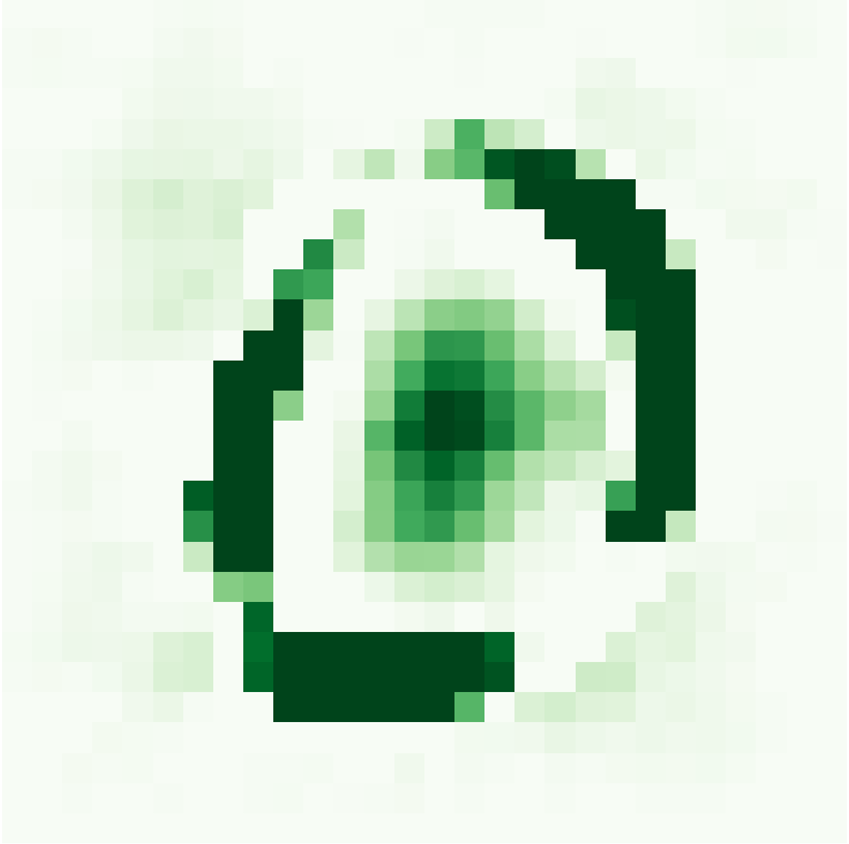} & 
\includegraphics[width=14.5mm]{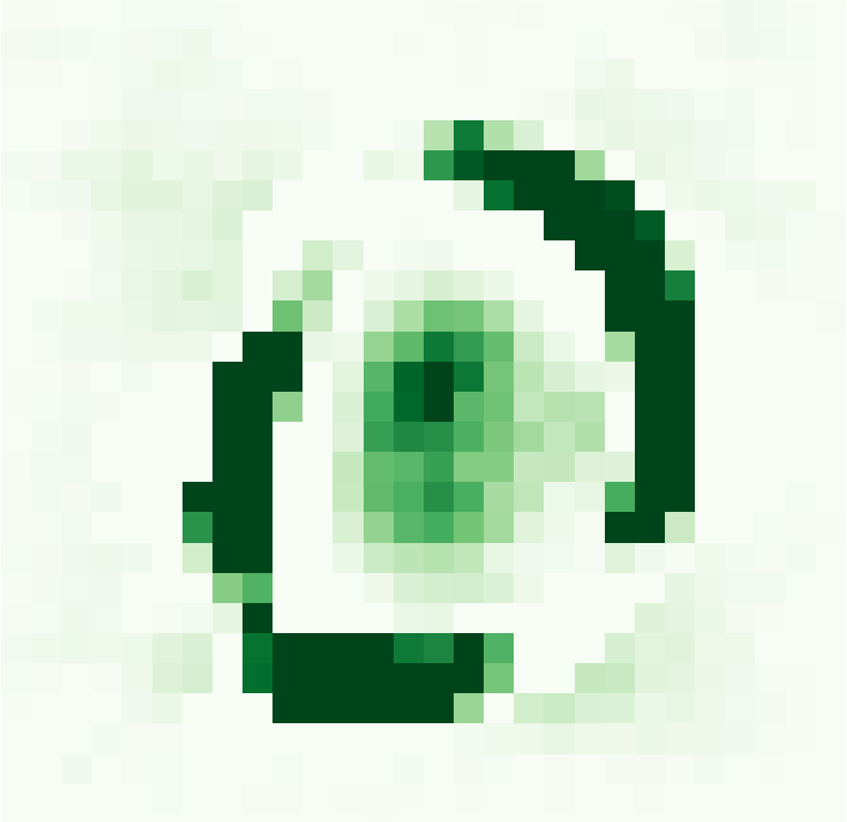} & 
\includegraphics[width=14.5mm]{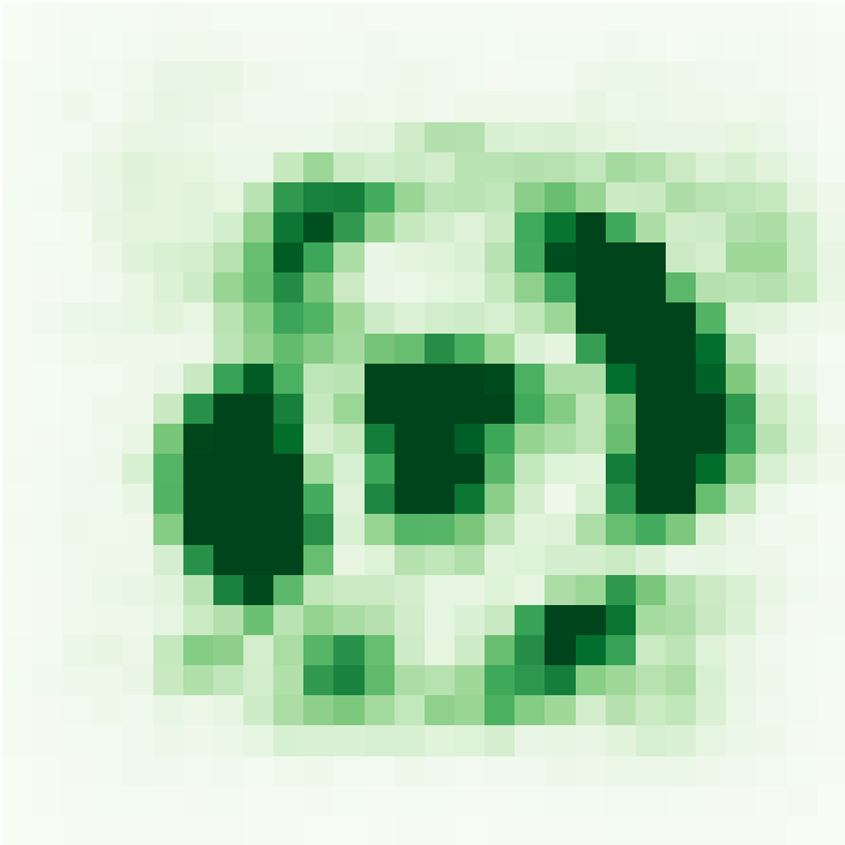} & 
\includegraphics[width=14.5mm]{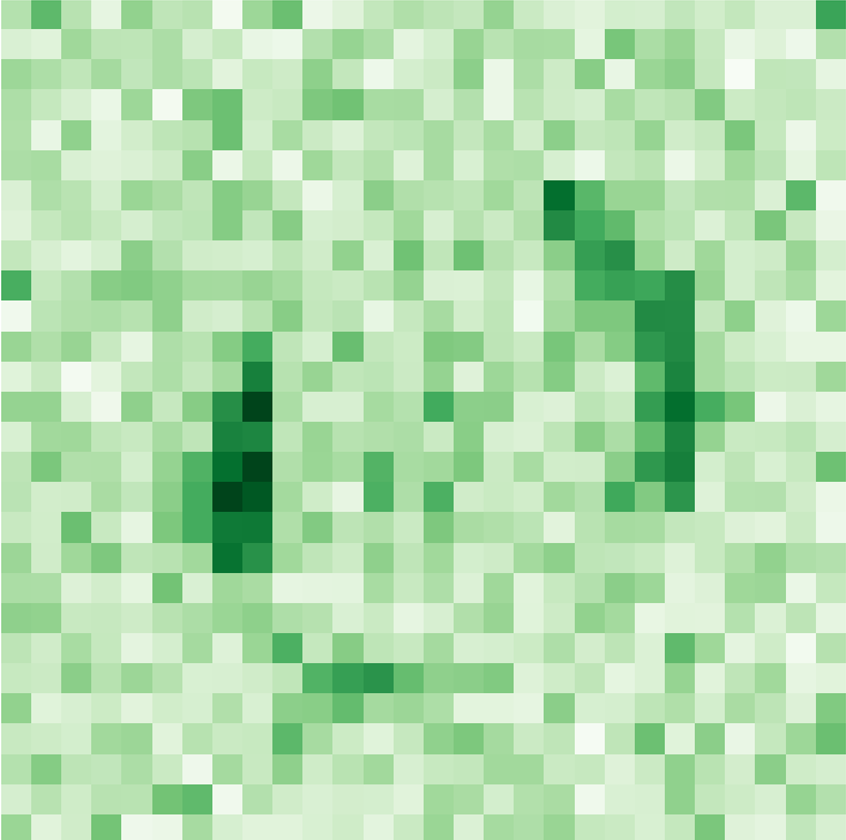} & 
\includegraphics[width=14.5mm]{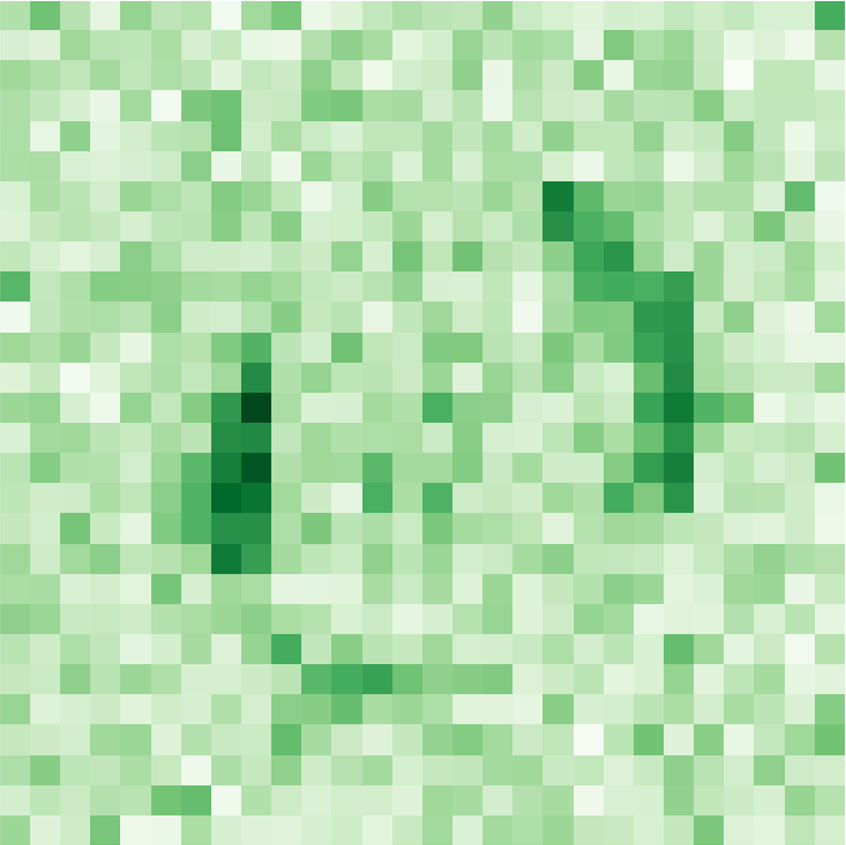} & 
\includegraphics[width=14.5mm]{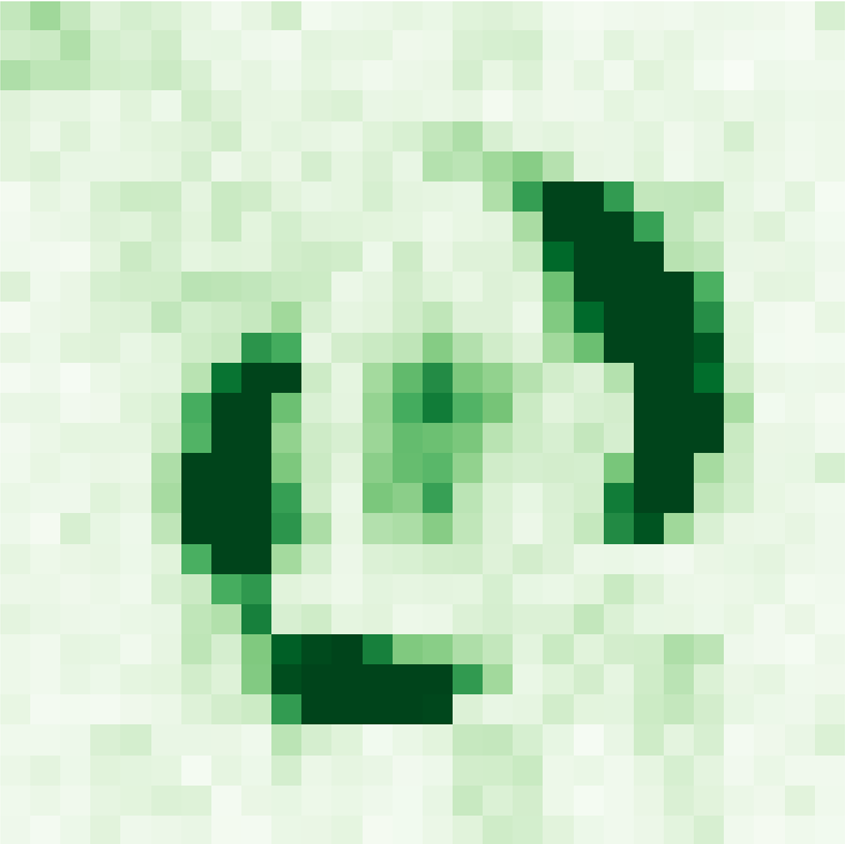} \\

\includegraphics[width=14.5mm]{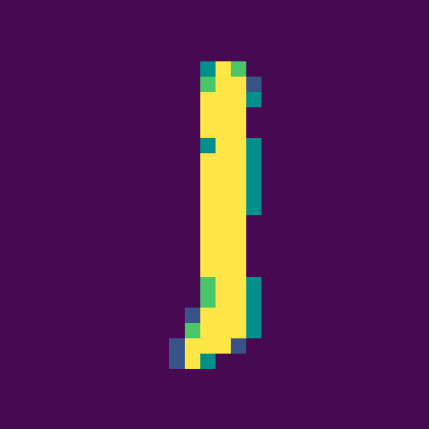} & 
\includegraphics[width=14.5mm]{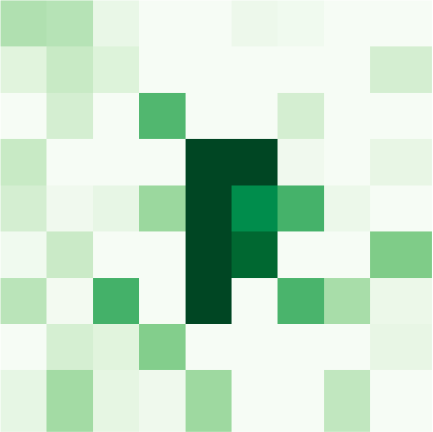} & 
\includegraphics[width=14.5mm]{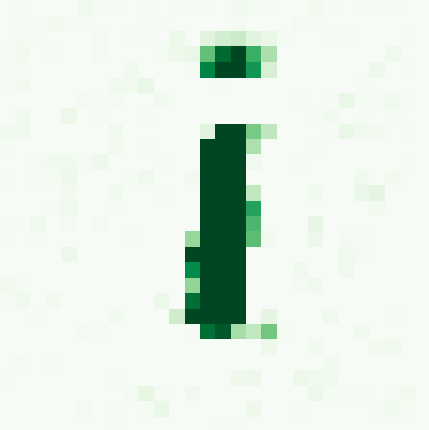} & 
\includegraphics[width=14.5mm]{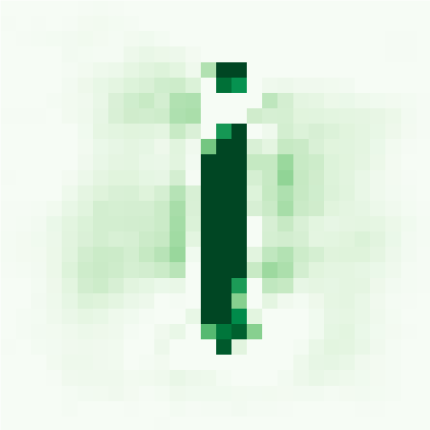} & 
\includegraphics[width=14.5mm]{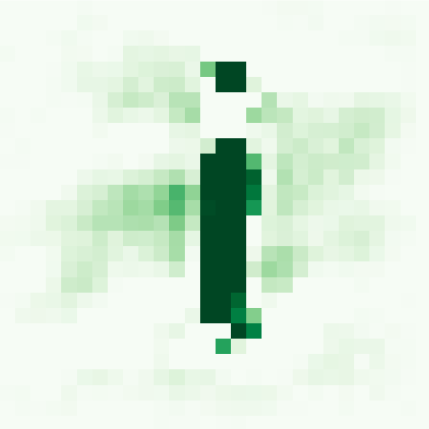} & 
\includegraphics[width=14.5mm]{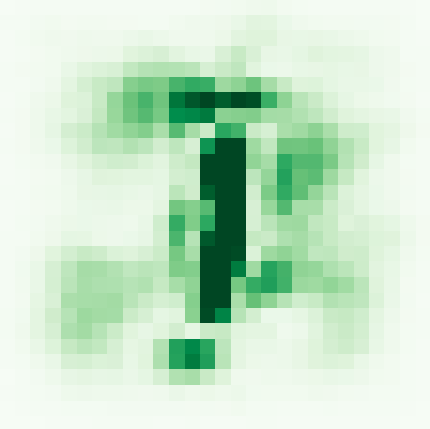} & 
\includegraphics[width=14.5mm]{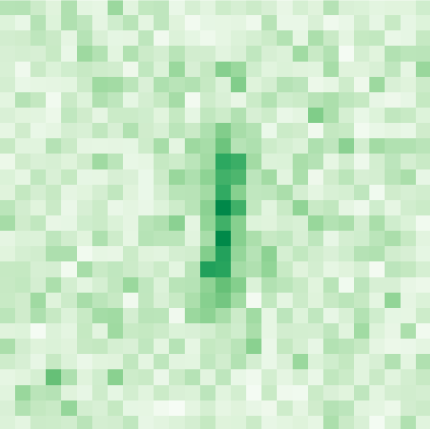} & 
\includegraphics[width=14.5mm]{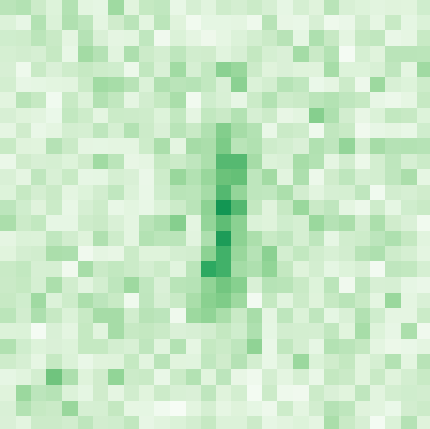} & 
\includegraphics[width=14.5mm]{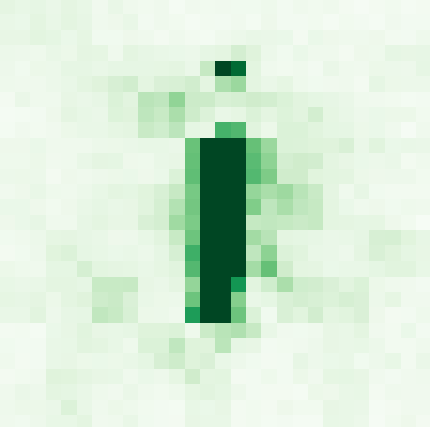} \\

\includegraphics[width=14.5mm]{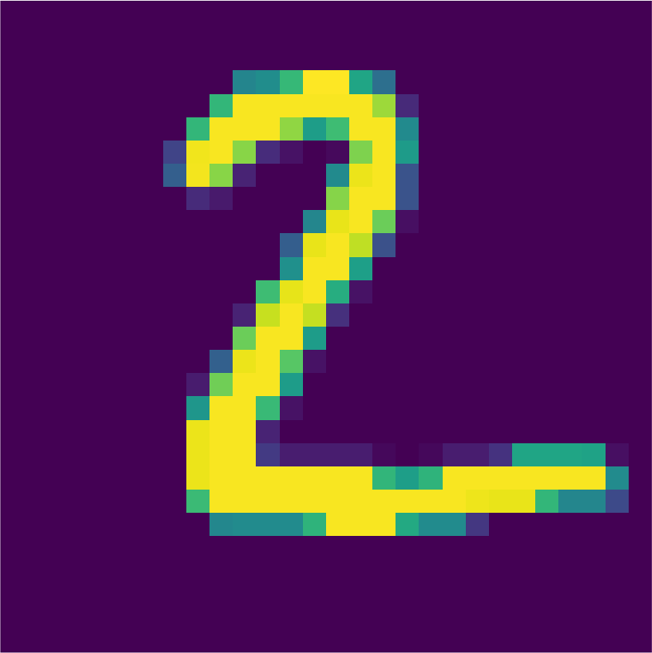} & 
\includegraphics[width=14.5mm]{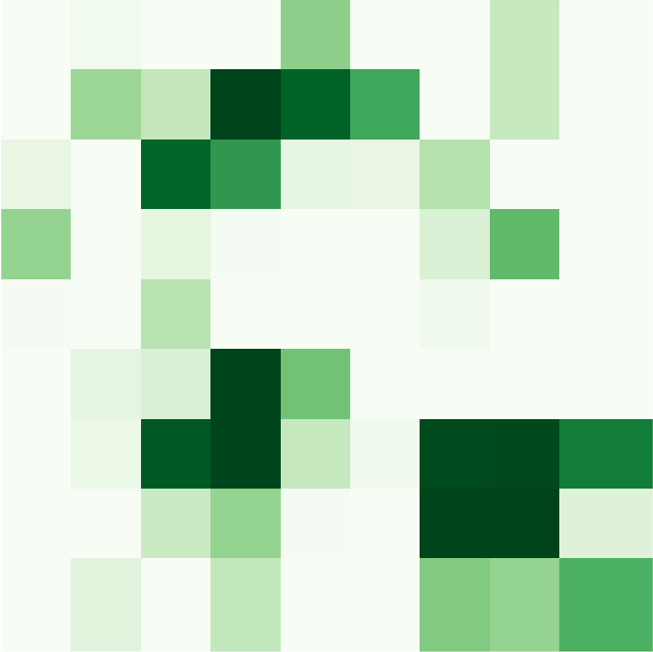} & 
\includegraphics[width=14.5mm]{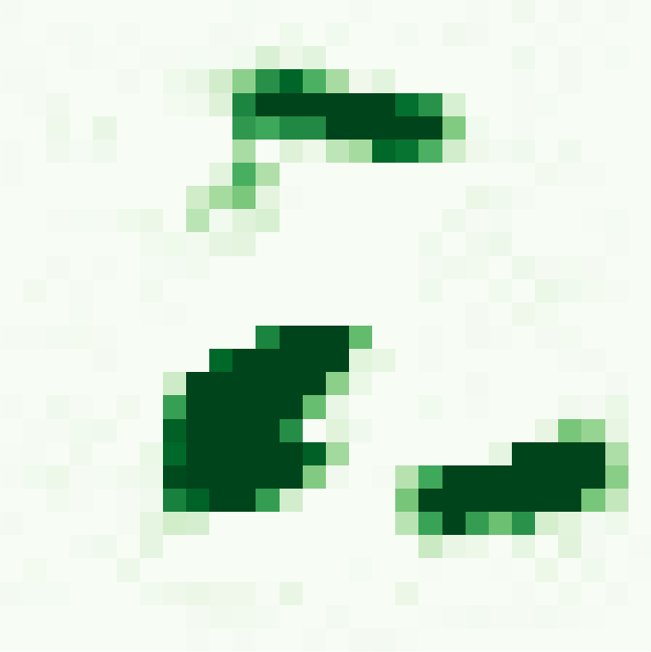} & 
\includegraphics[width=14.5mm]{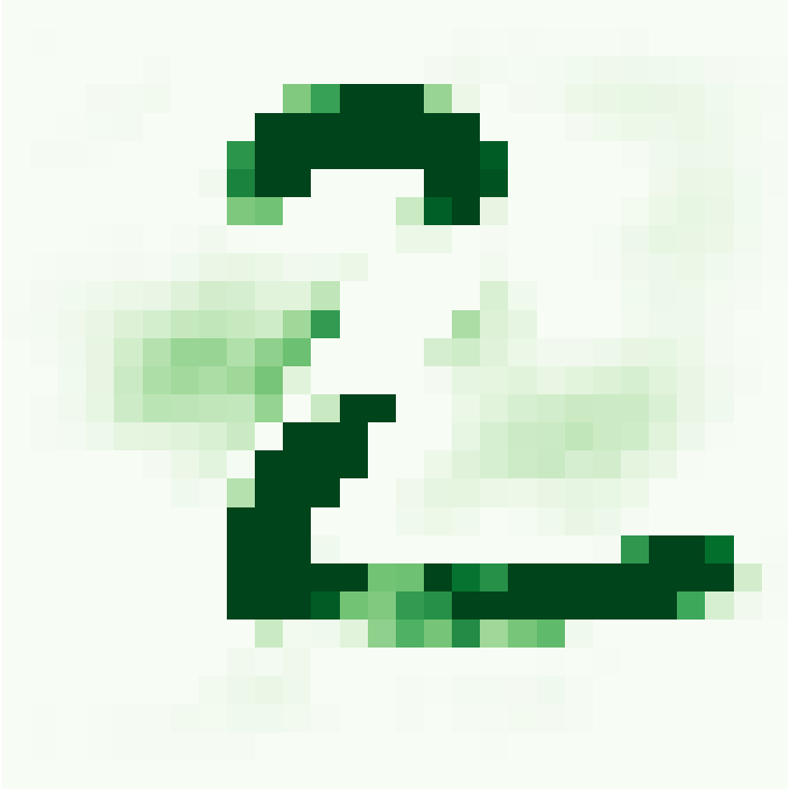} & 
\includegraphics[width=14.5mm]{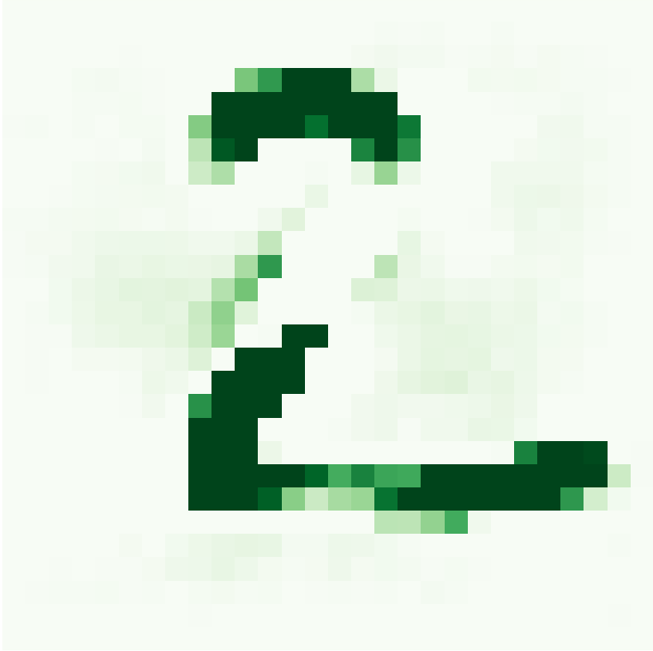} & 
\includegraphics[width=14.5mm]{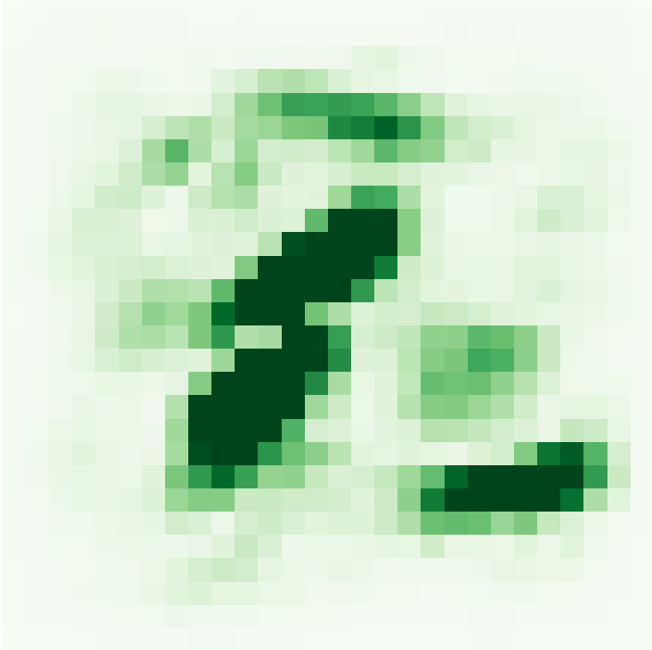} & 
\includegraphics[width=14.5mm]{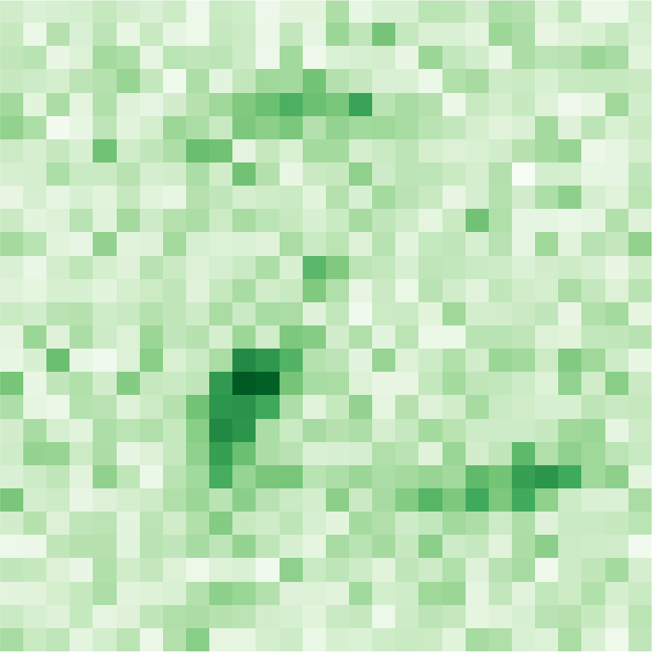} & 
\includegraphics[width=14.5mm]{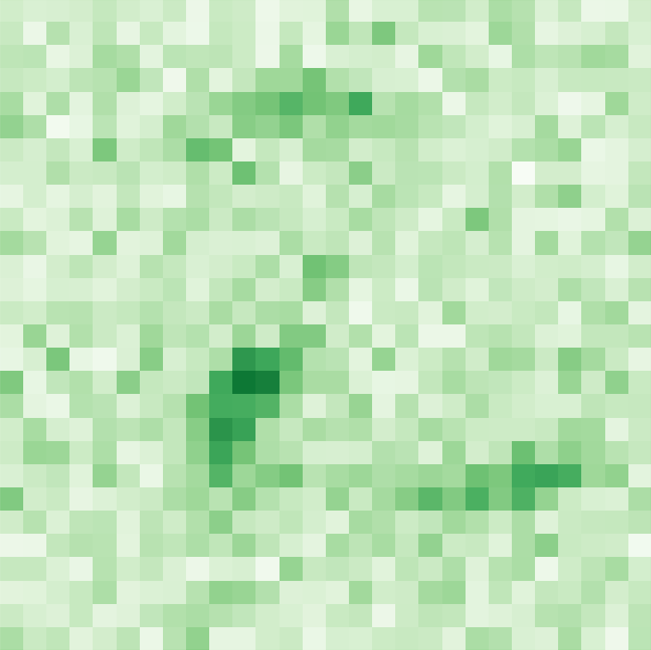} & 
\includegraphics[width=14.5mm]{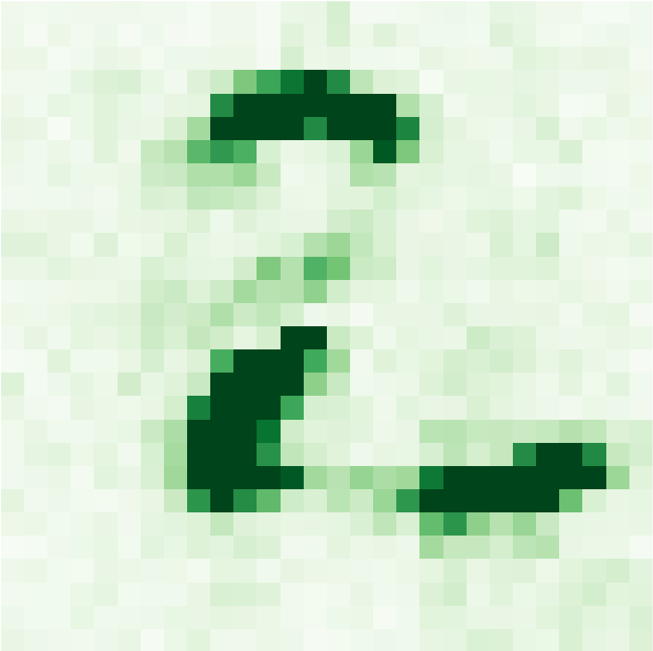} \\

\includegraphics[width=14.5mm]{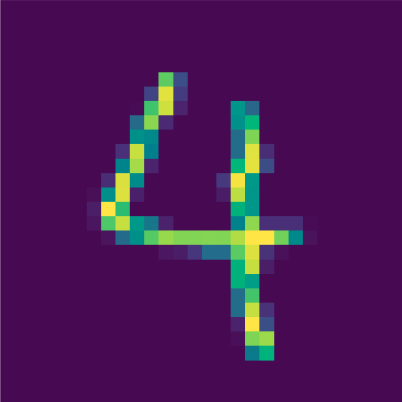} & 
\includegraphics[width=14.5mm]{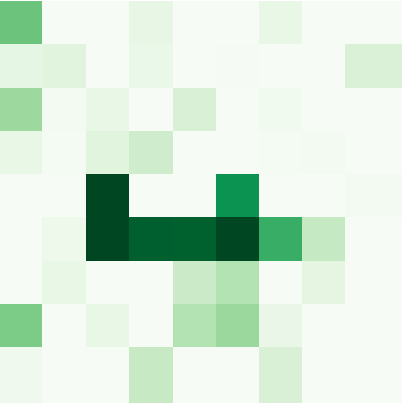} & 
\includegraphics[width=14.5mm]{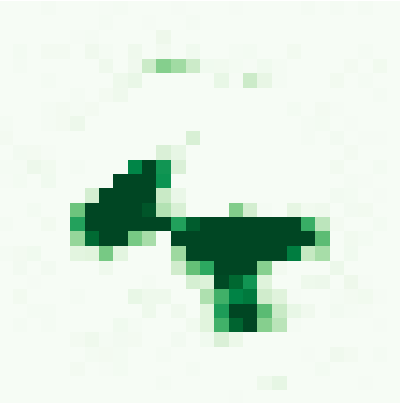} & 
\includegraphics[width=14.5mm]{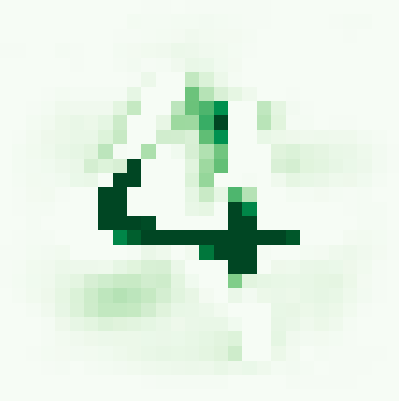} & 
\includegraphics[width=14.5mm]{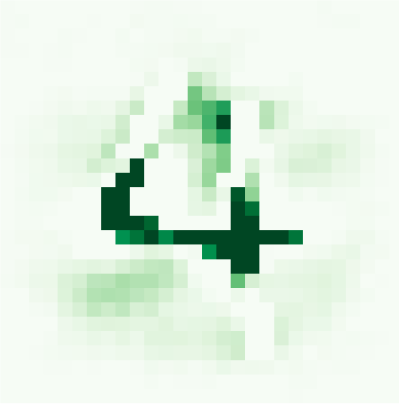} & 
\includegraphics[width=14.5mm]{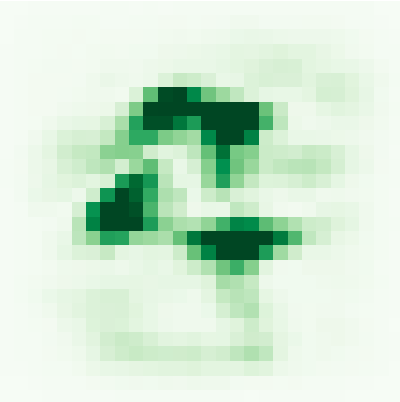} & 
\includegraphics[width=14.5mm]{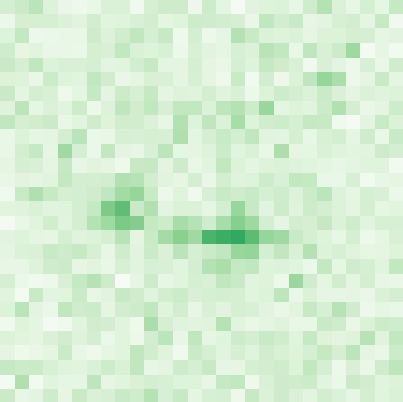} & 
\includegraphics[width=14.5mm]{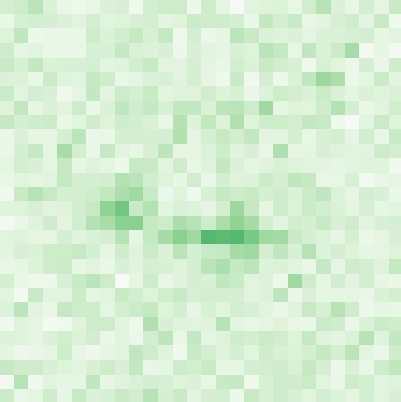} & 
\includegraphics[width=14.5mm]{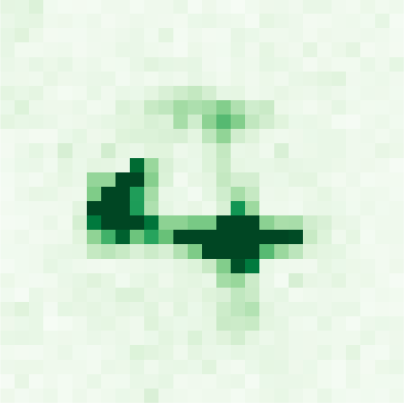} \\

\includegraphics[width=14.5mm]{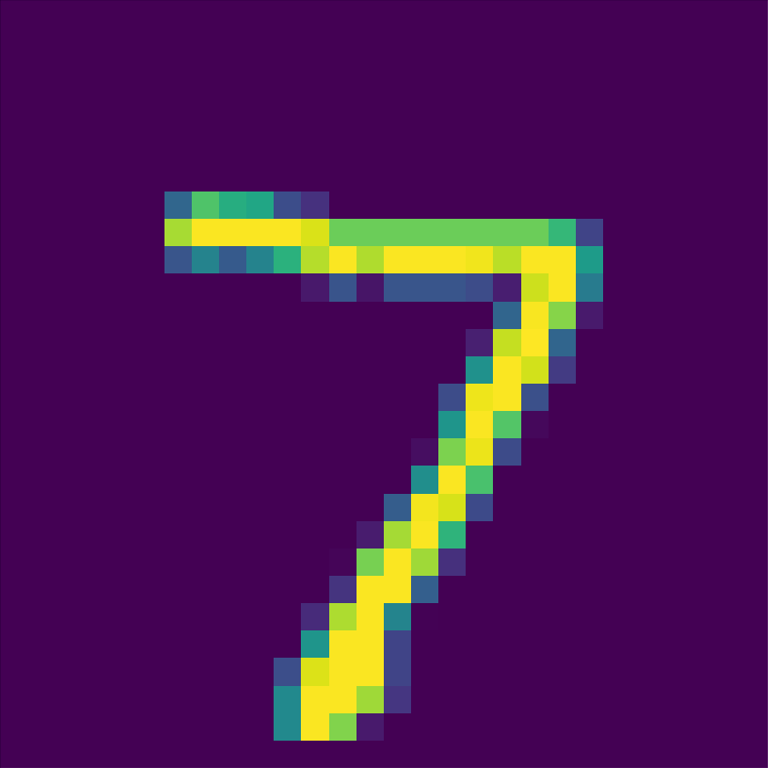} & 
\includegraphics[width=14.5mm]{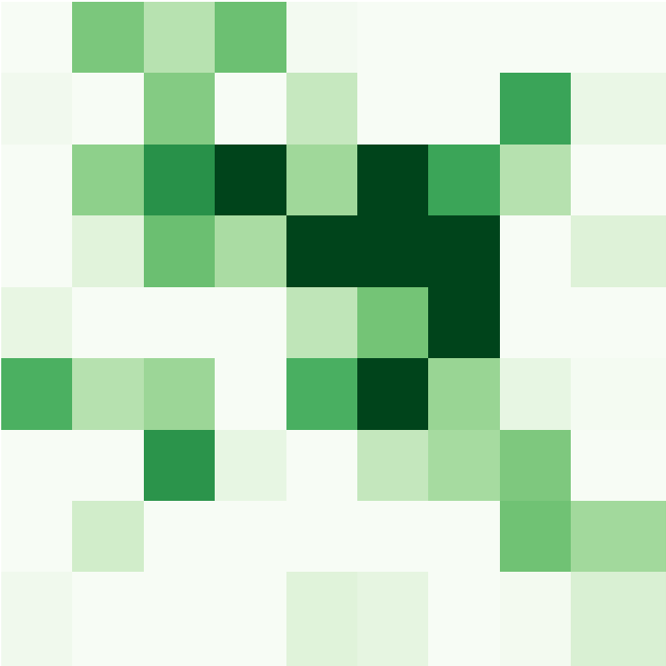} & 
\includegraphics[width=14.5mm]{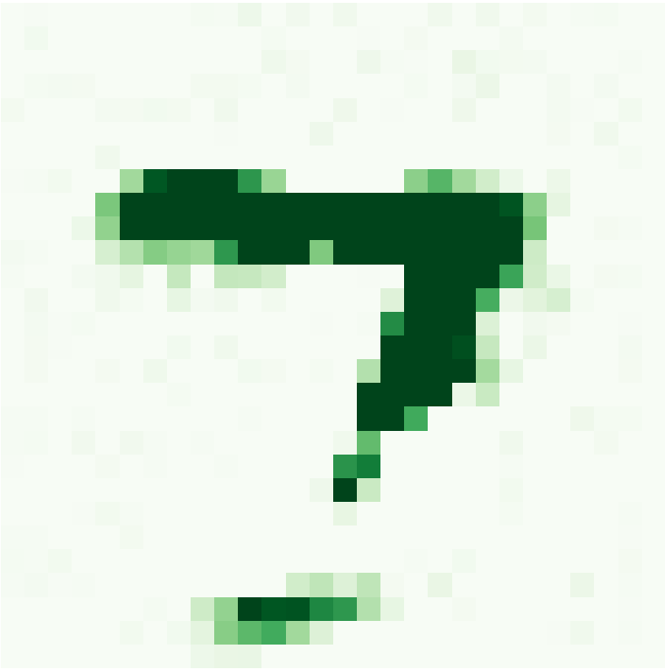} & 
\includegraphics[width=14.5mm]{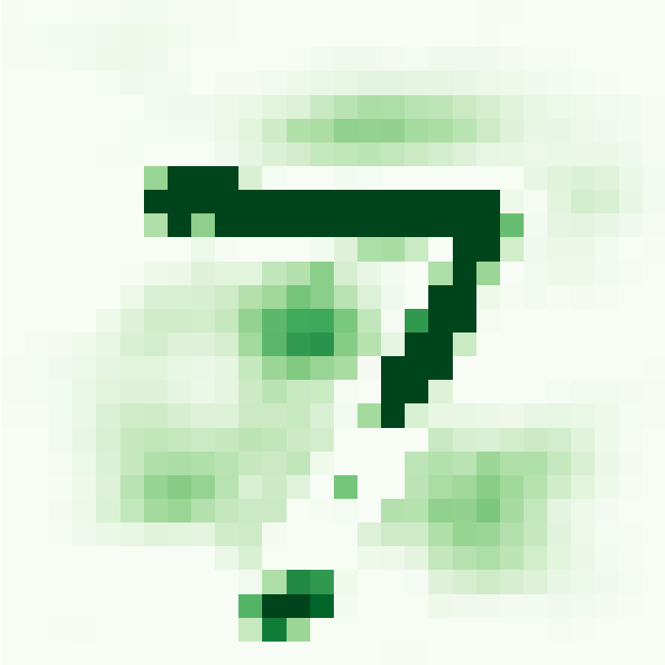} & 
\includegraphics[width=14.5mm]{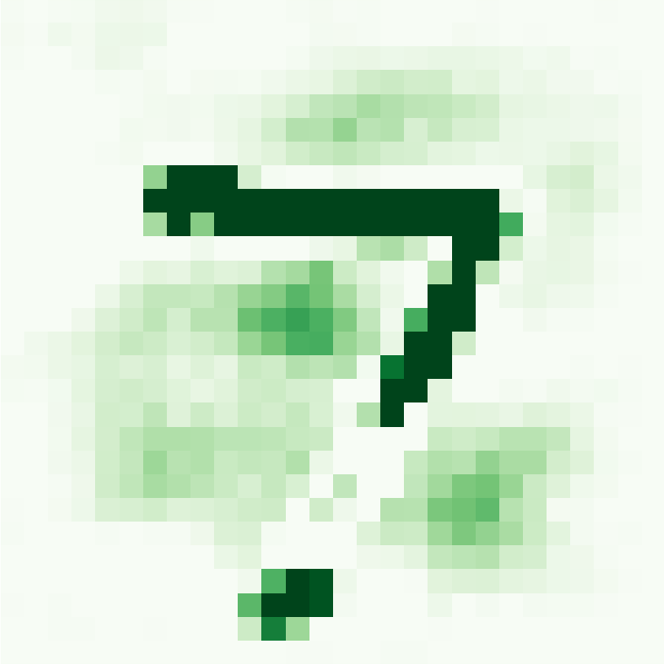} & 
\includegraphics[width=14.5mm]{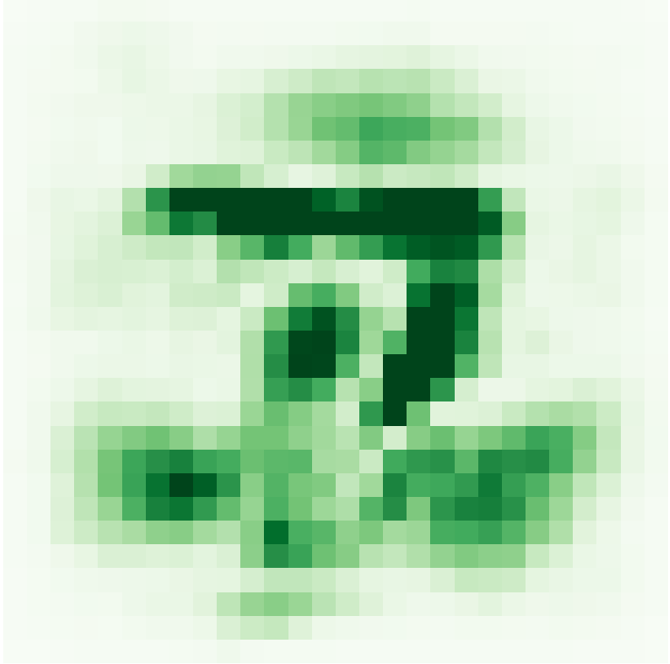} & 
\includegraphics[width=14.5mm]{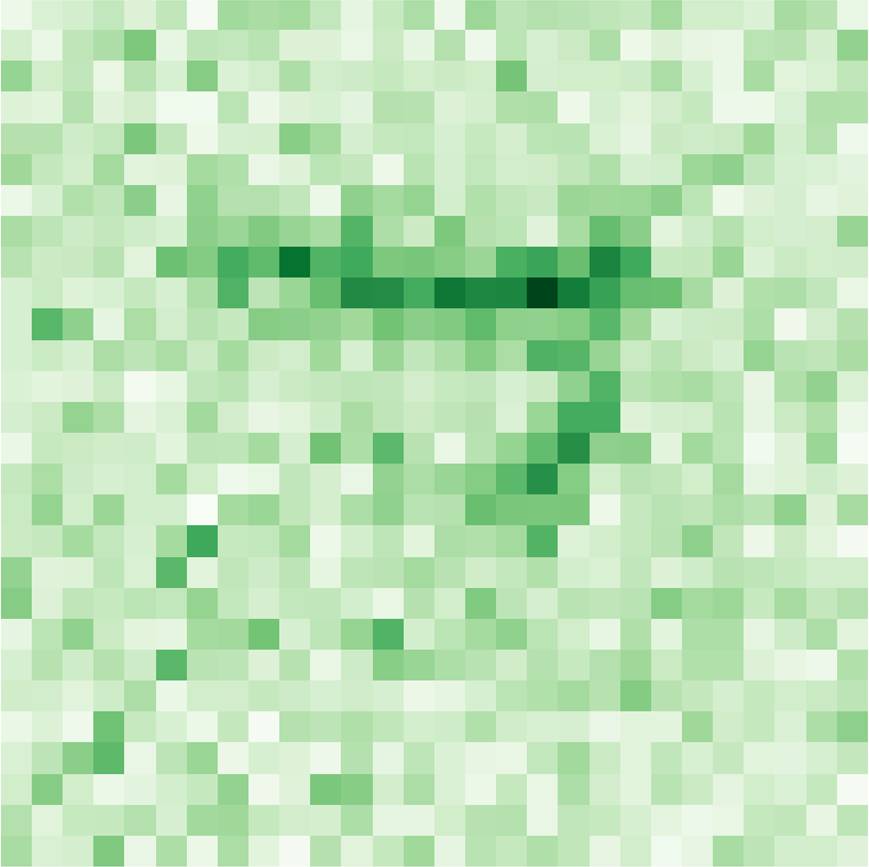} & 
\includegraphics[width=14.5mm]{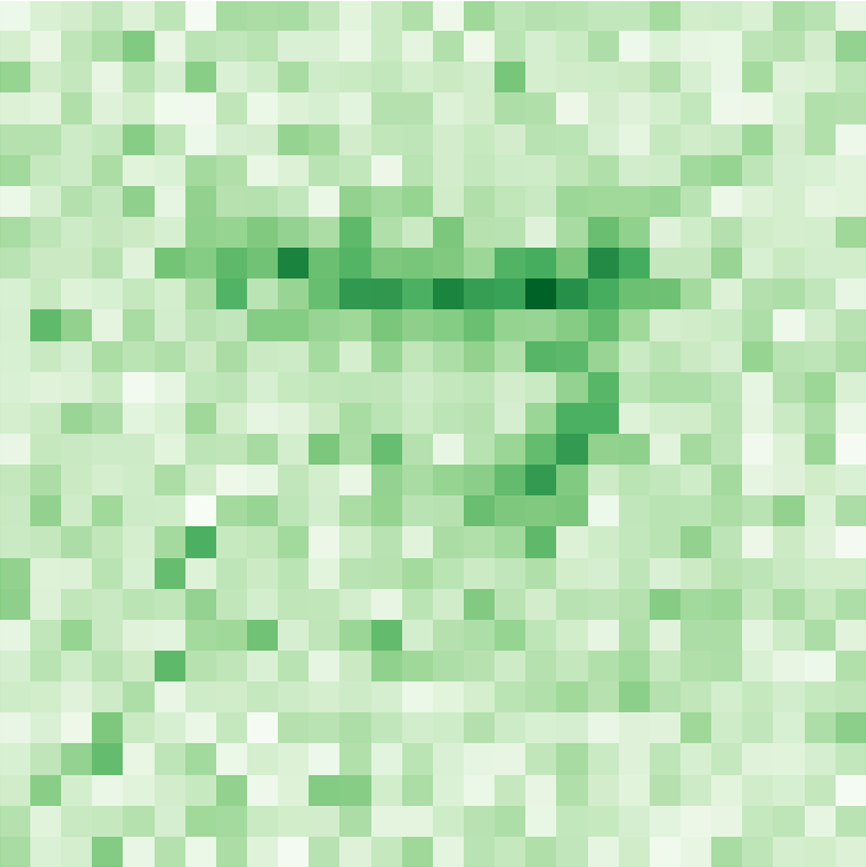} & 
\includegraphics[width=14.5mm]{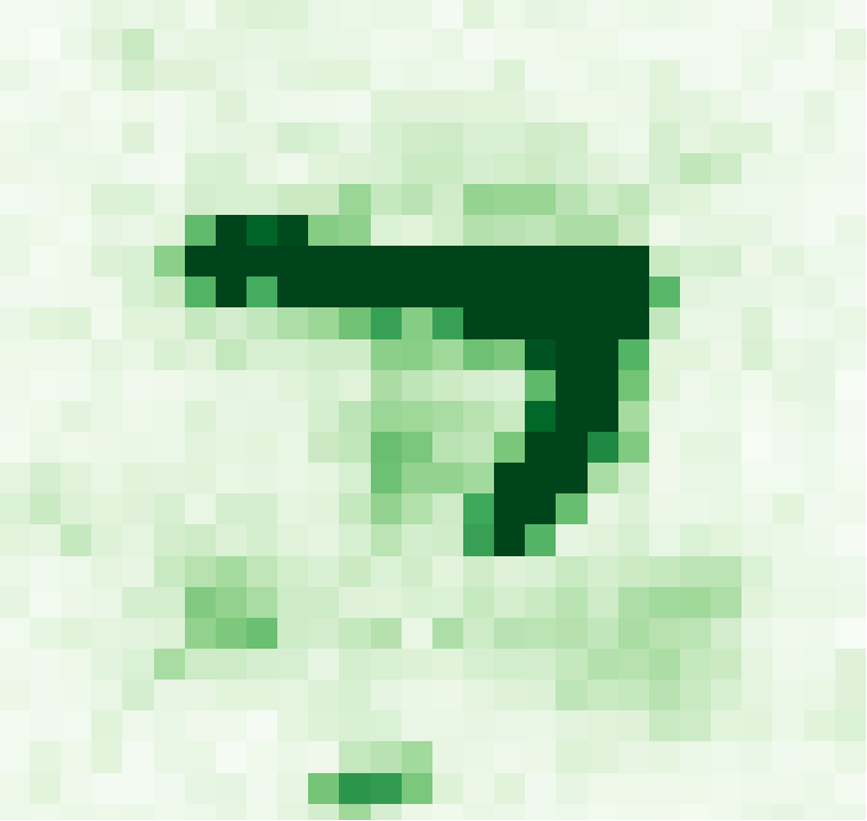} \\

\end{tabularx}
\caption{A visual comparison between baseline methods and ensemble methods on ImageNet \cite{deng2009imagenet} and MNIST \cite{lecun1998mnist} datasets.}
\label{tab:visual_experiment}
\end{table*}

\subsection{Visual Inspection for Image Data}
In our first experiment, a visual evaluation on images from ImageNet~\cite{deng2009imagenet} and MNIST~\cite{lecun1998mnist} is performed for several baseline and ensemble methods. We provide benchmark outcomes for two settings, with and without fifteen noisy baseline explanations in an ensemble. The results are depicted in Table \ref{tab:visual_experiment}.

\paragraph{Without artificial noise in the ensemble.}
We select four samples from ImageNet dataset~\cite{deng2009imagenet}  and five baseline explanation methods for the ensemble models: LIME~\cite{ribeiro2016should}, Guided Backpropagation (GB)~\cite{springenberg2015striving}, Integrated Gradients (IG)~\cite{SundararajanTY17}, Gradient SHAP (GS)~\cite{lundberg2017unified}, and SmoothGrad (SG) \cite{smilkov2017smoothgrad}. We compare the proposed RBM ensemble strategy to simple mean and variance ensembles~\cite{rieger2020aggregating}. The results in Table \ref{tab:visual_experiment} show that our approach produces sharp and visually appealing saliency maps in comparison to other ensemble baselines. In comparison to the baseline explanation methods, the proposed ensemble technique seems to produce more reliable and robust results by highlighting commonalities among the baseline methods and by mitigating the noise coming from the single baseline methods.  

\paragraph{With artificial noise in the ensemble.}

We challenge the discussed approaches by adding fifteen baselines with random noise sampled from the standard normal distribution $\textbf{e}_{rand} \sim \mathcal{N}(0,\,1)$ to the ensemble. 
The results in Table \ref{tab:visual_experiment} reveal that the proposed RBM-based aggregation method mitigates noise and hence results in more robust saliency maps in comparison to the other ensemble baselines.

\subsection{Pixel Perturbation Experiment}
\label{sec:pixe_perturb}

In the first quantitative experiments, we compare multiple baseline models and ensemble methods on the CIFAR10 dataset \cite{krizhevsky2009learning} by removing the most important pixels (according to a scoring function) and reporting the area under a curve score (DAUC). In addition, we also follow the approach of inserting the most important pixels into an empty image and again report the area under a curve (IAUC). Thus, an ideal feature scoring function has a large IAUC and low DAUC. These benchmark methods well accepted by the research community~\cite{petsiuk2018rise}. 
For this experiment, we select the following algorithms as baseline explanation methods: Gradient SHAP \cite{lundberg2017unified}, DeepLIFT \cite{shrikumar2017learning}, LIME \cite{ribeiro2016should}, Saliency maps \cite{simonyan2014}, SmoothGrad \cite{smilkov2017smoothgrad}, Integrated Gradients \cite{SundararajanTY17}, Guided Backpropagation \cite{springenberg2015striving}. 
As suggested in~\cite{zhang2014learning}, we add the original image as a baseline to the ensemble. However, according to our experiments, adding th original image to the ensemble does improve the overall ensemble performance. 
We report all scores for the baseline approaches and the ensemble methods in Table \ref{tab:quant_experiments}.

\subsection{IROF Experiment}

In~\cite{rieger2020aggregating} the authors propose the IROF  measure as an extension to the work~\cite{SamekBMBM15}. The main idea of the IROF benchmark is as follows: the image is divided into superpixels using the SLIC algorithm~\cite{Achanta2010}. Superpixels are regional blocks of pixels within an image where the contained pixels share a high similarity measure among each other. The relevancy for each superpixel is calculated by averaging over the attribution scores over all contained pixels (inside the superpixel). After, the superpixels are sorted descending by their relevancy. The entire superpixels are gradually replaced by a baseline and sent through the network again to measure the new recognition quality for the modified image wrt. to the target label. For more accurate attribution methods, the recognition quality decreases faster, and thus the area under the curve is lower. The IROF score is defined as the area over the curve (AOC): $AOC = 1-AUC$. Higher values, therefore, indicate a better attribution. We use the same baseline methods as in the pixel perturbation experiment (Sec. \ref{sec:pixe_perturb}). The results are listed in Table \ref{tab:quant_experiments}. 

% In this work we also use a modified version of the IROF benchamrk \cite{rieger2020irof} to measure the accuracy of the feature attribution by pixel relevancy. Hereby we set all but the $k, k-1, ..., 2, 1$ most important pixels to a baseline and calculate the new network's probability estimate for the modified images wrt. to the target label. The new probabilities are divided by the probability of the original image in order to normalize them and calculate the average AOC. Higher pixel relevancy values indicate a better attribution. For images of higher resolution we remove the $k$ most important pixels in steps of $g$ (e.g. $40$) to reduce computational cost. 

% \textcolor{red}{The table size should be modified.}

\begin{table*}
\begin{tabularx}{\textwidth}{l @{\extracolsep{\fill}} cccccc}
\toprule

\textbf{Method} & Insertion (IAUC) & Deletion (DAUC) & IROF \cite{rieger2020irof} \\
\midrule
Gradient SHAP \cite{lundberg2017unified} & 0.61 $\pm\,$ 0.42 & 0.22 $\pm\,$ 0.29   & 0.73 $\pm\,$ 0.24 \\
DeepLIFT \cite{shrikumar2017learning} & 0.62 $\pm\,$ 0.42 & 0.23 $\pm\,$ 0.30      & 0.73 $\pm\,$ 0.23 \\
LIME \cite{ribeiro2016should} & \textbf{0.80 $\pm\,$ 0.31} & 0.23 $\pm\,$ 0.23     & \textbf{0.76 $\pm\,$ 0.22} \\
Saliency map \cite{simonyan2014} & 0.50 $\pm\,$ 0.35 & 0.37 $\pm\,$ 0.32           & 0.65 $\pm\,$ 0.25 \\
SmoothGrad \cite{smilkov2017smoothgrad} & 0.60 $\pm\,$ 0.26 & 0.38 $\pm\,$ 0.29    & 0.63 $\pm\,$ 0.26 \\
Integrated Gradients  \cite{SundararajanTY17} & 0.66 $\pm\,$ 0.42 & \textbf{0.19 $\pm\,$ 0.27}      & 0.75 $\pm\,$ 0.23 \\
Guided Backpropagation \cite{springenberg2015striving} & 0.54 $\pm\,$ 0.38 & 0.49 $\pm\,$ 0.36      & 0.65 $\pm\,$ 0.25 \\
Original Image & 0.52 $\pm\,$ 0.32 & 0.53 $\pm\,$ 0.34                            & 0.47 $\pm\,$ 0.30 \\
\midrule
Mean Ensemble & \textbf{0.79 $\pm\,$ 0.33} & 0.25 $\pm\,$ 0.28                               & 0.70 $\pm\,$ 0.26 \\
Variance Ensemble \cite{rieger2020aggregating} & 0.62 $\pm\,$ 0.36 & 0.39 $\pm\,$ 0.31       & 0.71 $\pm\,$ 0.26 \\
RBM ensemble \scriptsize{with the flip detection} & 0.76 $\pm\,$ 0.38 & 0.19 $\pm\,$ 0.26    & 0.76 $\pm\,$ 0.22 \\
RBM ensemble \scriptsize{with the metric optimization} & 0.77 $\pm\,$ 0.37 & \textbf{0.18 $\pm\,$ 0.24}      & \textbf{0.76 $\pm\,$ 0.22} \\
\bottomrule
\end{tabularx}
\caption{A quantitative comparison between single and ensemble methods for the pixel perturbation: IAUC (higher is better), DAUC (lower is better), and IROF (higher is better) experiments on 10,000 samples from the CIFAR10 validation dataset \cite{krizhevsky2009learning}.}
\label{tab:quant_experiments}

\end{table*}

\subsection{An RBM Ensemble Within a Singe Explanation Framework}

\begin{figure}
    \centering
    \includegraphics[width=\textwidth]{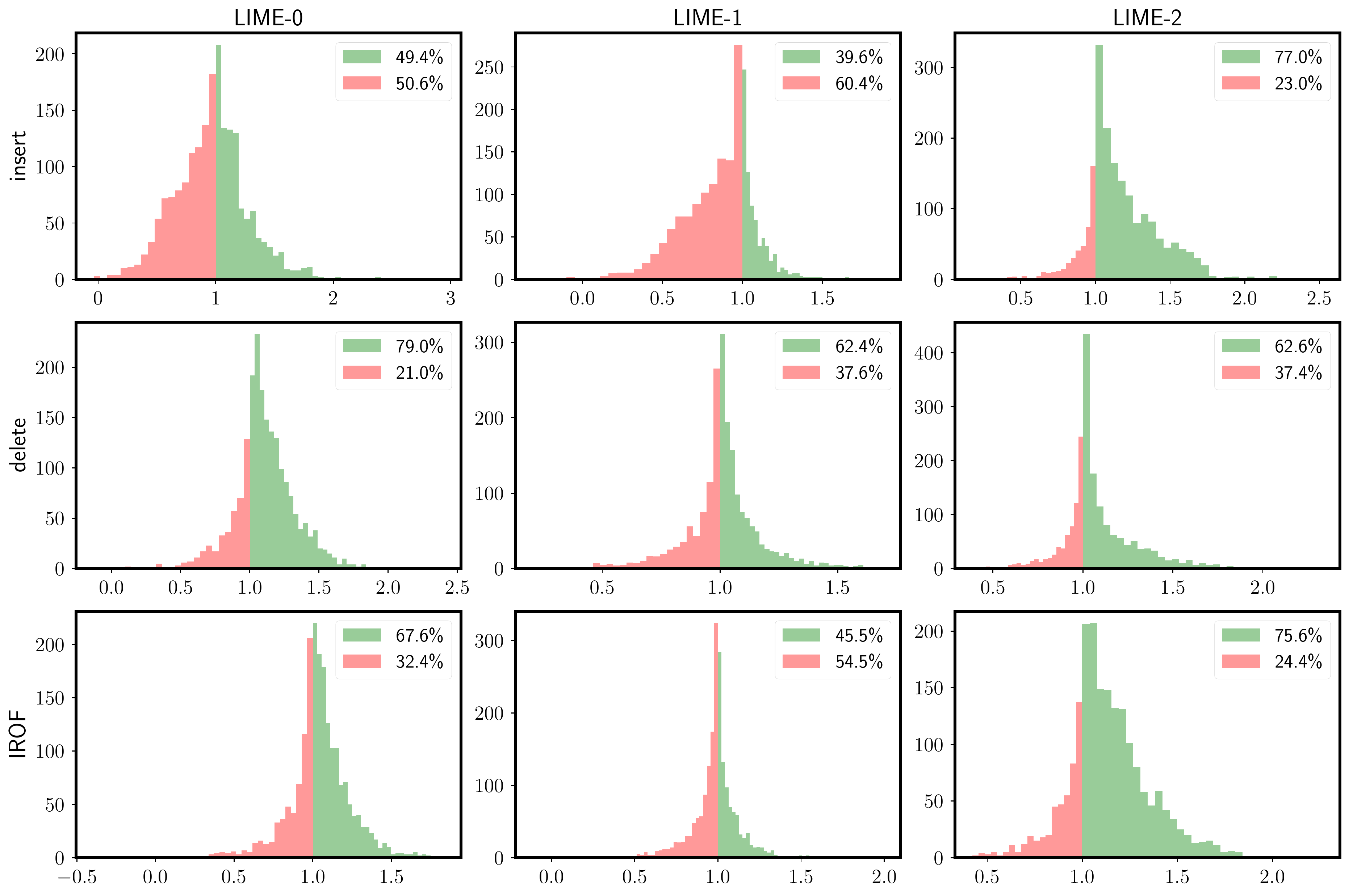}
    \caption{Distributions of differences where the proposed RBM ensemble shows better (green) and inferior (red) results in comparison to a baseline explanation technique according to insertion, deletion, and IROF metrics (score 1 means them being equal). The ensemble consists of the feature attributions from the same algorithm - LIME, but different hyperparameters. We randomly sampled $2000$ images from CIFAR10 \cite{krizhevsky2009learning} for this experiment.
    }
    \label{fig:lime_ensemble}
\end{figure}

In this experiment, we demonstrate that even for multiple baseline explanations of the same explanation method, the unsupervised ensemble with an RBM can lead to an improvement. To this end, we select the LIME \cite{ribeiro2016should} method with a different hyperparameter - the number of superpixels in the image. For the baselines (LIME-0, LIME-1, and LIME-2), we used $10$, $100$, and $1000$ superpixels per image, respectively. The results can be seen in Fig. \ref{fig:lime_ensemble}. The main idea is that each lime method has a different granularity level, thus highlighting distinct detail levels, and the proposed method's aggregation may help improve the reliability and robustness of feature attributions. 

\begin{figure}
    %\centering
    \hspace{0.8cm} Original  \hspace{1.5cm} LIME-0\hspace{1.4cm} LIME-1 \hspace{1.4cm} LIME-2 \hspace{0.95cm} RBM-ensemble\hfill

    \includegraphics[width=\textwidth]{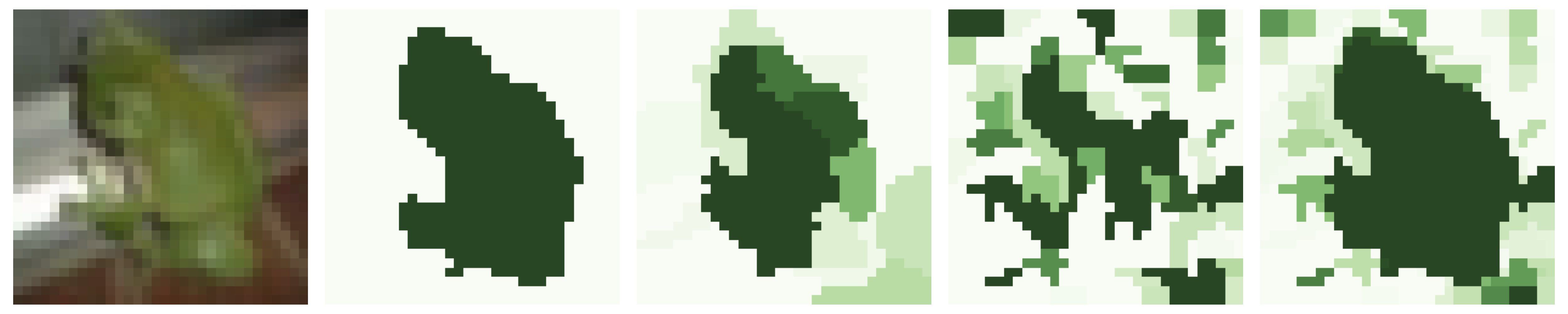}
    \textbf{\# of superpixels:}  \hspace{1.4cm} \textbf{10}\hspace{2.23cm} \textbf{50}\hspace{2.23cm} \textbf{100} \hfill
    \caption{Three local feature attribution maps for a single data sample from CIFAR10 dataset \cite{krizhevsky2009learning} using the LIME algorithm \cite{ribeiro2016should} with different number of superpixels and the proposed RBM ensemble of the selected feature attribution maps.}
    \label{fig:lime_superpixels}
\end{figure}

% Using multiple LIME attribution methods with different hyperparameters as base learners for the RBM ensemble. The scores shown are the differences between RBM and LIME base learners. Where 1. means them being equal. In the legend the percentage how often the RBM is better or worse is shown. In majority of tests ... 

\subsection{Reproducibility}

For reproducibility reasons, we describe the data preprocessing step used for all the experiments and provide information about packages used in this work. Also, the code for every experiment is publicly available online (see the links provided in Section \ref{sec:intro}). 

To achieve a fair comparison, the image data from all datasets was preprocessed in the same way for each baseline. We performed a per saliency map normalization before the aggregation.
In every experiment, we used the ResNet18 neural network architecture \cite{he2016deep}, except for the experiment on the MNIST dataset where we utilized a simple five layers convolutional neural network.
We use a pre-trained model for ImageNet dataset \cite{deng2009imagenet} from torchvision library \cite{PyTorch}.  

For the experiments we used the Bernoulli RBM implementation from the Scikit-Learn library \cite{scikit-learn} with following hyperparameters for each experiment: for the MNIST dataset we set the batch size to $5$, the learning rate to $0.01$, and the number of iterations to $100$. For CIFAR10 and ImageNet datasets we use the following hyperparameters: a batch size of $35$, a learning rate of $0.001$, and a number of iterations is $250$. The rest of hyperparameters are default to the scikit-learn package. For all baseline explanation techniques we use the publicly available open-source implementations from the captum library \cite{kokhlikyan2020captum} with their default hyperparameters.

% \subsection{Evaluation methods} 

\section{Discussion and Future Work}
\label{sec:discussion}

The results of multiple experiments with the proposed RBM ensemble show its competitive performance compared to base explanation techniques and other ensemble approaches.  We hypothesize moderate performance of the RBM ensemble on the insertion (IAUC) benchmark is connected to our data preparation step since we filter the negative values for every saliency map in the ensemble.

The computational complexity of an ensemble method primarily depends on the base learner. In our case, the base explanation techniques are relatively fast, especially on specialized hardware (GPU or TPU), where an RBM has low computational complexity. 

The gradient-based methods frequently produce noisy explanations. We empirically demonstrated that our approach reduces the noise in the final ensemble (Tab. \ref{tab:visual_experiment}). Therefore, we believe that the RBM aggregation of multiple saliency maps from gradient-based feature attributions is a powerful tool for improving the overall reliability of local explanations.   

% We also believe that the competitive learning \cite{lobov2020competitive} might improve the final ensemble performance. Also, the other unsupervised approaches 

As part of our future work, we aim to evaluate our aggregation approach on larger datasets. Furthermore, methods for selecting a few quite reliable base explanations for aggregation might lead to efficient explanations ensembles for larger datasets.

Finally we expect that the proposed approach can be easily adapted to handle local explanations over structured tabular data, where the explanation of deep neural networks is an essential task for many crucial applications such as healthcare and finance \cite{borisov2021deep}. 
% In addition, several other unsupervised ensemble methods can be applied to this problem, such as the competitive learning \cite{lobov2020competitive}. 

%%%%%%%%%%%%%%%%%%%%%%%%%%%%%%%%%%%%%%%%%%%%%%%%%%%%%%%%%%%%
%%%%%%%%%%%%%%%%%%%%%%%%%%%%%%%%%%%%%%%%%%%%%%%%%%%%%%%%%%%%
%%%%%%%%%%%%%%%%%%%%%%%%%%%%%%%%%%%%%%%%%%%%%%%%%%%%%%%%%%%%
%%%%%%%%%%%%%%%%%%%%%%%%%%%%%%%%%%%%%%%%%%%%%%%%%%%%%%%%%%%%
%%%%%%%%%%%%%%%%%%%%%%%%%%%%%%%%%%%%%%%%%%%%%%%%%%%%%%%%%%%%

\section{Conclusion}
\label{sec:conclusion}

In this work, we presented a novel approach to unsupervised aggregation of feature-based explanations using Restricted Boltzmann Machines with the aim of reliably interpreting the influence of inputs on the output of deep neural networks. In addition to explanatory reasons, the latter is also essential for debugging and diagnostic purposes and serves the long-term acceptance of deep learning in real-world applications.

Using the proposed approach, we demonstrated through visual and quantitative experiments its ability to obtain more robust and reliable explanations than other existing ensemble methods. In a setting with noisy attribution maps in an ensemble, the proposed approach successfully selects only the valuable information, mitigating noise. 
Moreover, our work illuminates and mitigates the problem of possible contradictory results that may be obtained by different explanation and evaluation methods.
Finally, we note that our approach can also be used within a single interpretability framework to reduce the sensitivity of a feature-based explanatory approach to its hyperparameters.

% achieve competitive performance

%%%%%%%%%%%%%%%%%%%%%%%%%%%%%%%%%%%%%%%%%%%%%%%%%%%%%%%%%%%%
%%%%%%%%%%%%%%%%%%%%%%%%%%%%%%%%%%%%%%%%%%%%%%%%%%%%%%%%%%%%
%%%%%%%%%%%%%%%%%%%%%%%%%%%%%%%%%%%%%%%%%%%%%%%%%%%%%%%%%%%%
\bibliographystyle{unsrt}
\bibliography{refs.bib}
%%%%%%%%%%%%%%%%%%%%%%%%%%%%%%%%%%%%%%%%%%%%%%%%%%%%%%%%%%%%
% \newpage

% \textbf{\textit{At the end of your paper submission, please indicate whether you would like an extended version of the submission to be considered for publication in a journal special issue.}}\\
% Yes, we would like to be considered. 

% \section{Reproducibility}
% \label{app:reproducibility}

% For the reproducibility reasons, we provide information about programming framework used in this study as well as hyper-parameters. 

\newpage
\appendix

\section{Additional Experiments}

\begin{table*}
%  \caption{Visual results of MNIST and ImageNet(VGG-19) in the noise case}
% \label{vis_results}
\begin{tabularx}{\textwidth}{l @{\extracolsep{\fill}} c @{\extracolsep{\fill}} c@{\extracolsep{\fill}}c@{\extracolsep{\fill}} c@{\extracolsep{\fill}}c@{\extracolsep{\fill}}c@{\extracolsep{\fill}} c@{\extracolsep{\fill}}c@{\extracolsep{\fill}}}
\toprule
$\,\,$ Original & LIME \cite{ribeiro2016should} & GB \cite{springenberg2015striving} & IG \cite{SundararajanTY17} & GS  \cite{lundberg2017unified} & SG \cite{smilkov2017smoothgrad} & Mean & Variance & \textbf{RBM} \\
 & & & & & & ensemble & ensemble &  \textbf{ensemble} \\
\midrule

\multicolumn{9}{c}{\textbf{With} noisy feature attribution maps in the ensemble} \\

\includegraphics[width=14.5mm]{visual_results/imagenet_2_orig.png} & 
\includegraphics[width=14.5mm]{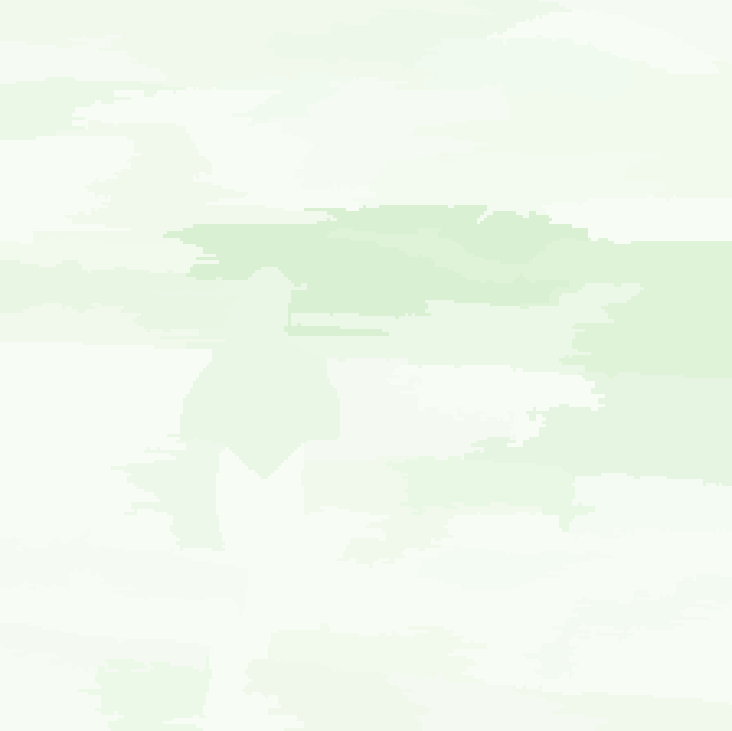} & 
\includegraphics[width=14.5mm]{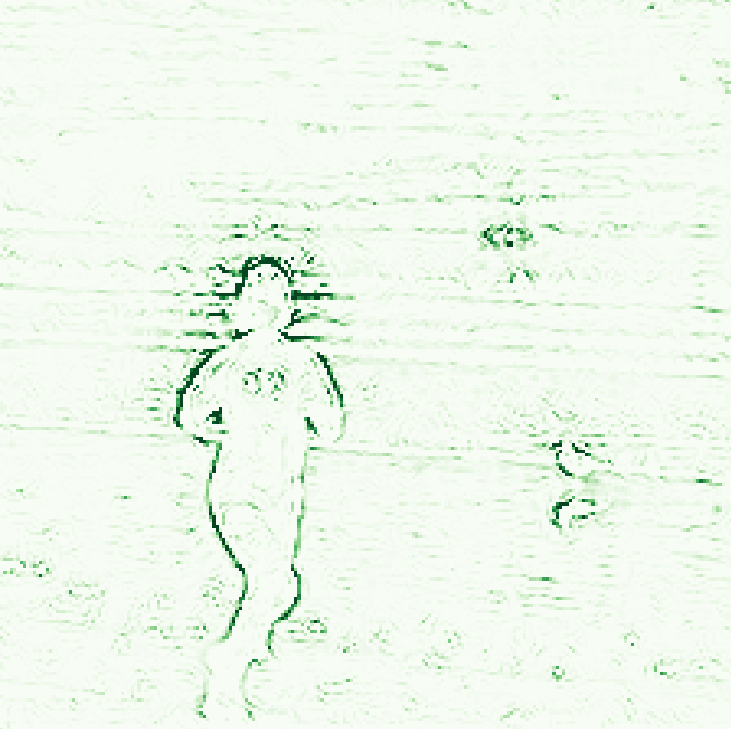} & 
\includegraphics[width=14.5mm]{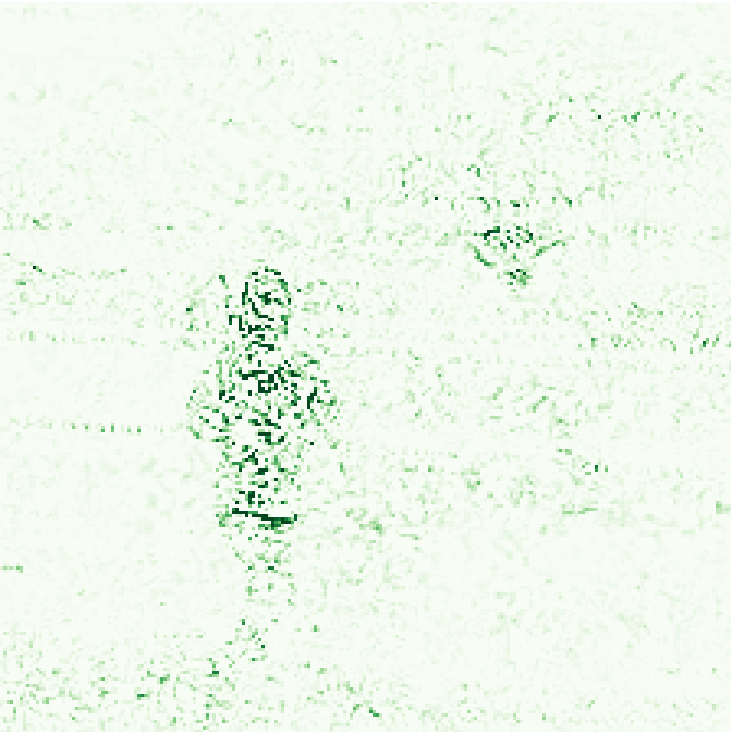} & 
\includegraphics[width=14.5mm]{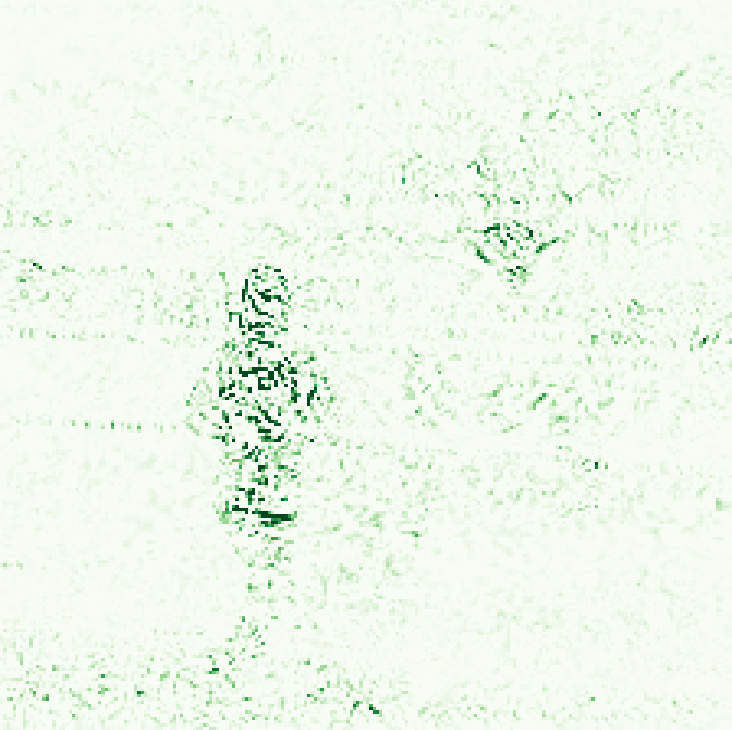} &
\includegraphics[width=14.5mm]{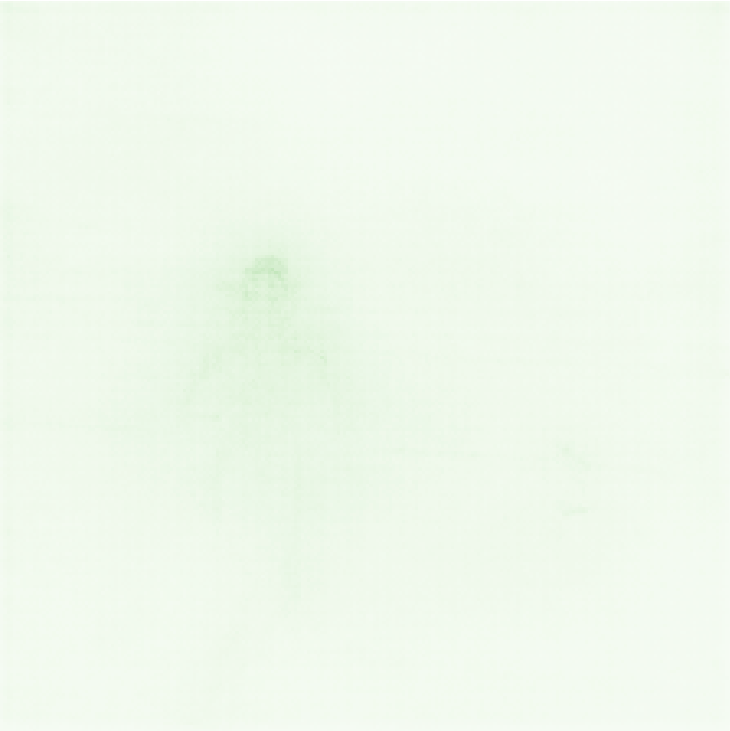} & 
\includegraphics[width=14.5mm]{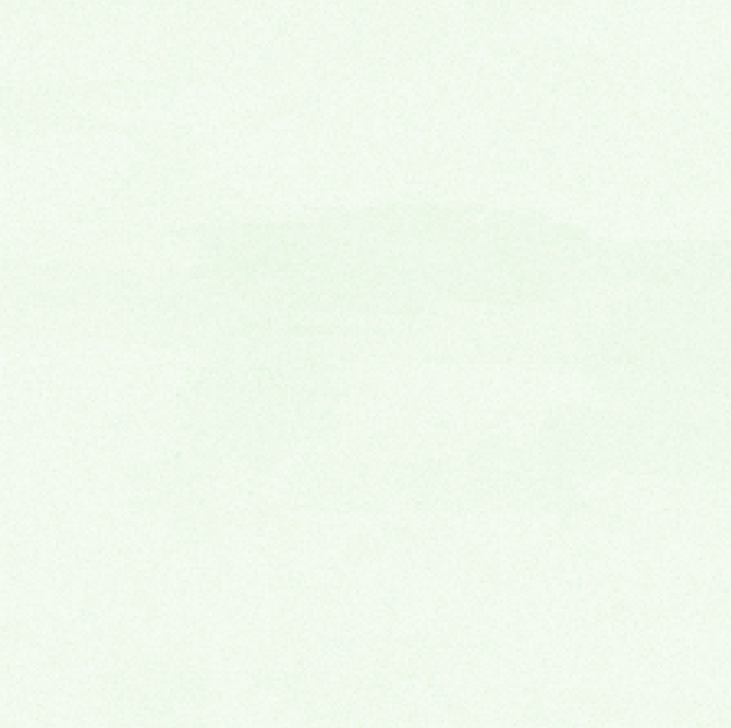} & 
\includegraphics[width=14.5mm]{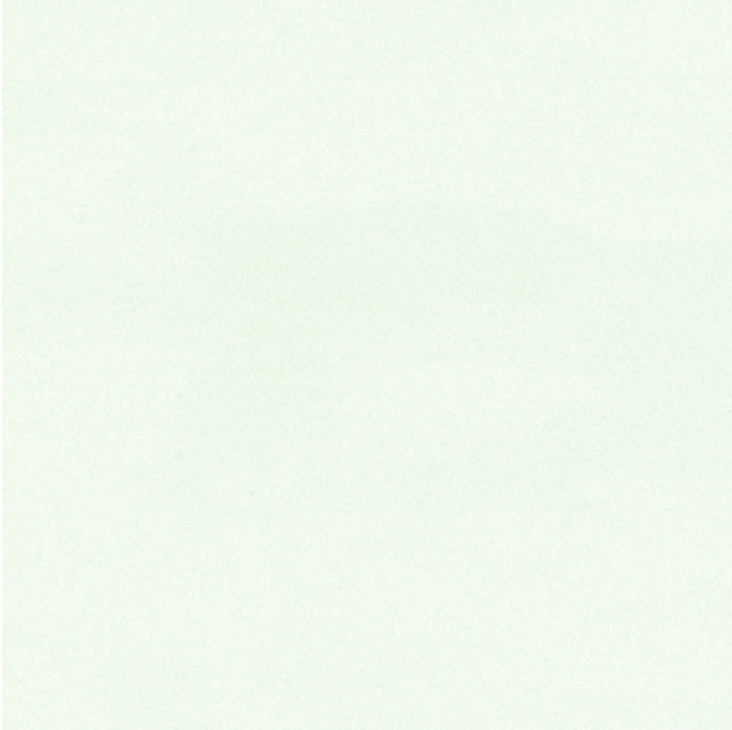} & 
\includegraphics[width=14.5mm]{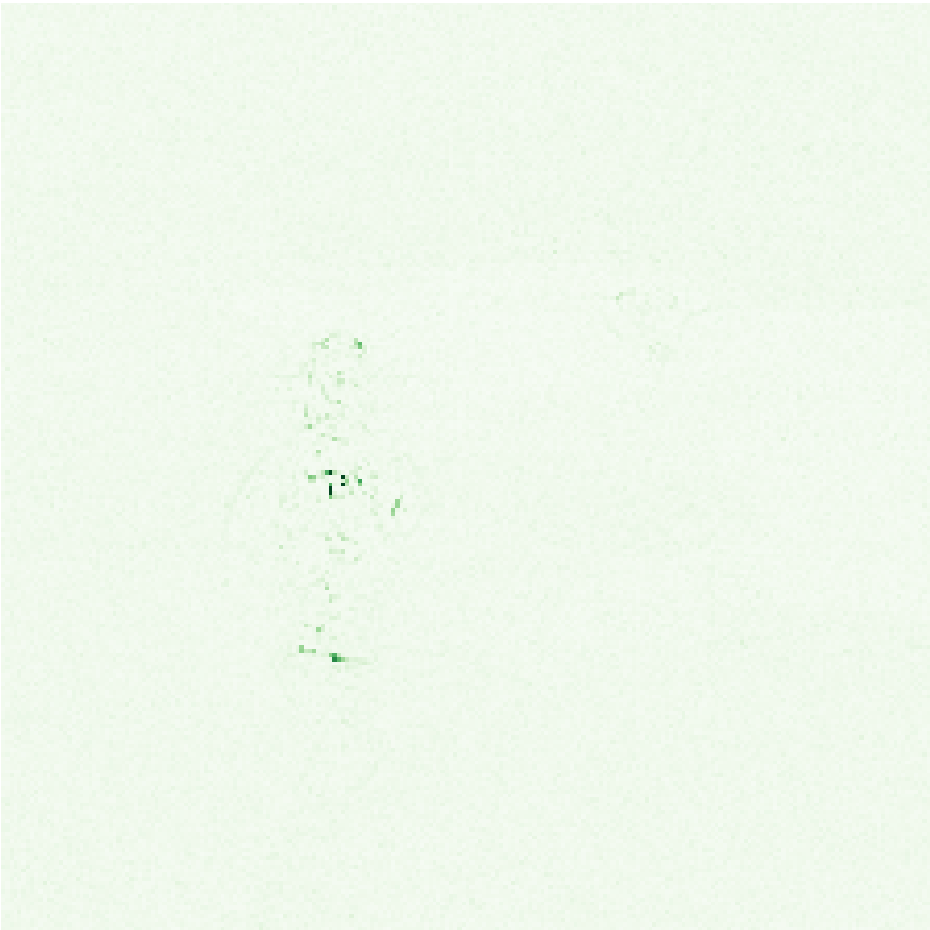} \\

\includegraphics[width=14.5mm]{visual_results/imagenet_3_orig.png} & 
\includegraphics[width=14.5mm]{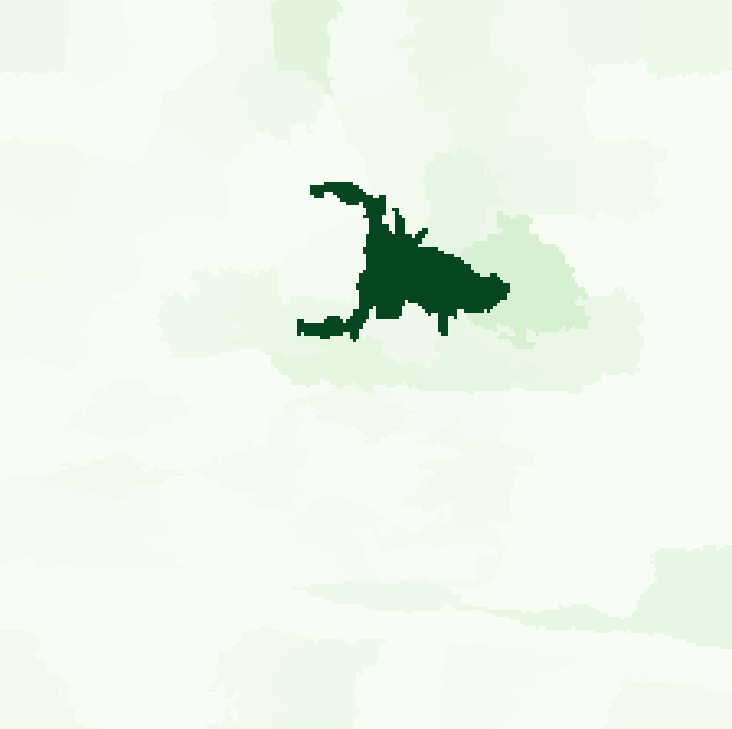} & 
\includegraphics[width=14.5mm]{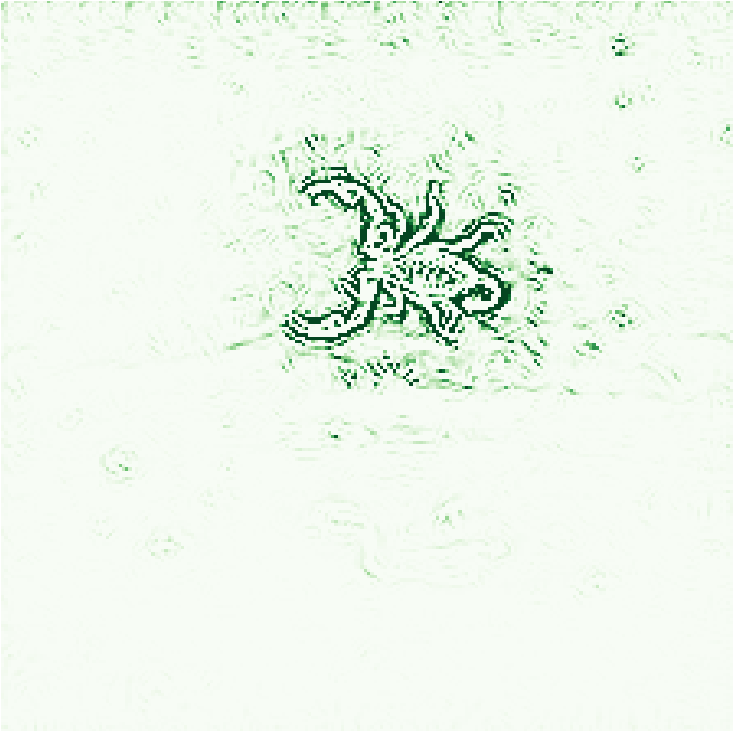} & 
\includegraphics[width=14.5mm]{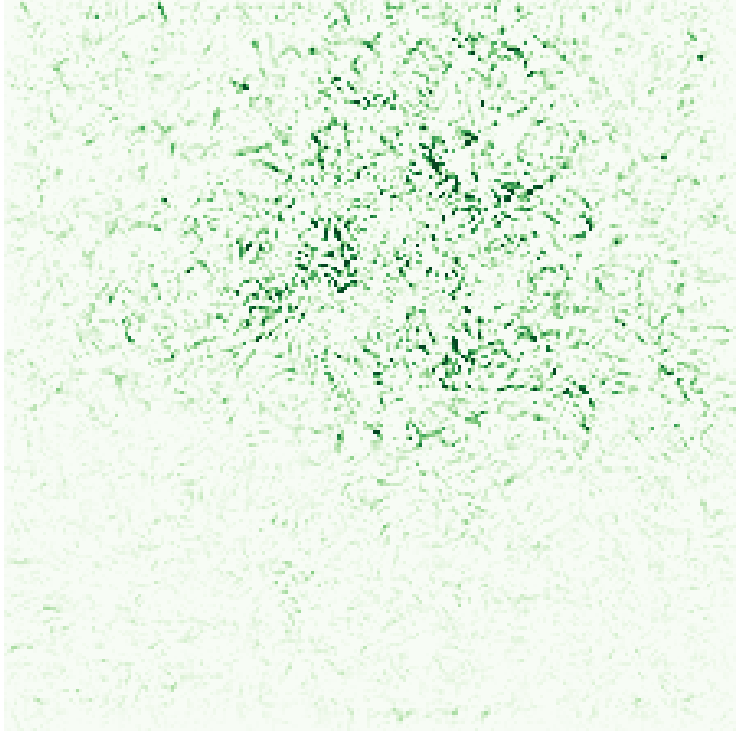} & 
\includegraphics[width=14.5mm]{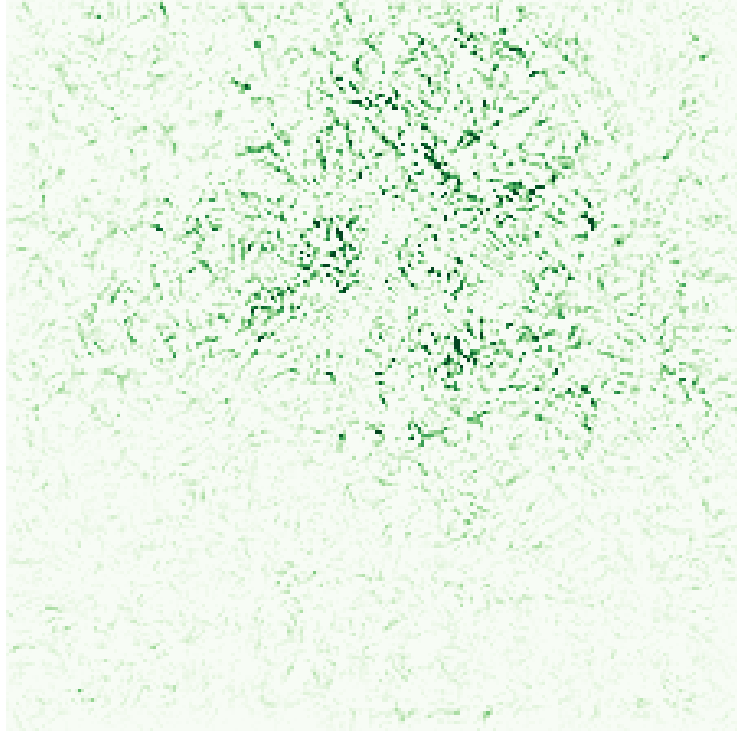} & 
\includegraphics[width=14.5mm]{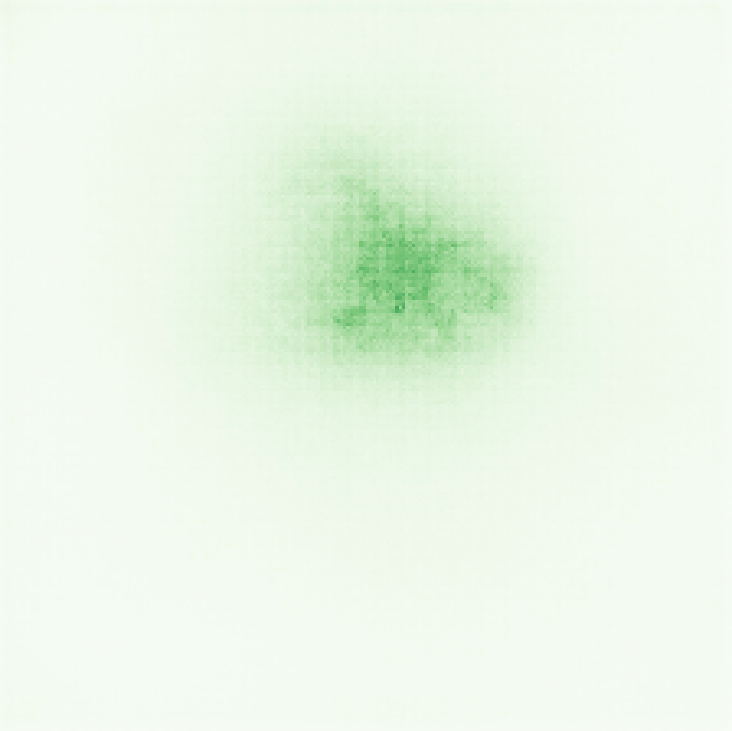} & 
\includegraphics[width=14.5mm]{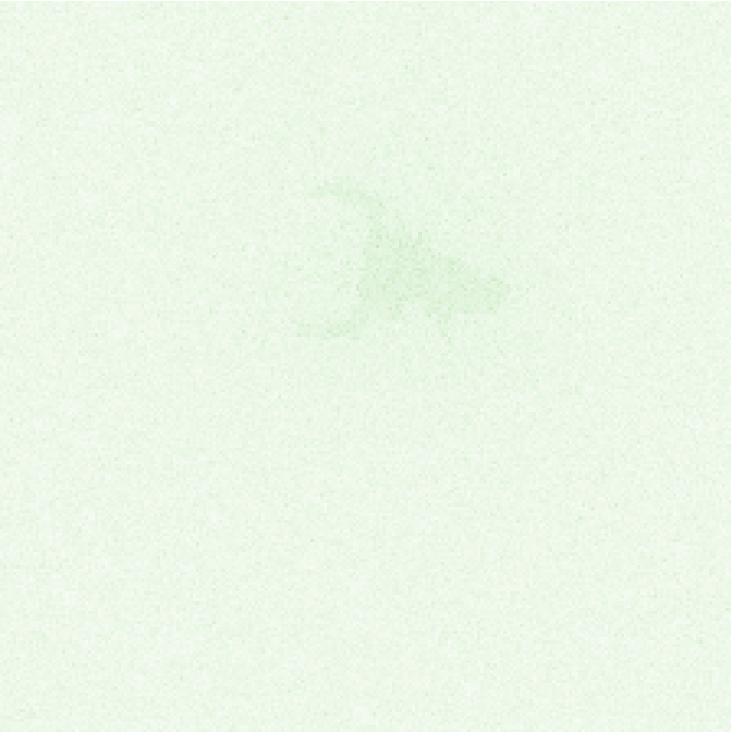} & 
\includegraphics[width=14.5mm]{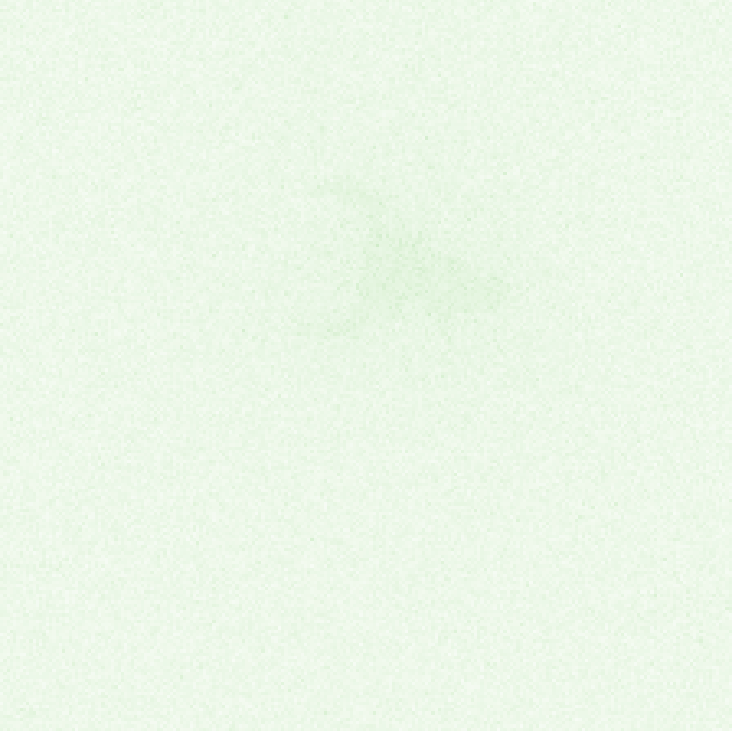} & 
\includegraphics[width=14.5mm]{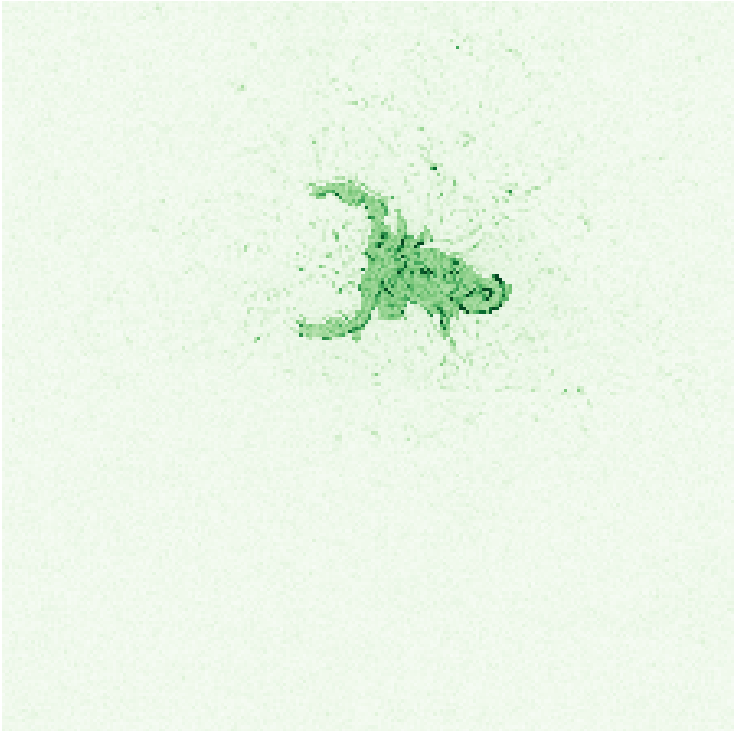} \\

\includegraphics[width=14.5mm]{visual_results/imagenet_5_orig.png} & 
\includegraphics[width=14.5mm]{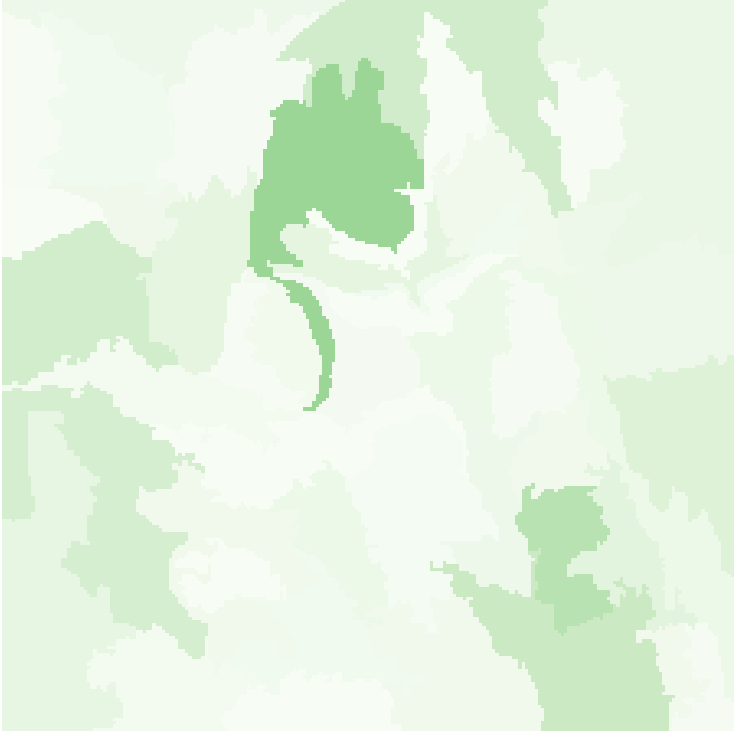} & 
\includegraphics[width=14.5mm]{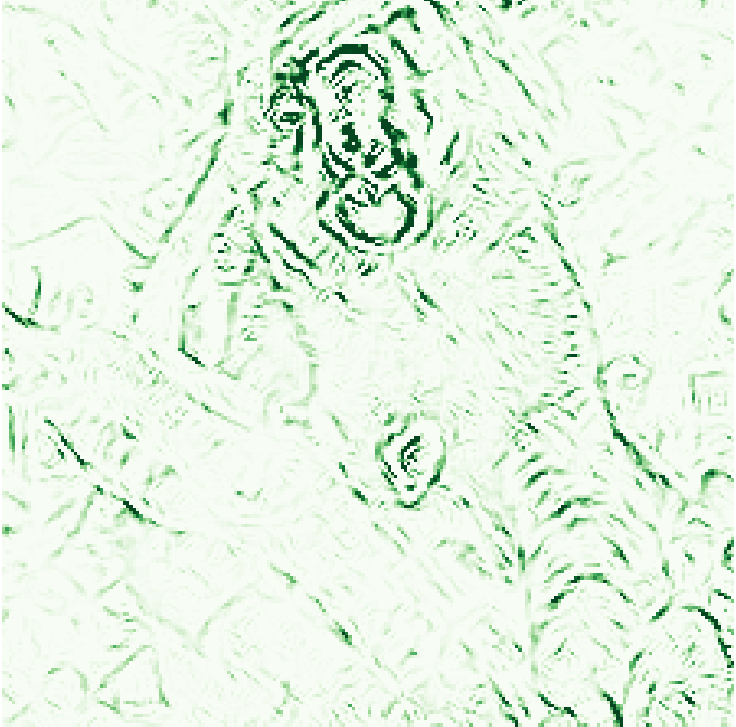} & 
\includegraphics[width=14.5mm]{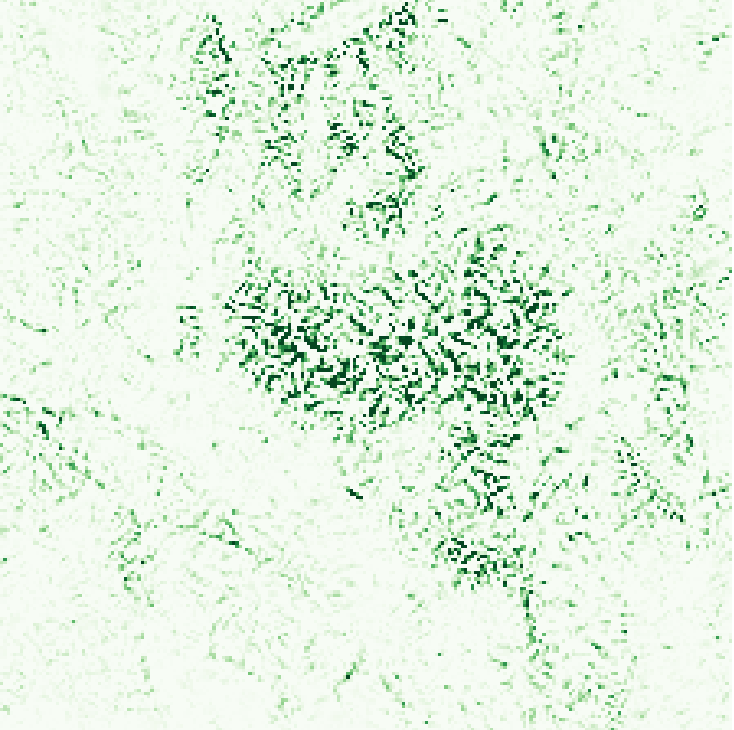} & 
\includegraphics[width=14.5mm]{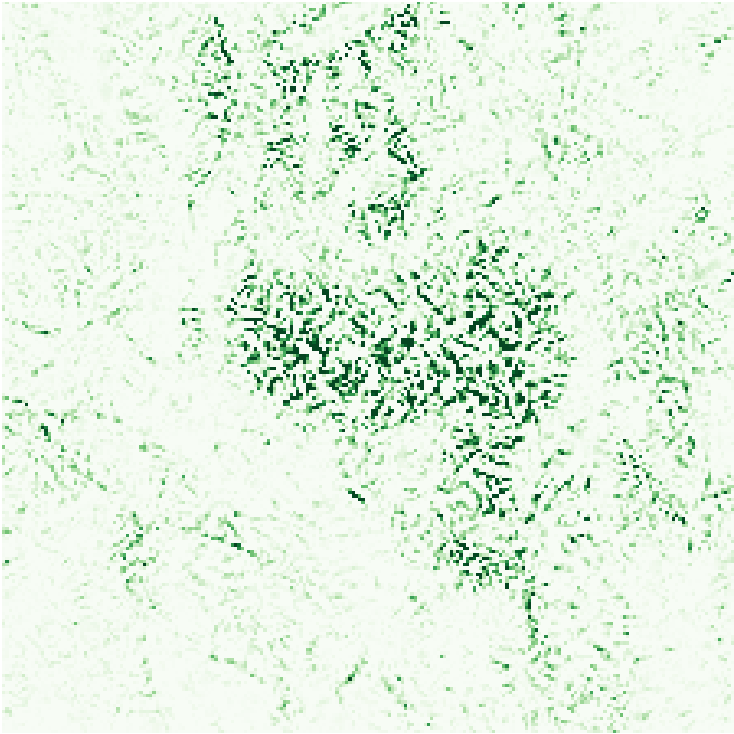} & 
\includegraphics[width=14.5mm]{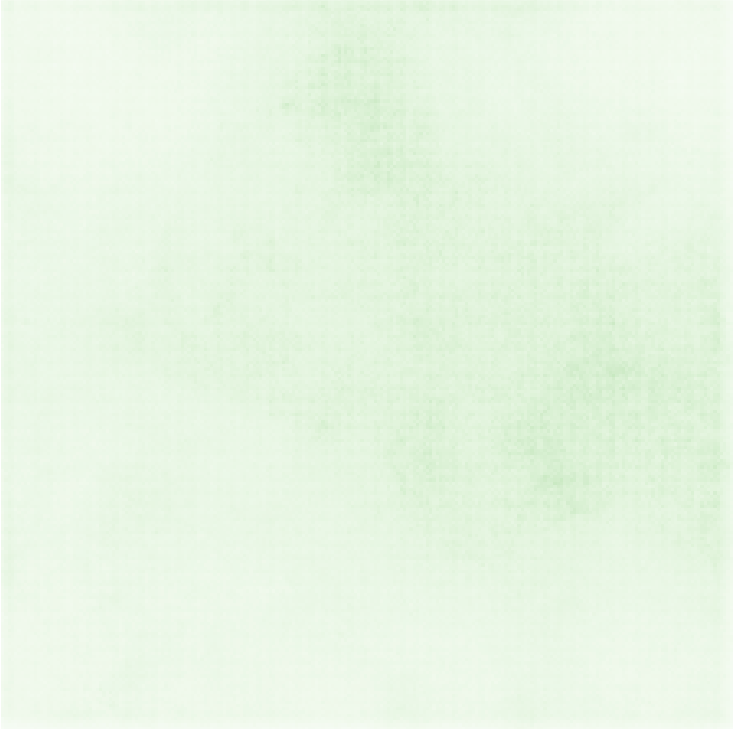} & 
\includegraphics[width=14.5mm]{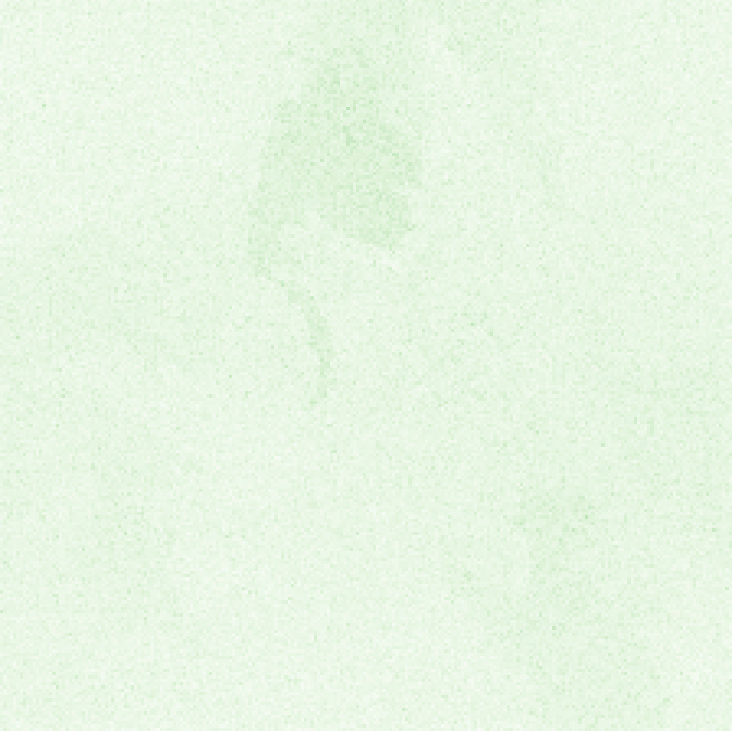} & 
\includegraphics[width=14.5mm]{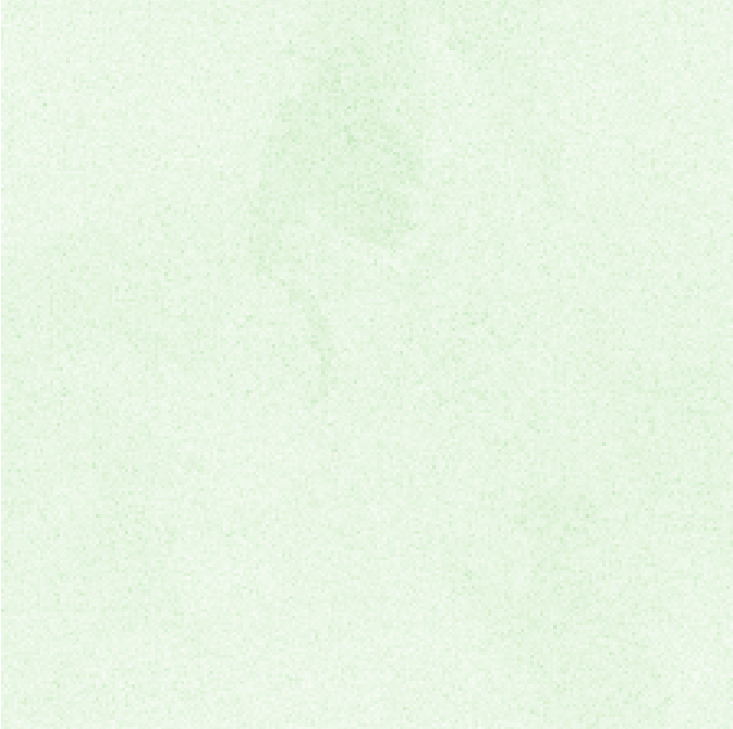} & 
\includegraphics[width=14.5mm]{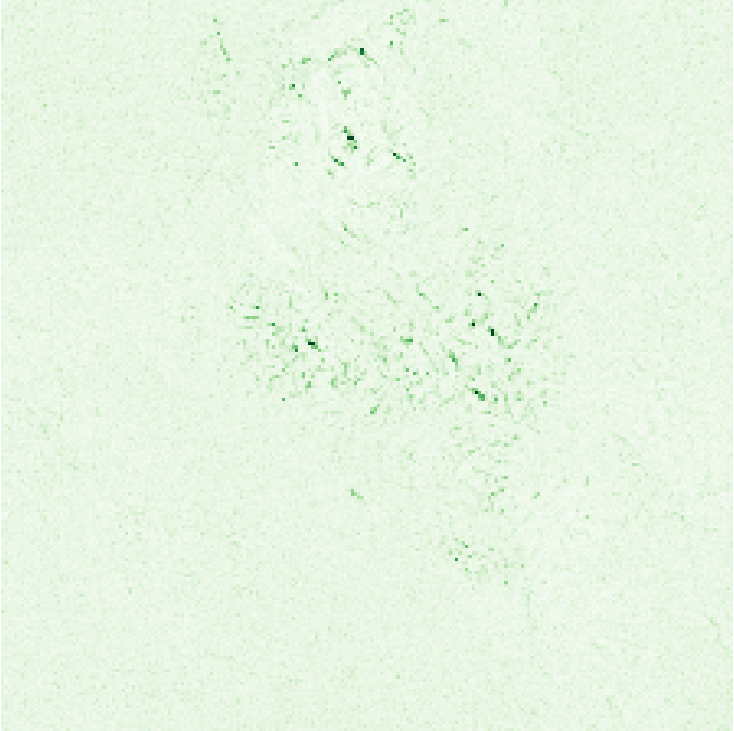} \\

\includegraphics[width=14.5mm]{visual_results/imagenet_4_orig.png} & 
\includegraphics[width=14.5mm]{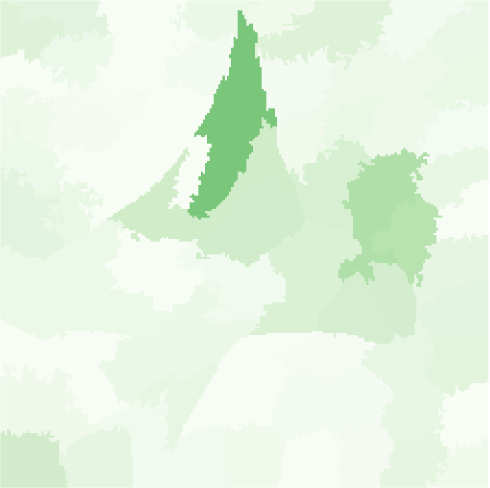} & 
\includegraphics[width=14.5mm]{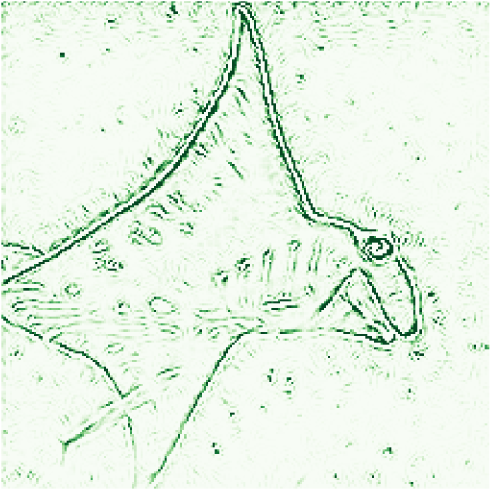} & 
\includegraphics[width=14.5mm]{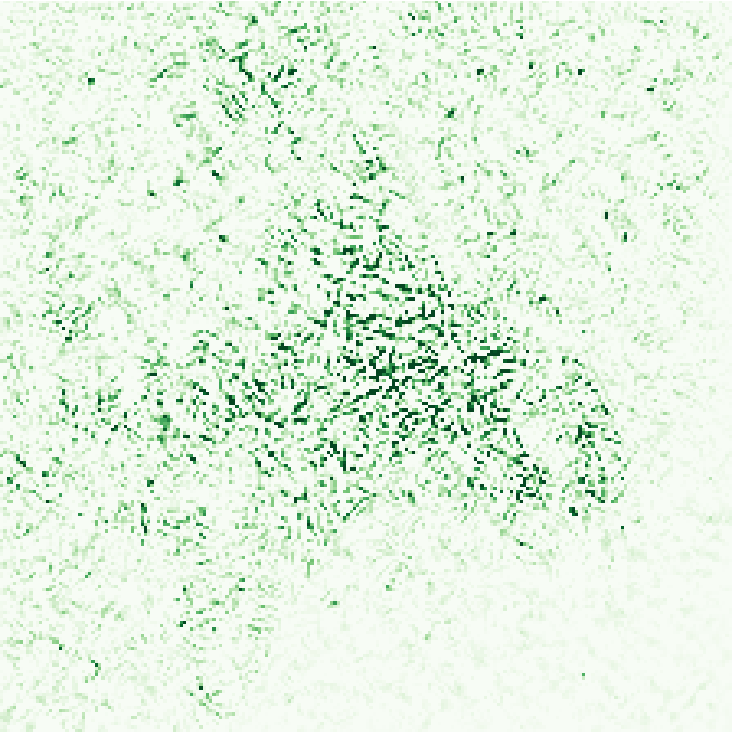} & 
\includegraphics[width=14.5mm]{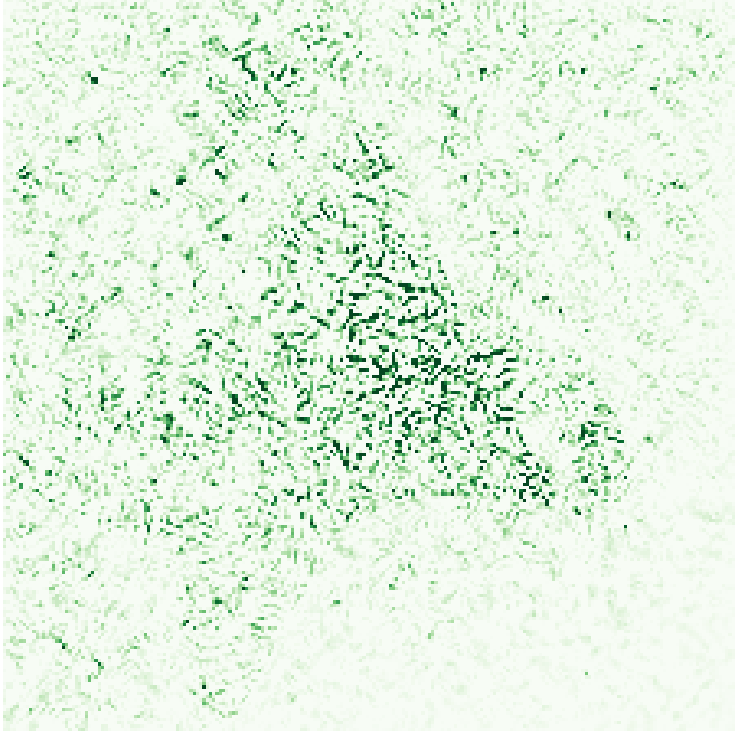} &\includegraphics[width=14.5mm]{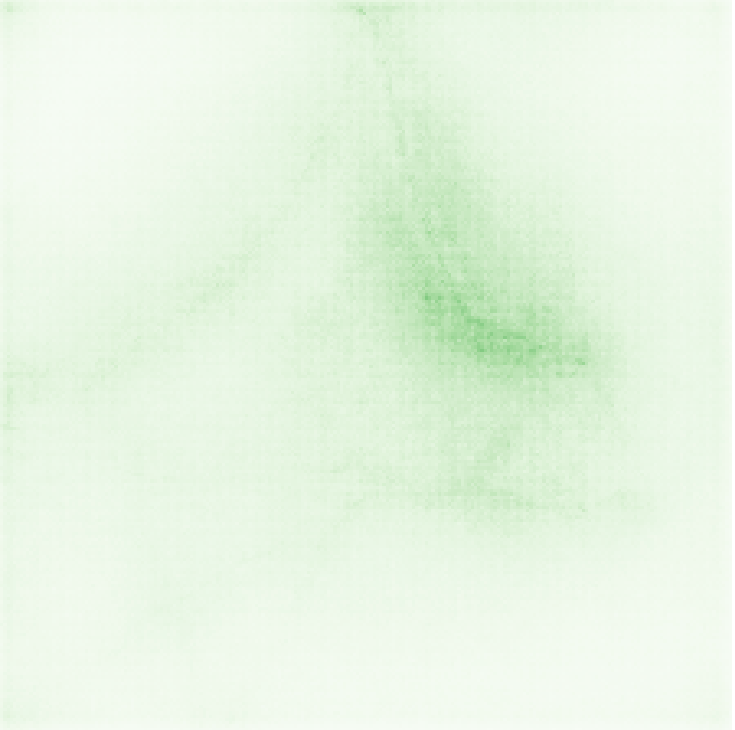} & 
\includegraphics[width=14.5mm]{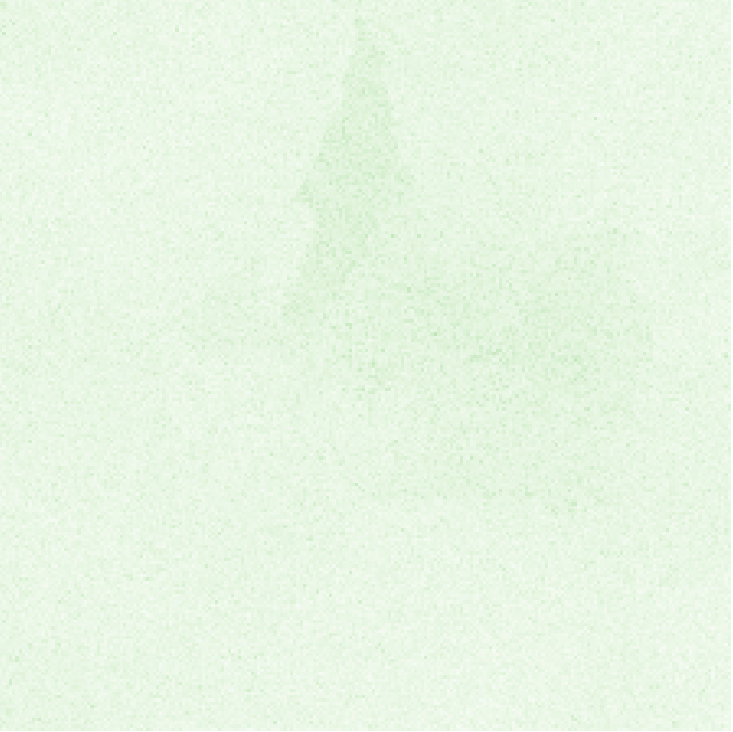} & 
\includegraphics[width=14.5mm]{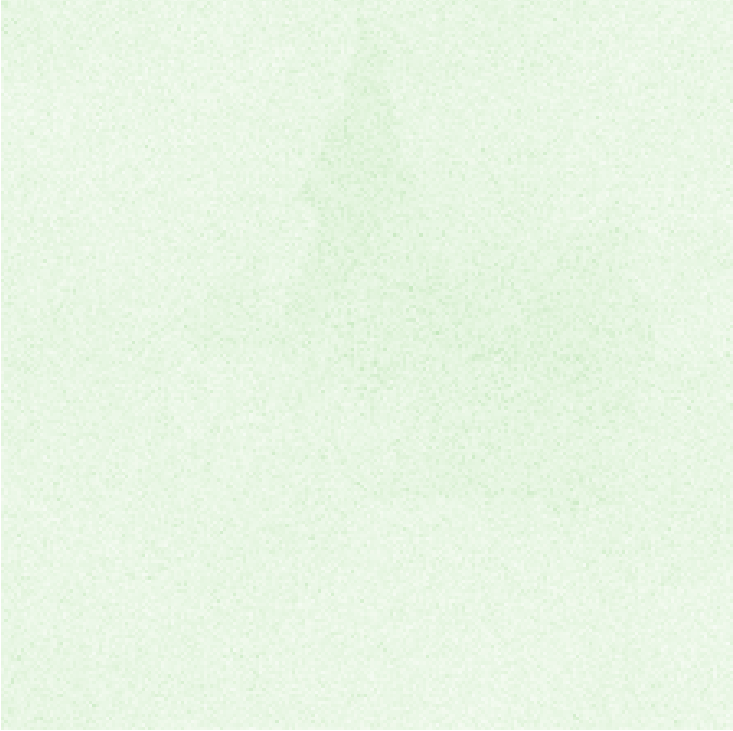} & 
\includegraphics[width=14.5mm]{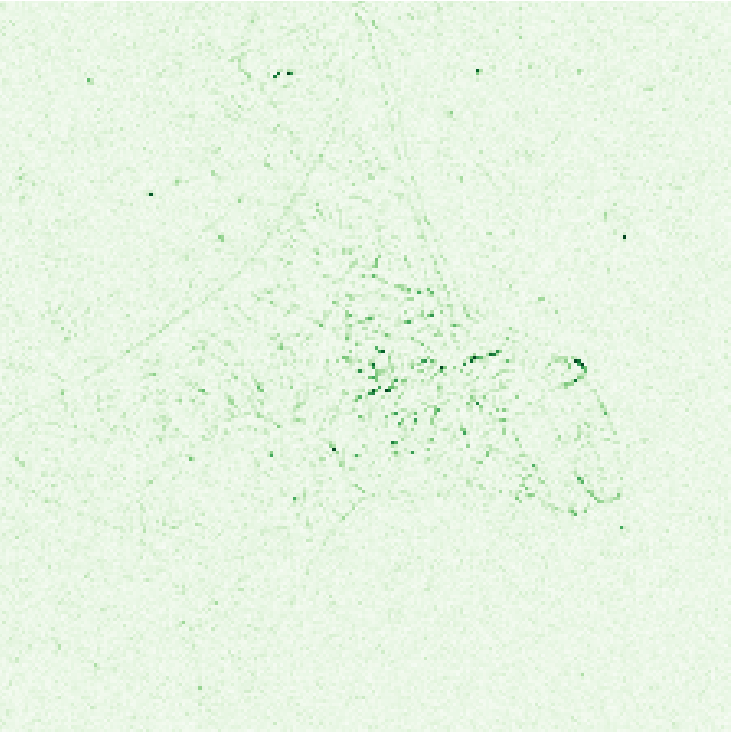} \\

\includegraphics[width=14.5mm]{visual_results/imagenet_1_orig.png} & 
\includegraphics[width=14.5mm]{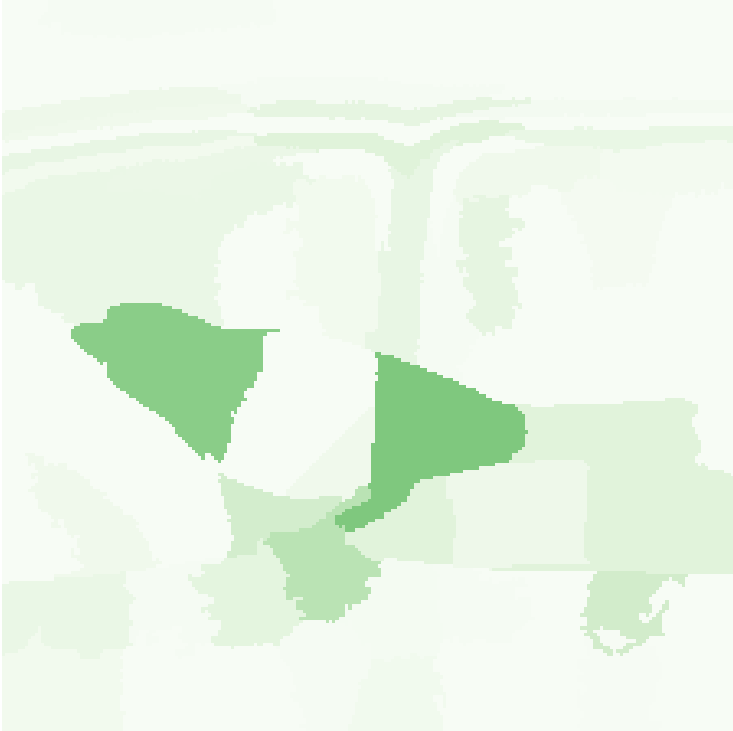} & 
\includegraphics[width=14.5mm]{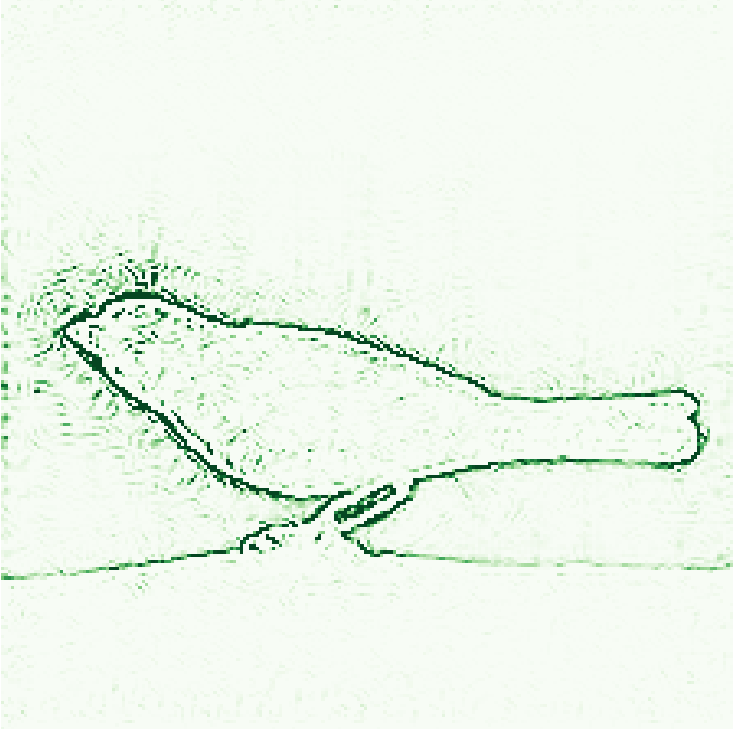} & 
\includegraphics[width=14.5mm]{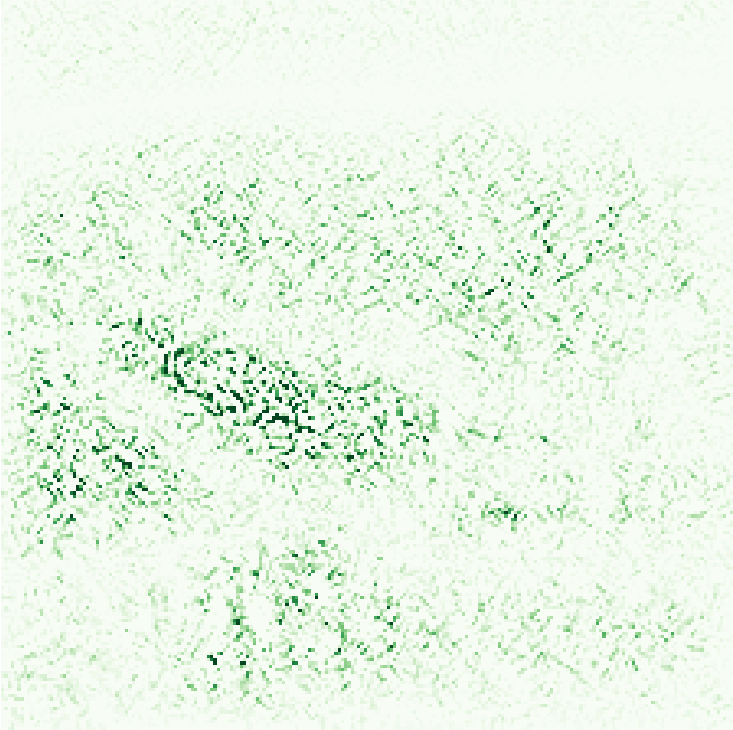} & 
\includegraphics[width=14.5mm]{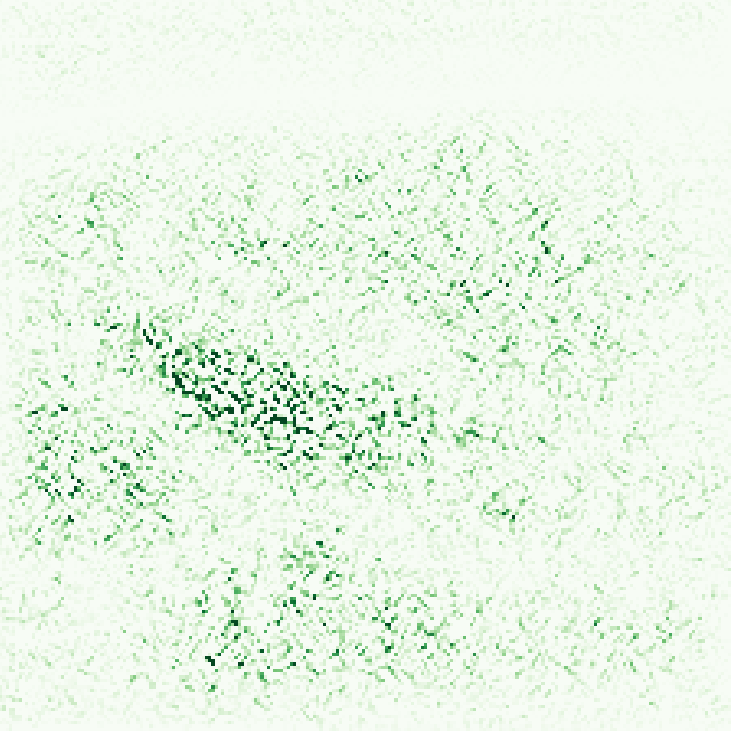} & 
\includegraphics[width=14.5mm]{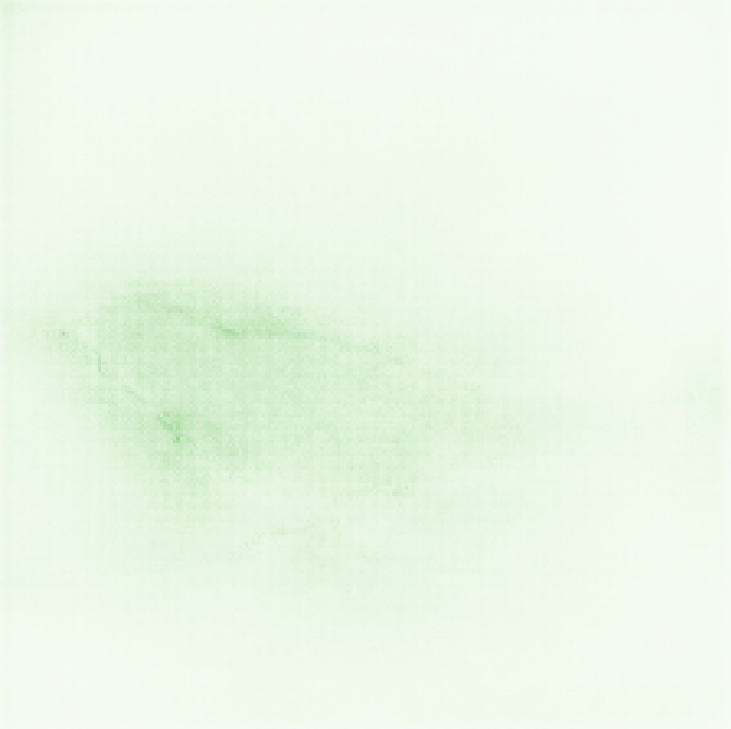} & 
\includegraphics[width=14.5mm]{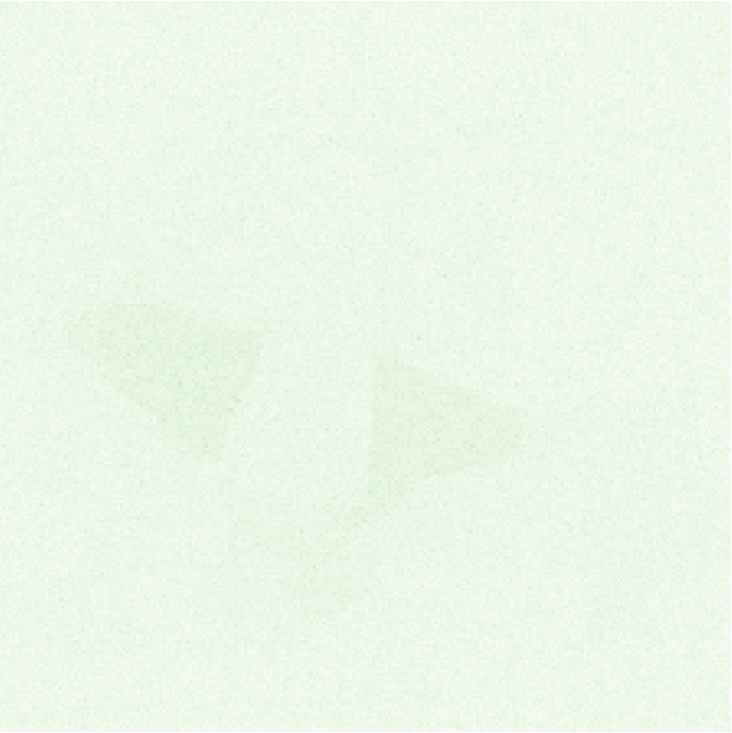} & 
\includegraphics[width=14.5mm]{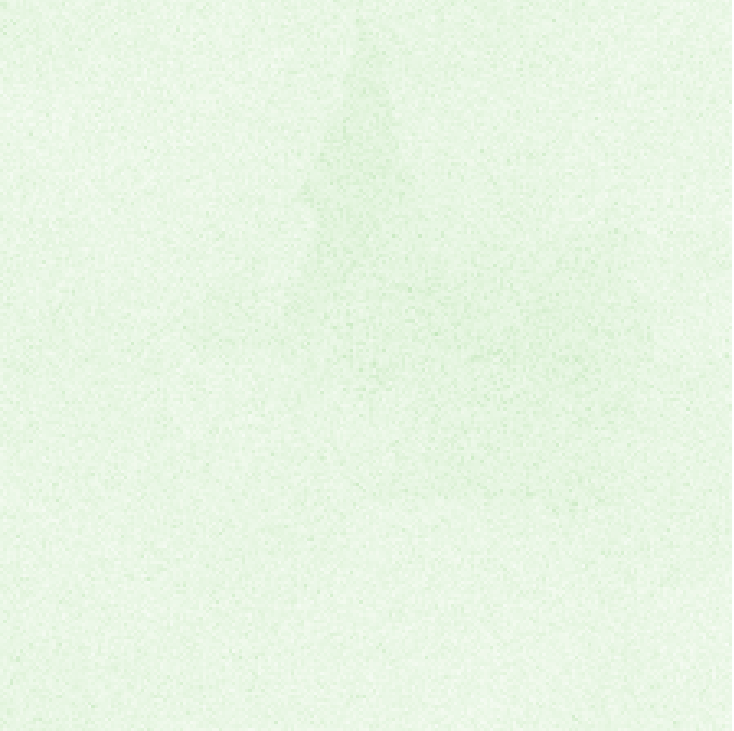} & 
\includegraphics[width=14.5mm]{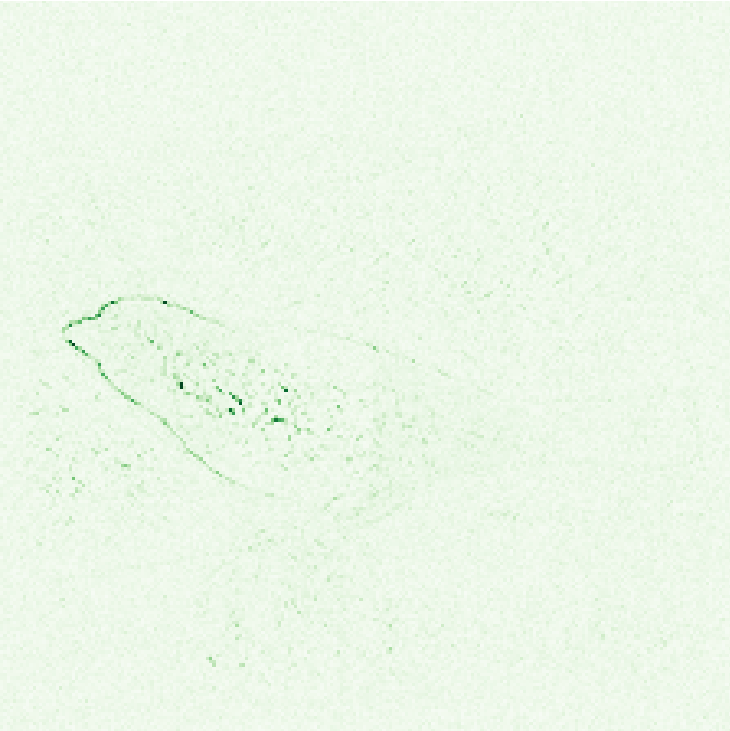} \\

\multicolumn{9}{c}{\textbf{Without} noisy feature attribution maps in the ensemble} \\ 

\includegraphics[width=14.5mm]{visual_results/mnist_3_orig.png} & 
\includegraphics[width=14.5mm]{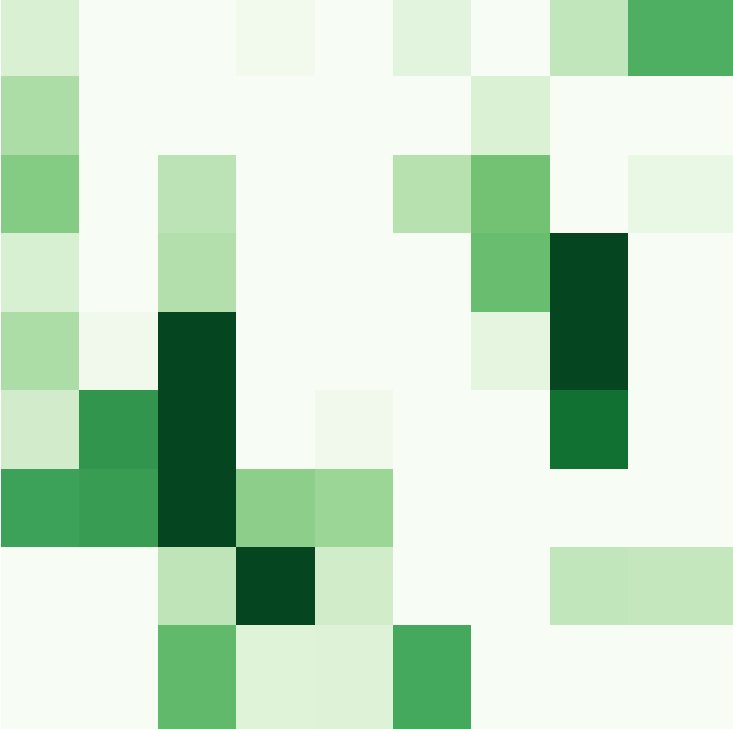} & 
\includegraphics[width=14.5mm]{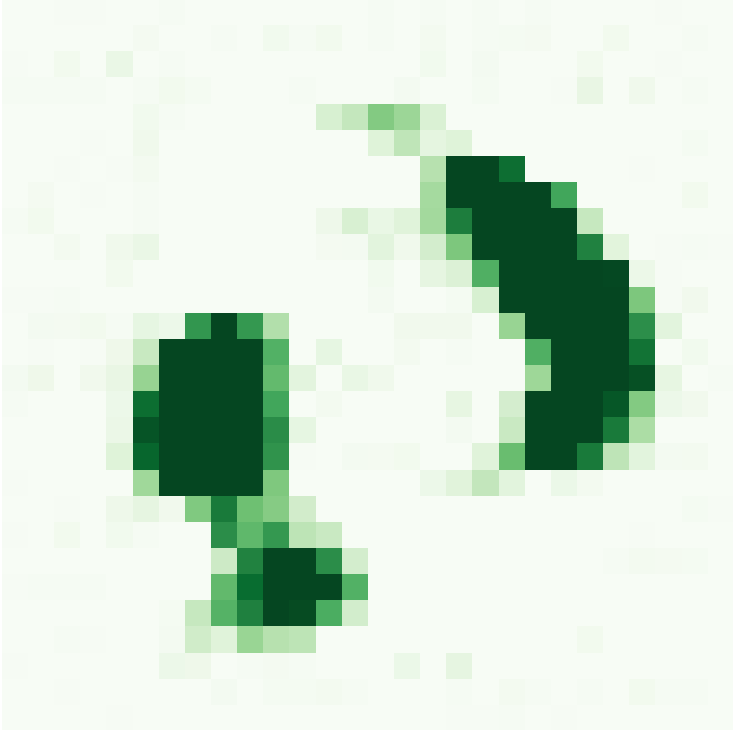} & 
\includegraphics[width=14.5mm]{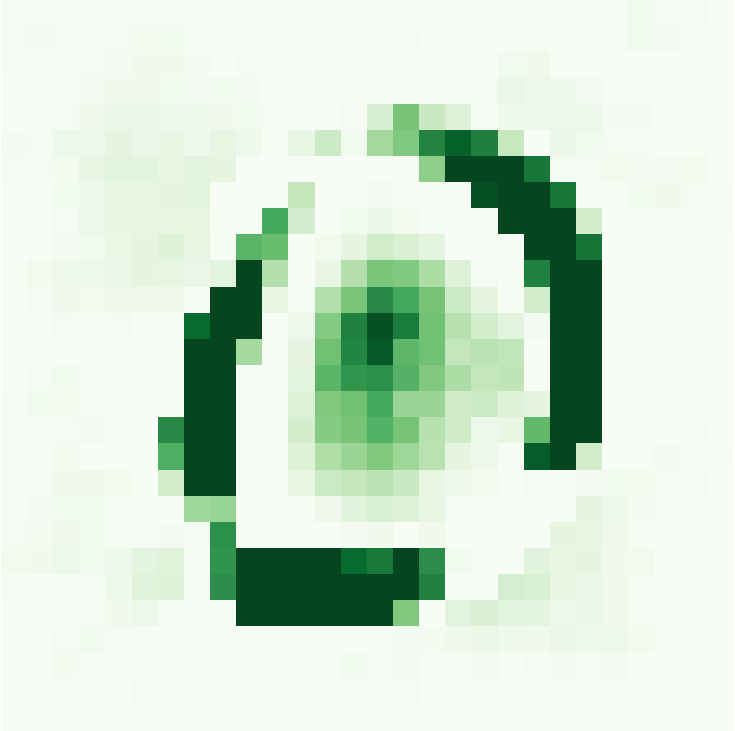} & 
\includegraphics[width=14.5mm]{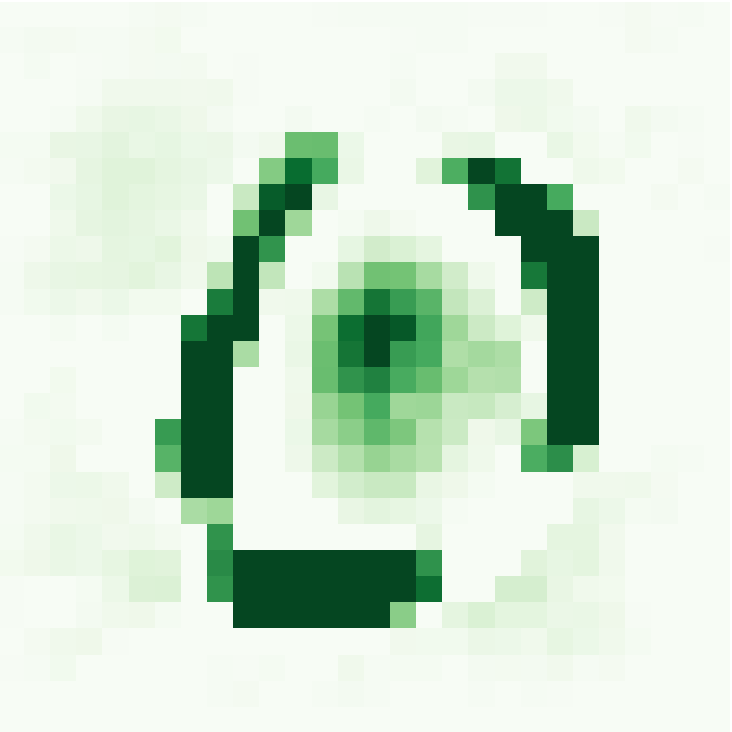} & 
\includegraphics[width=14.5mm]{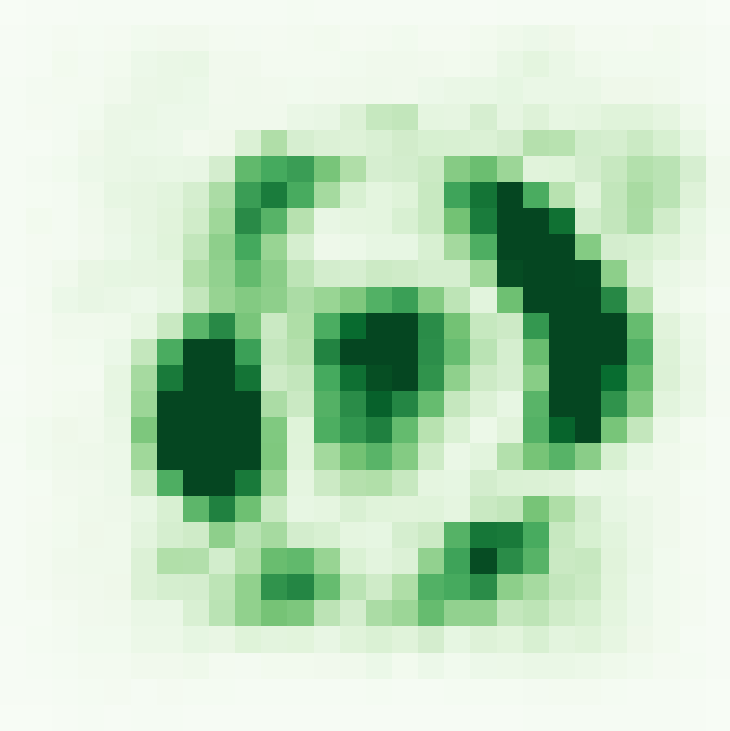} & 
\includegraphics[width=14.5mm]{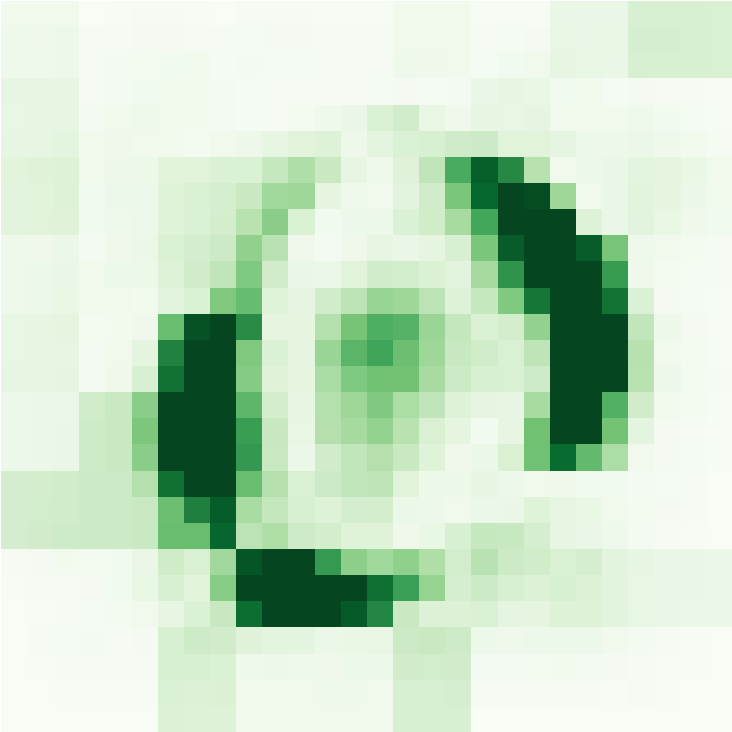} & 
\includegraphics[width=14.5mm]{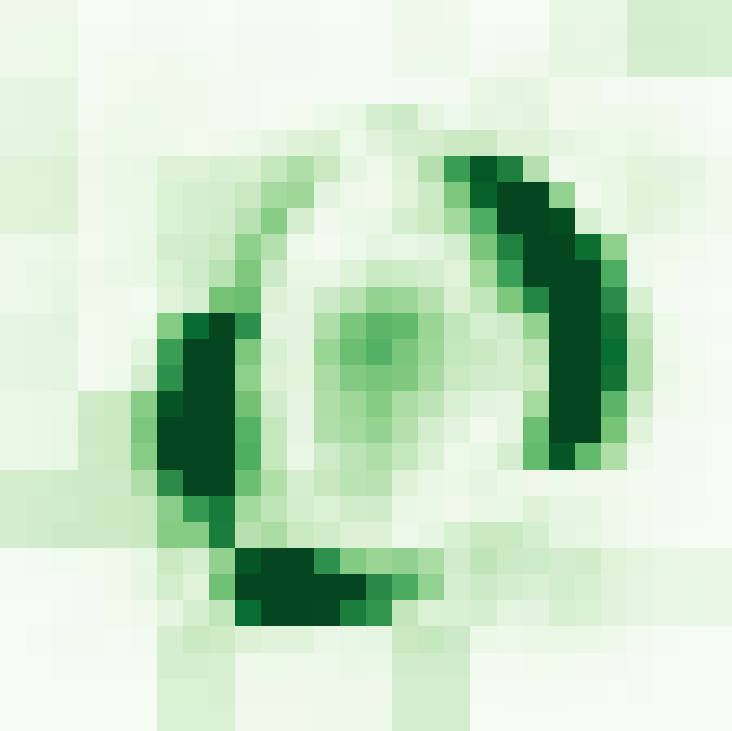} & 
\includegraphics[width=14.5mm]{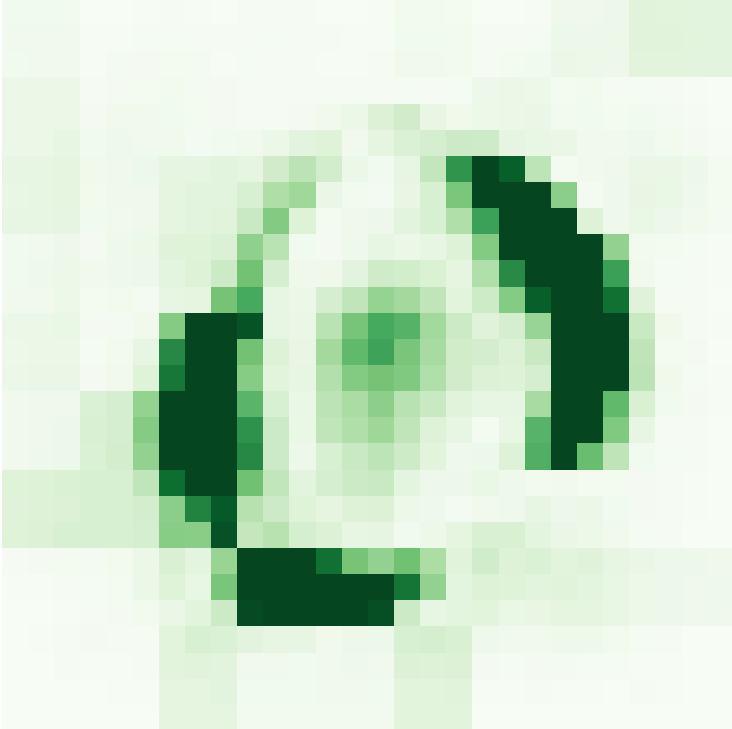} \\

\includegraphics[width=14.5mm]{visual_results/mnist_5_orig.png} & 
\includegraphics[width=14.5mm]{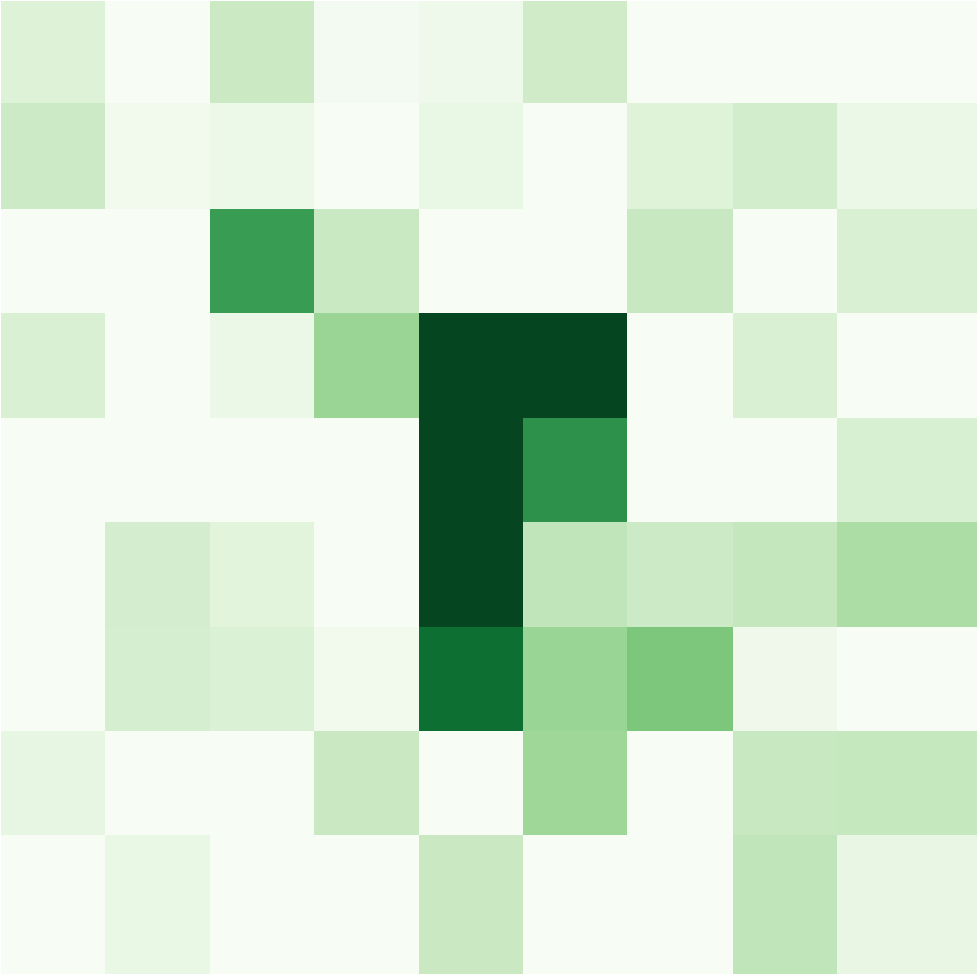} & 
\includegraphics[width=14.5mm]{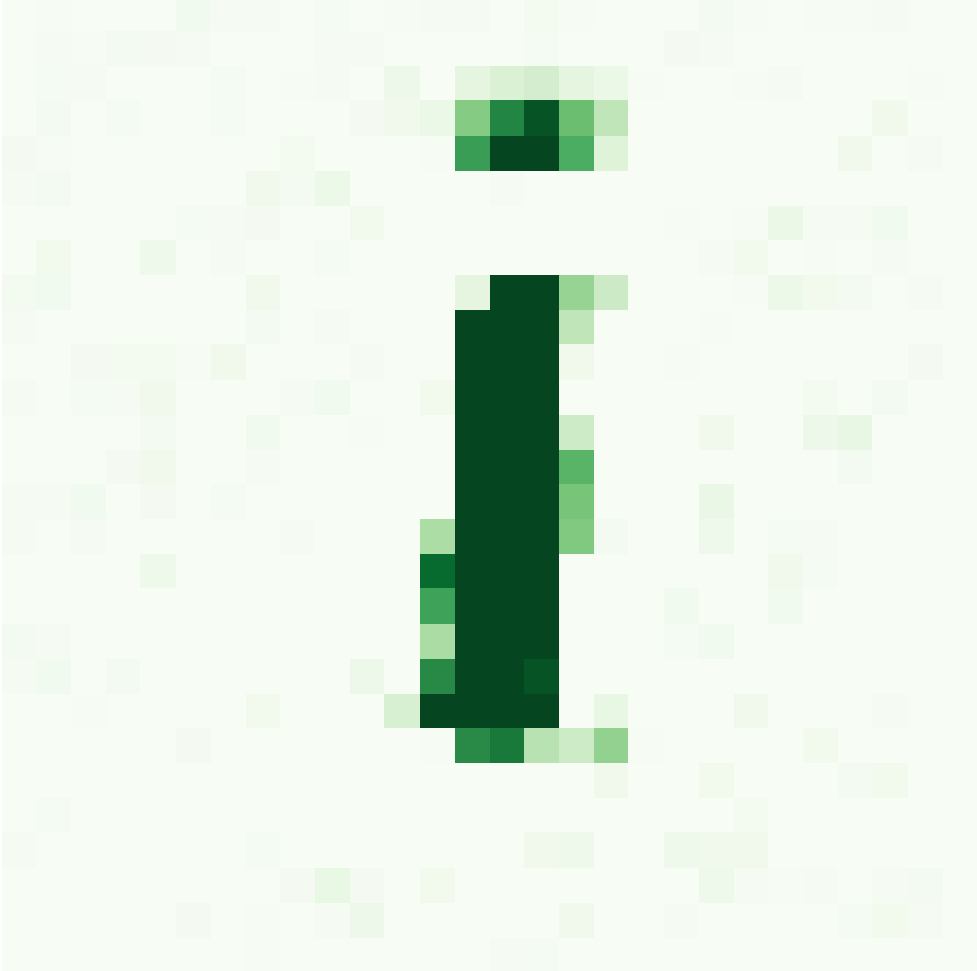} & 
\includegraphics[width=14.5mm]{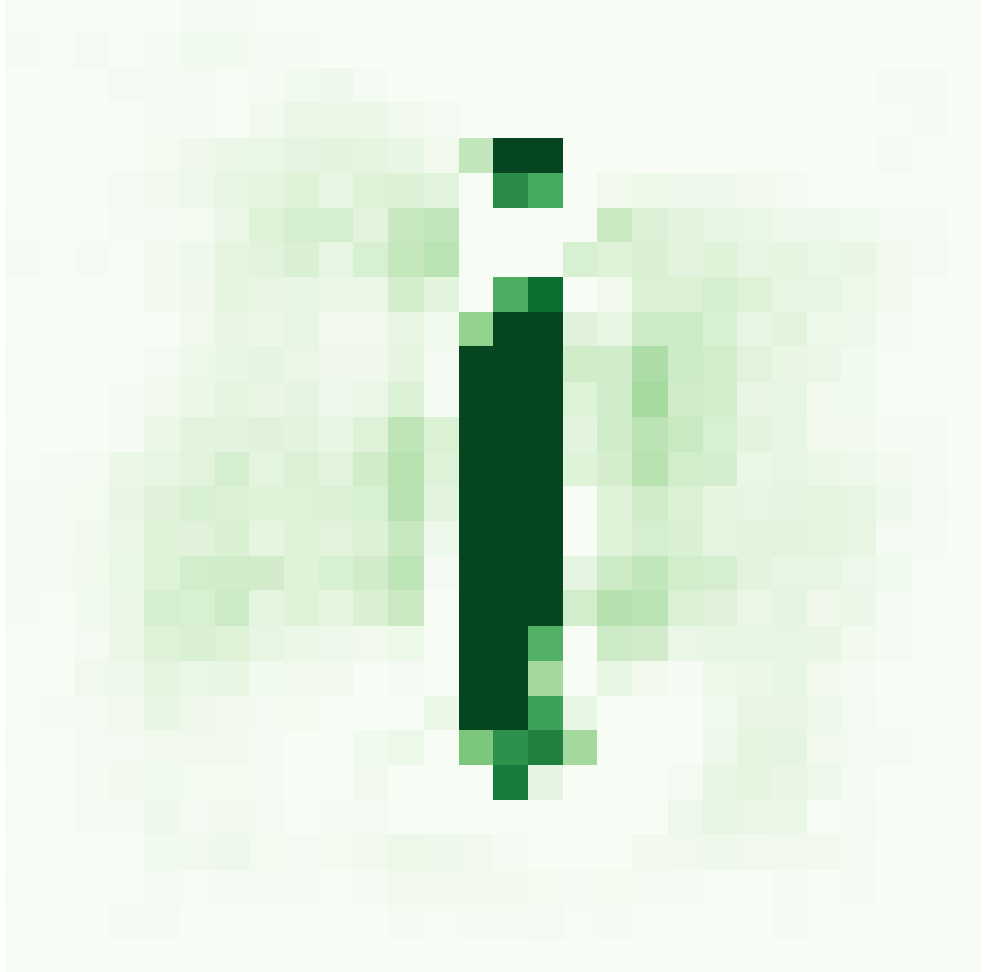} & 
\includegraphics[width=14.5mm]{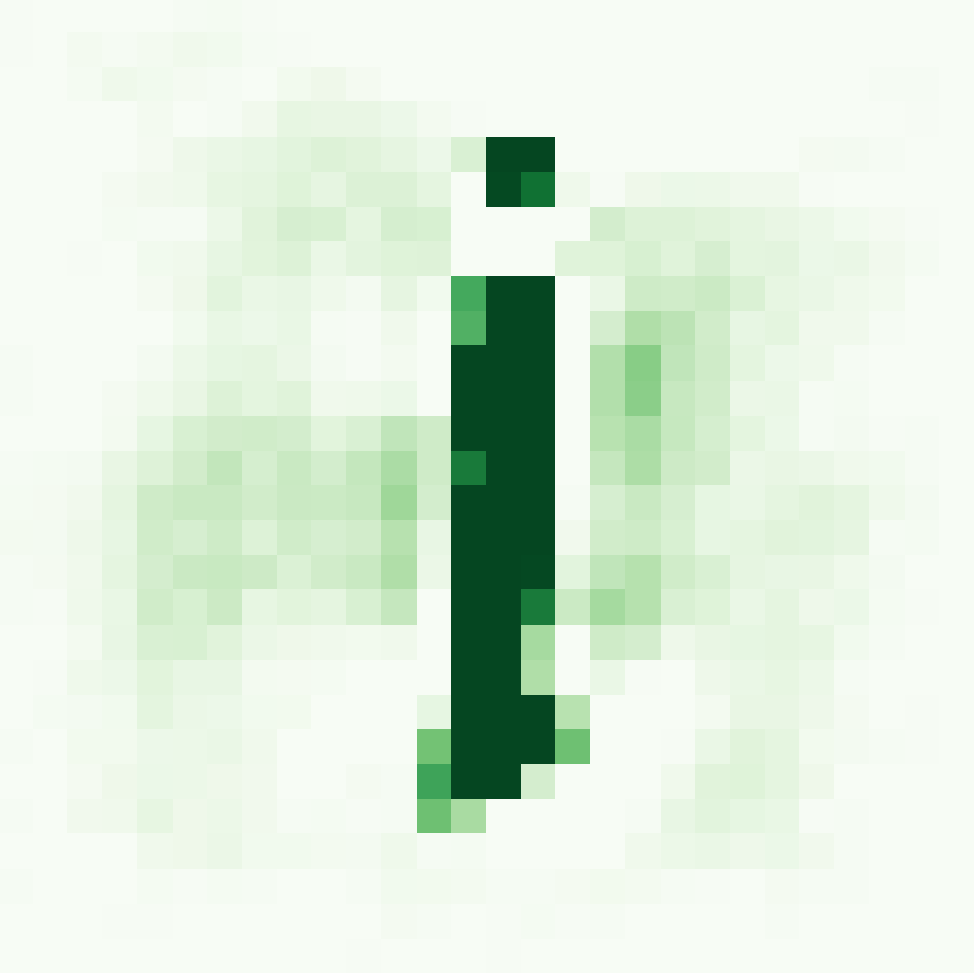} & 
\includegraphics[width=14.5mm]{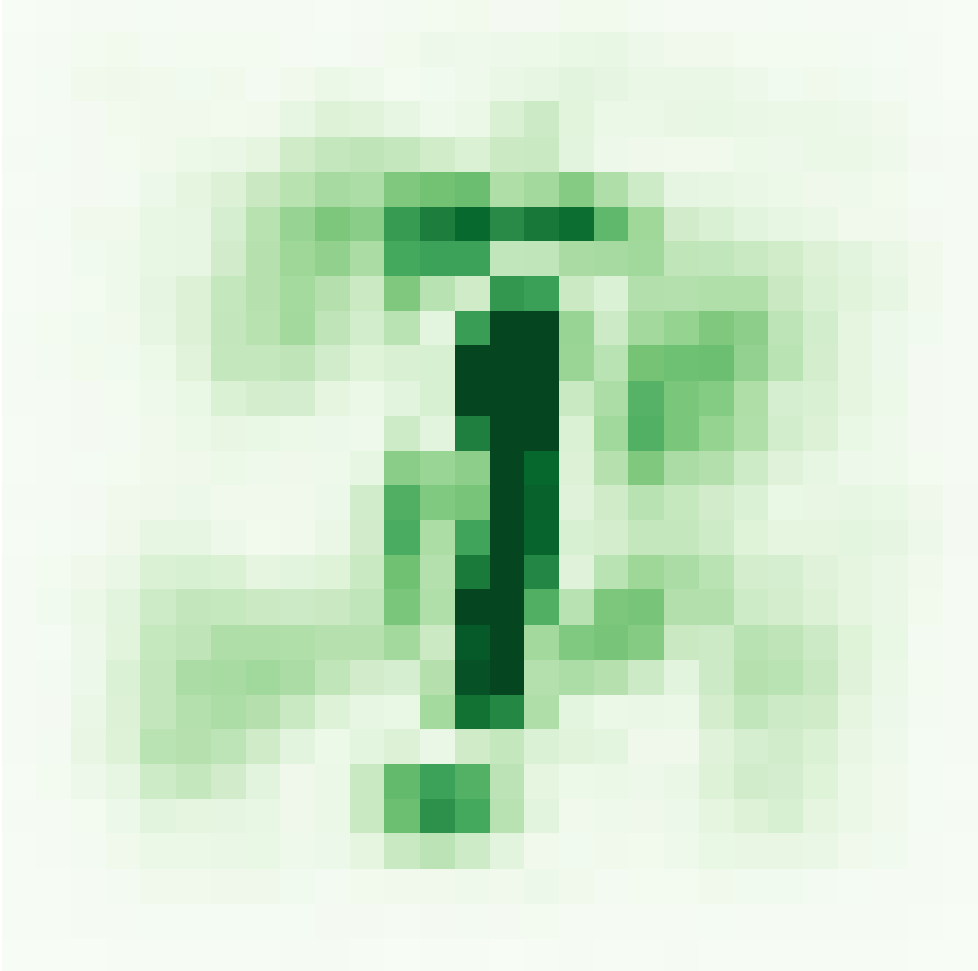} & 
\includegraphics[width=14.5mm]{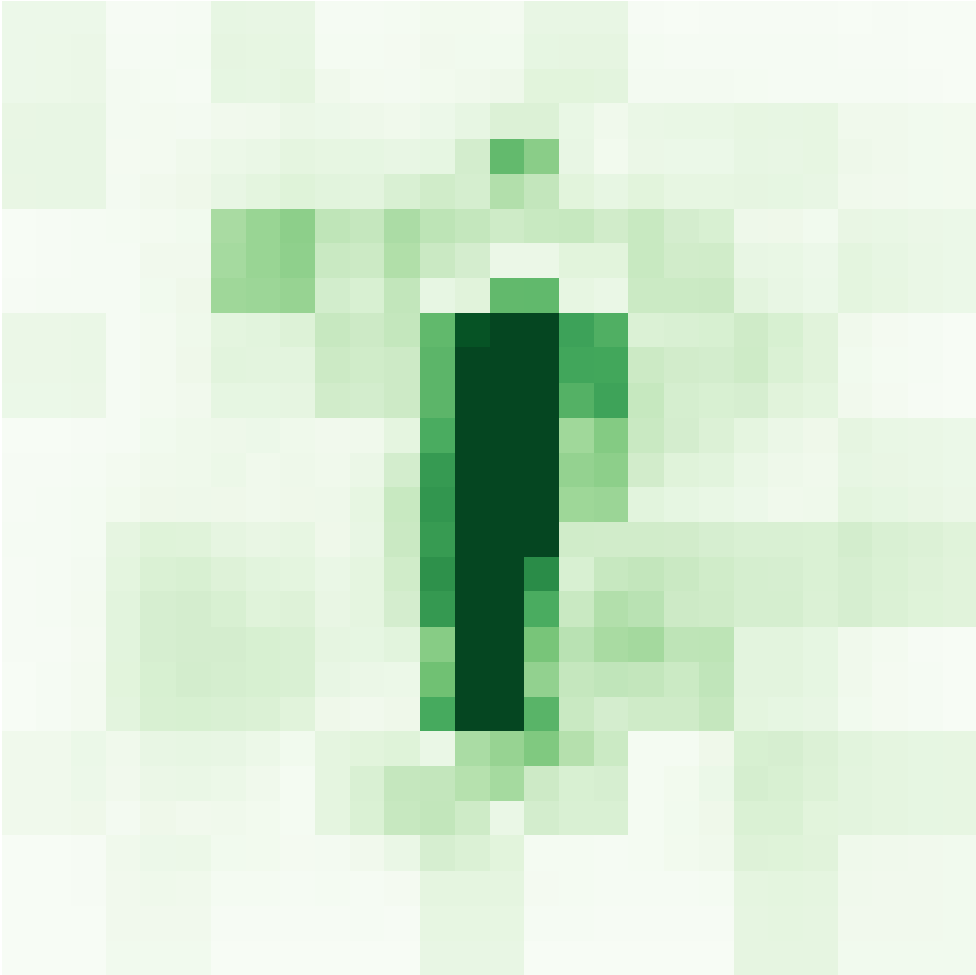} & 
\includegraphics[width=14.5mm]{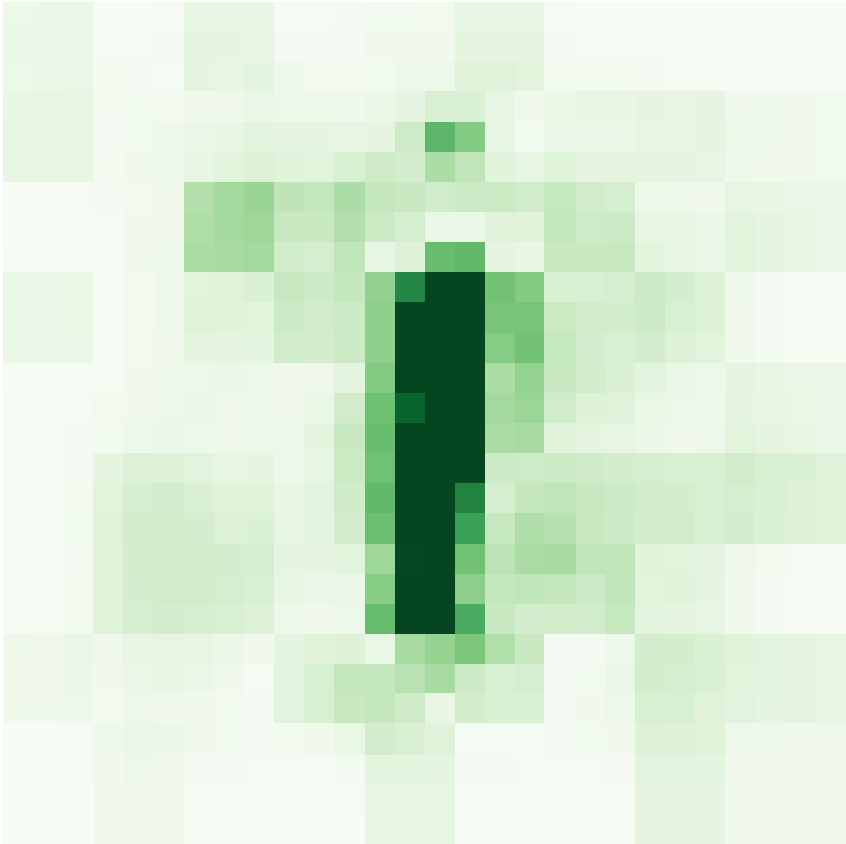} & 
\includegraphics[width=14.5mm]{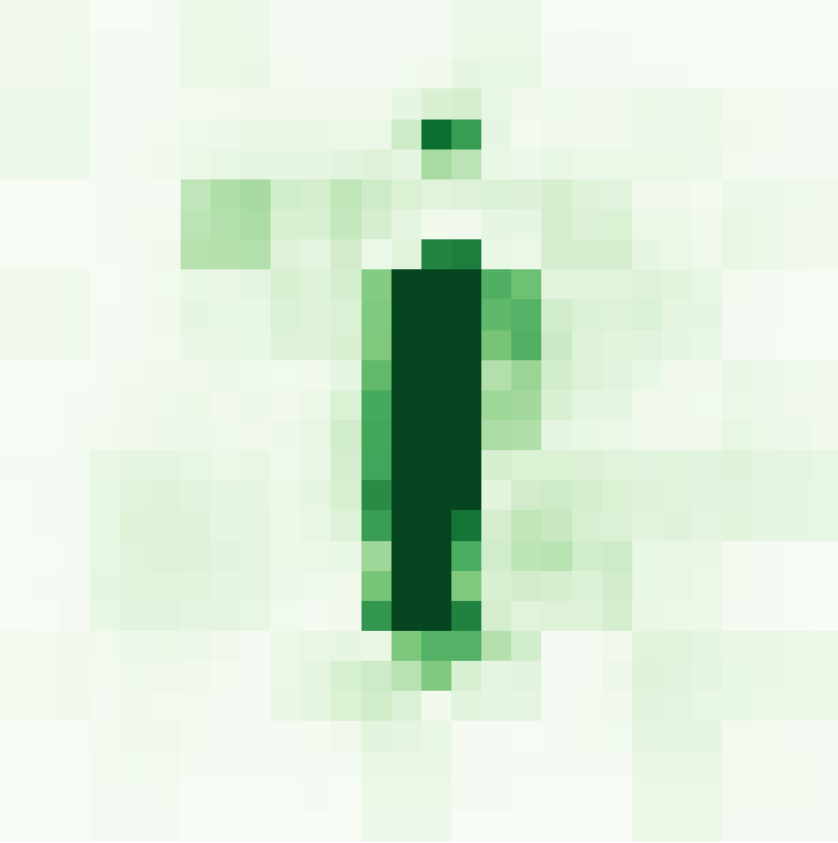} \\

\includegraphics[width=14.5mm]{visual_results/mnist_2_orig.png} & 
\includegraphics[width=14.5mm]{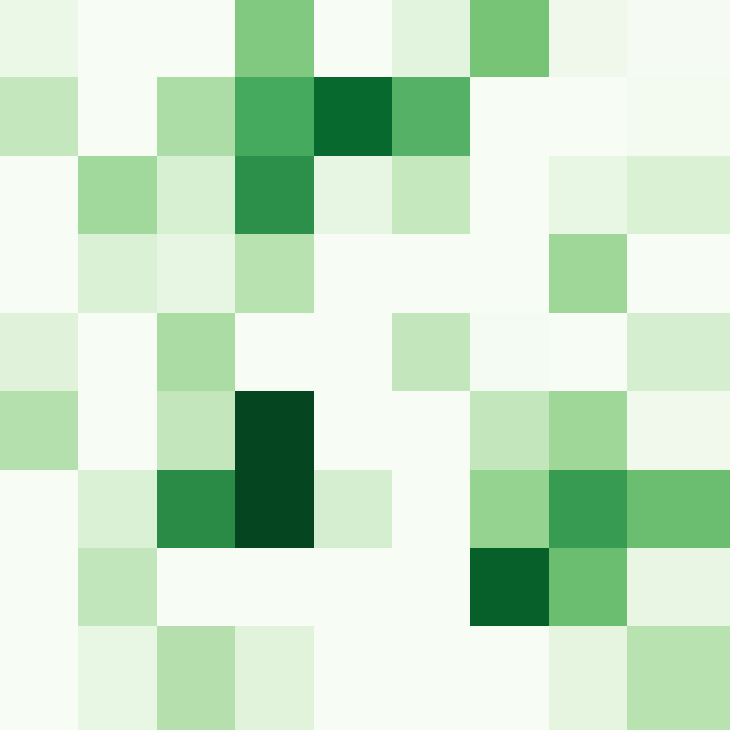} & 
\includegraphics[width=14.5mm]{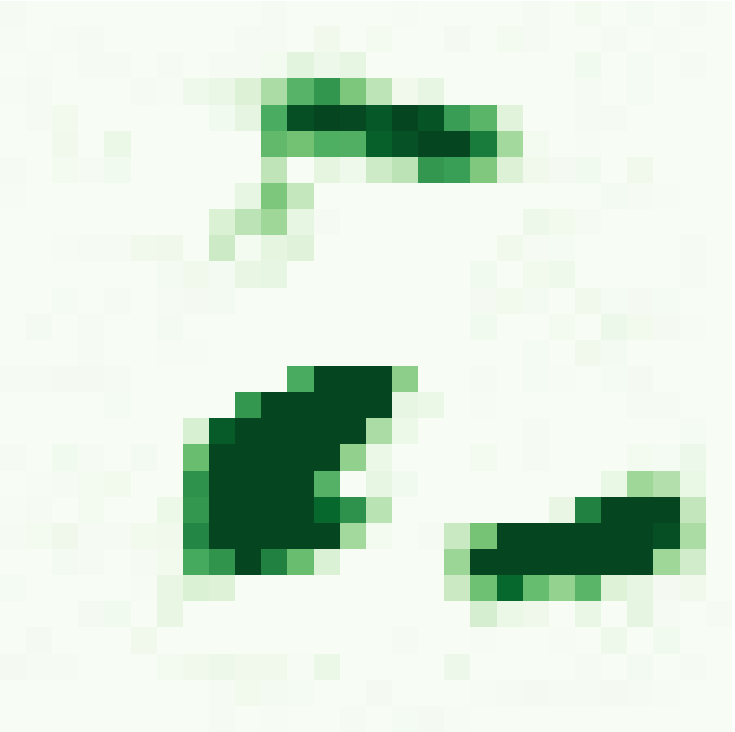} & 
\includegraphics[width=14.5mm]{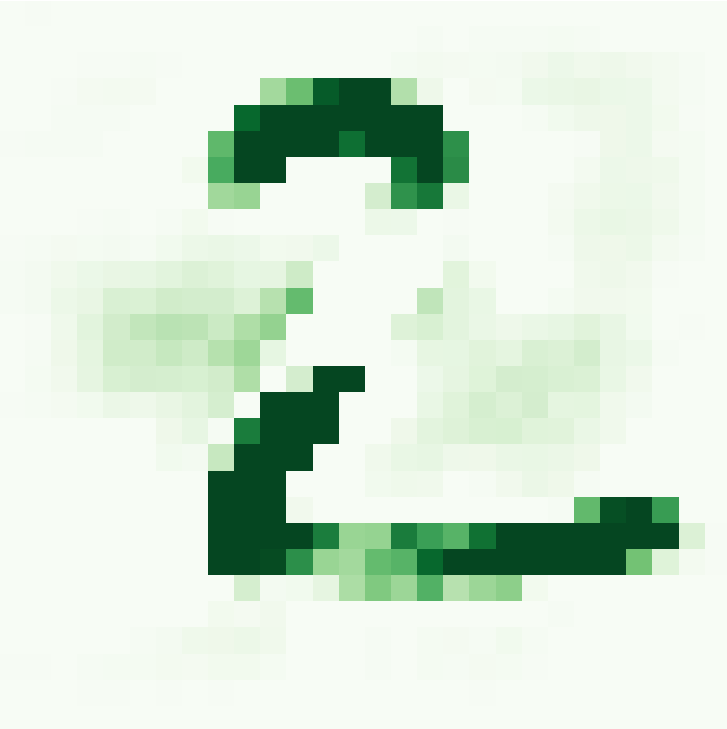} & 
\includegraphics[width=14.5mm]{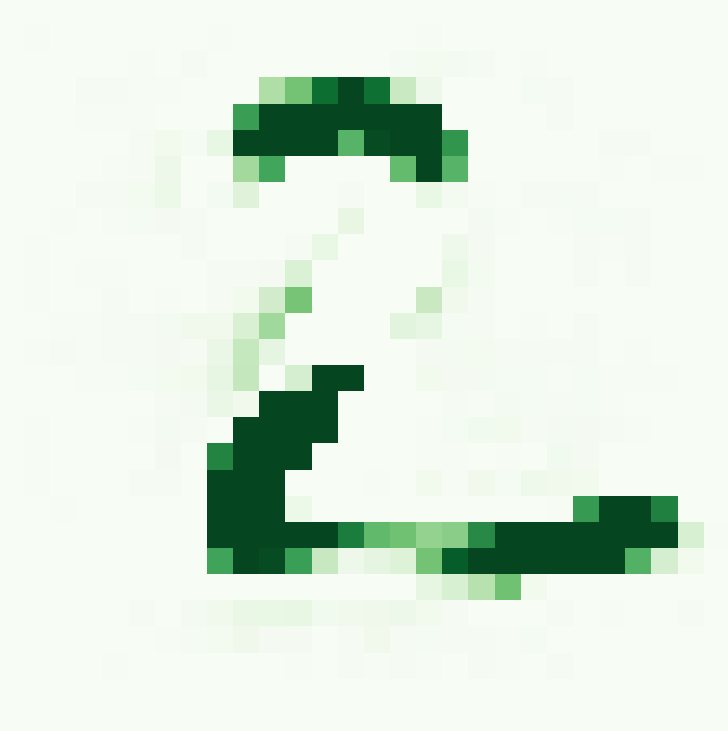} & 
\includegraphics[width=14.5mm]{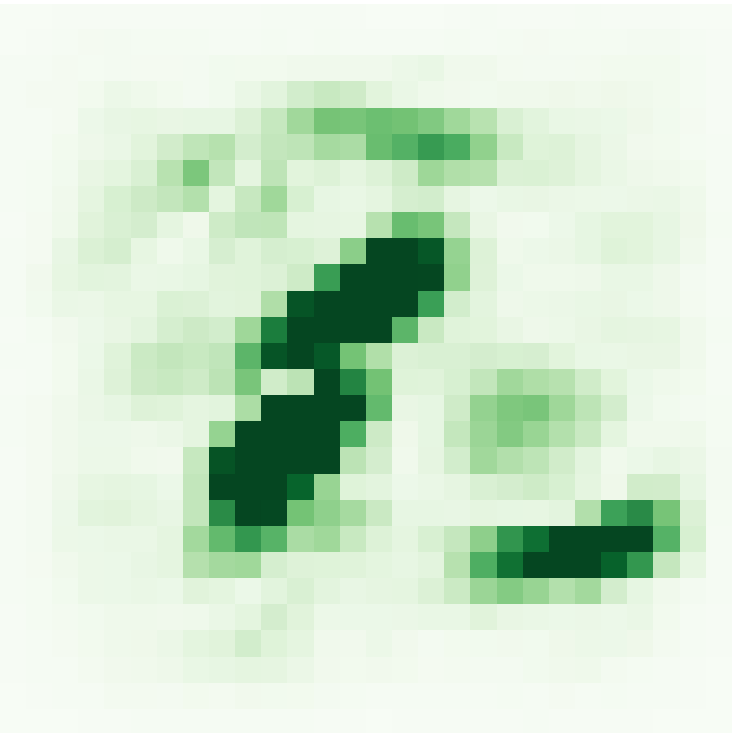} & 
\includegraphics[width=14.5mm]{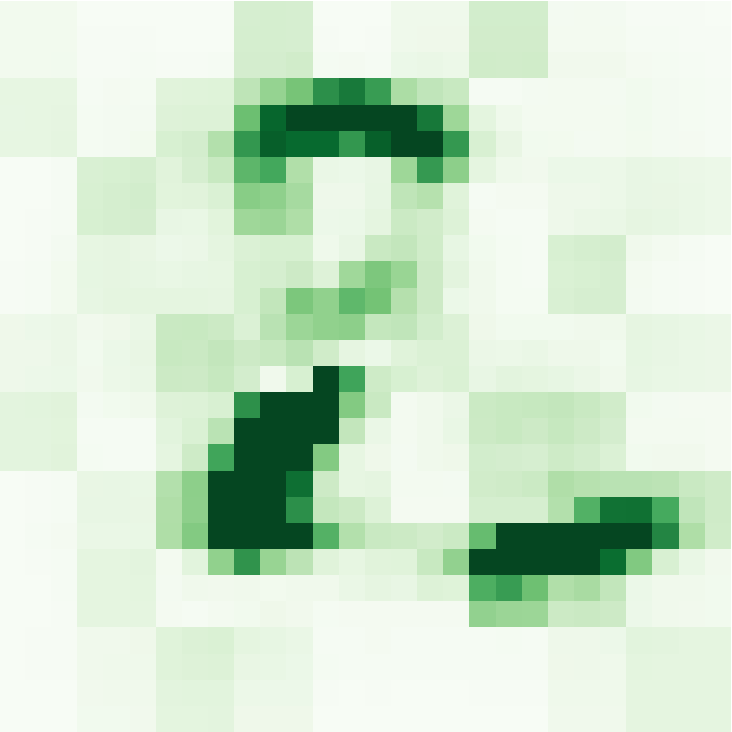} & 
\includegraphics[width=14.5mm]{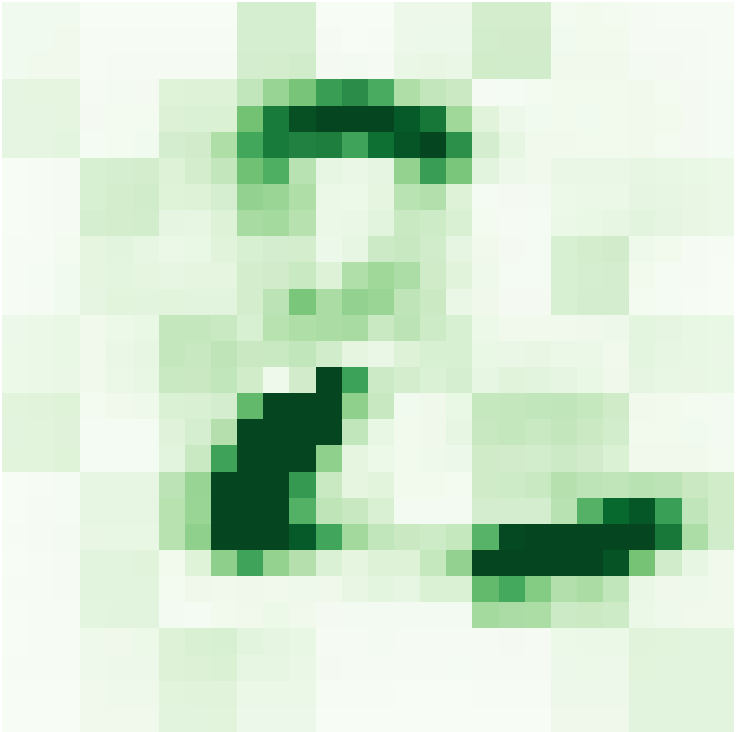} & 
\includegraphics[width=14.5mm]{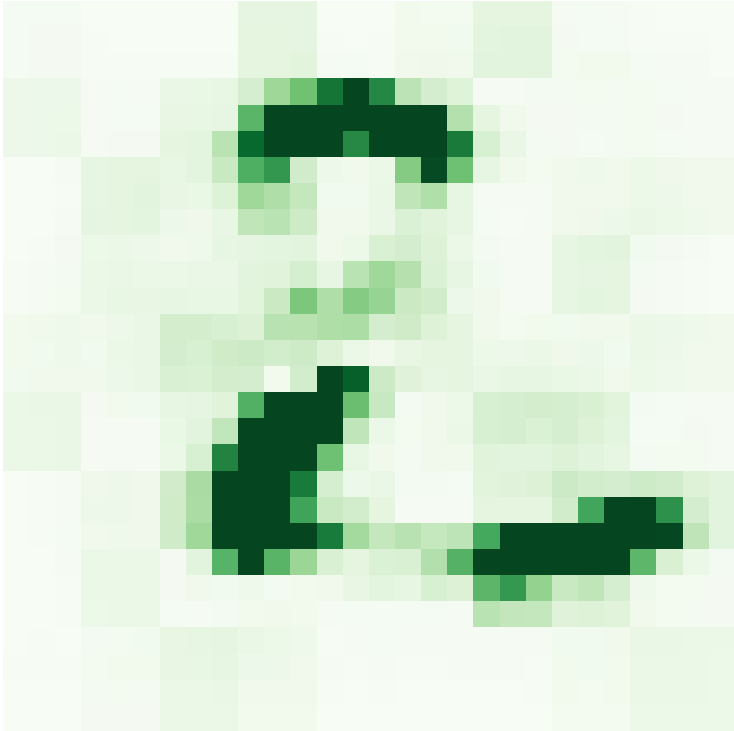} \\

\includegraphics[width=14.5mm]{visual_results/mnist_4_orig.png} & 
\includegraphics[width=14.5mm]{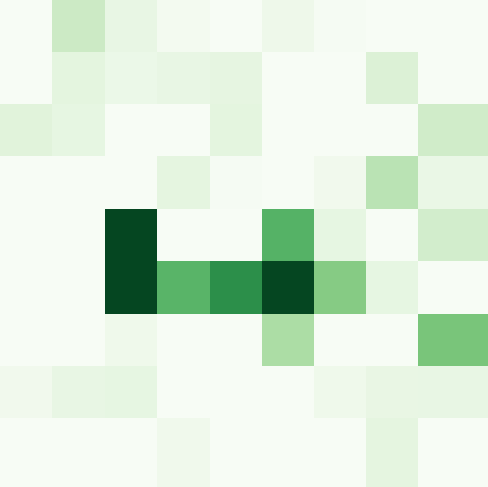} & 
\includegraphics[width=14.5mm]{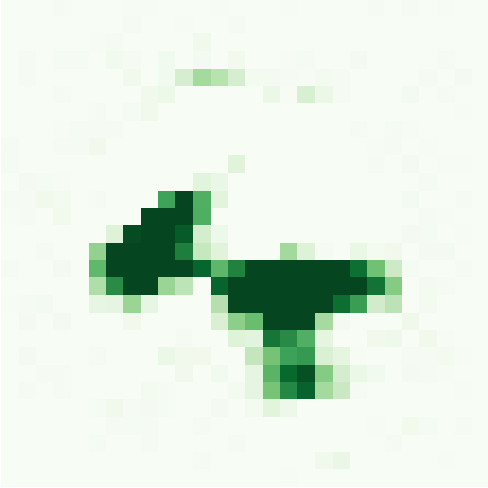} & 
\includegraphics[width=14.5mm]{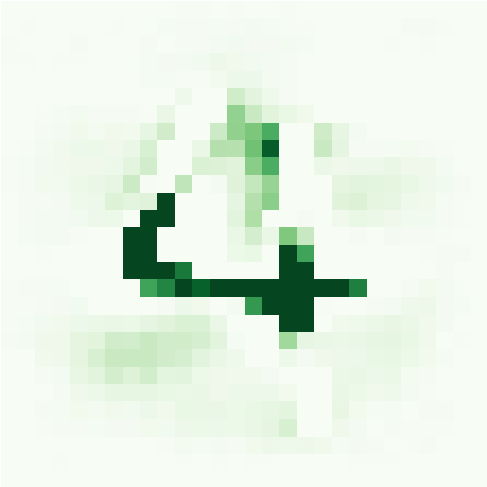} & 
\includegraphics[width=14.5mm]{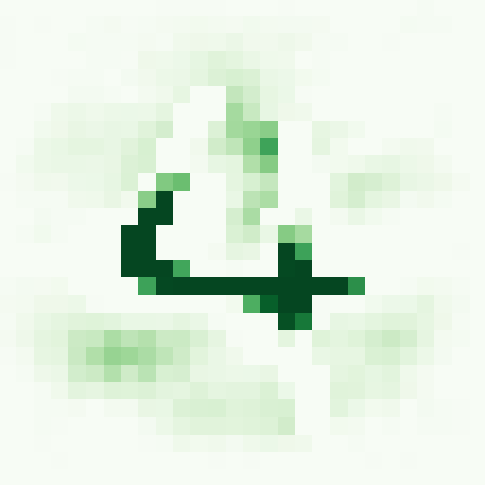} & 
\includegraphics[width=14.5mm]{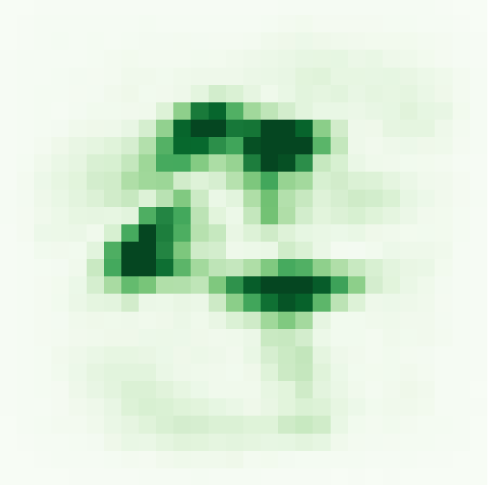} & 
\includegraphics[width=14.5mm]{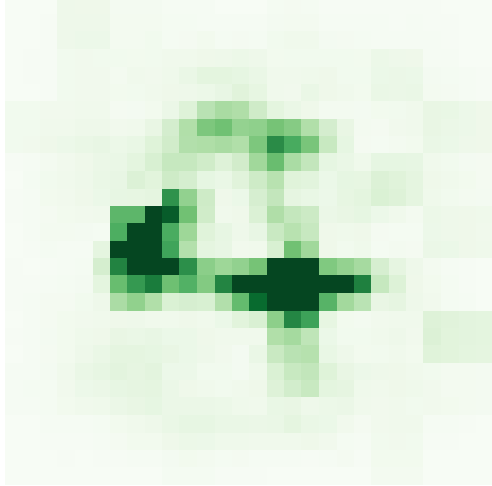} & 
\includegraphics[width=14.5mm]{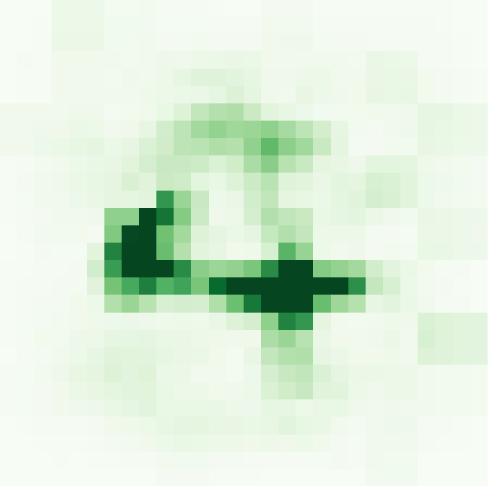} & 
\includegraphics[width=14.5mm]{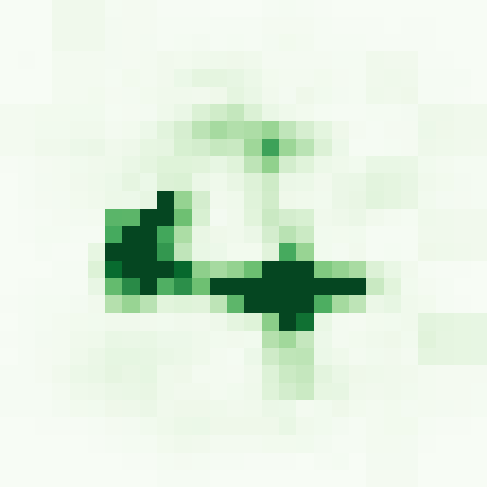} \\

\includegraphics[width=14.5mm]{visual_results/mnist_1_orig.png} & 
\includegraphics[width=14.5mm]{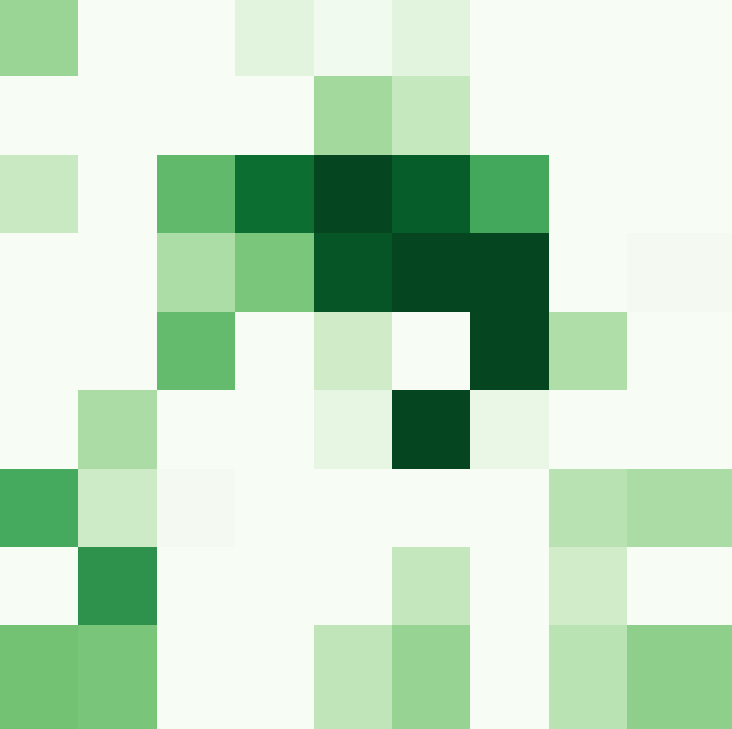} & 
\includegraphics[width=14.5mm]{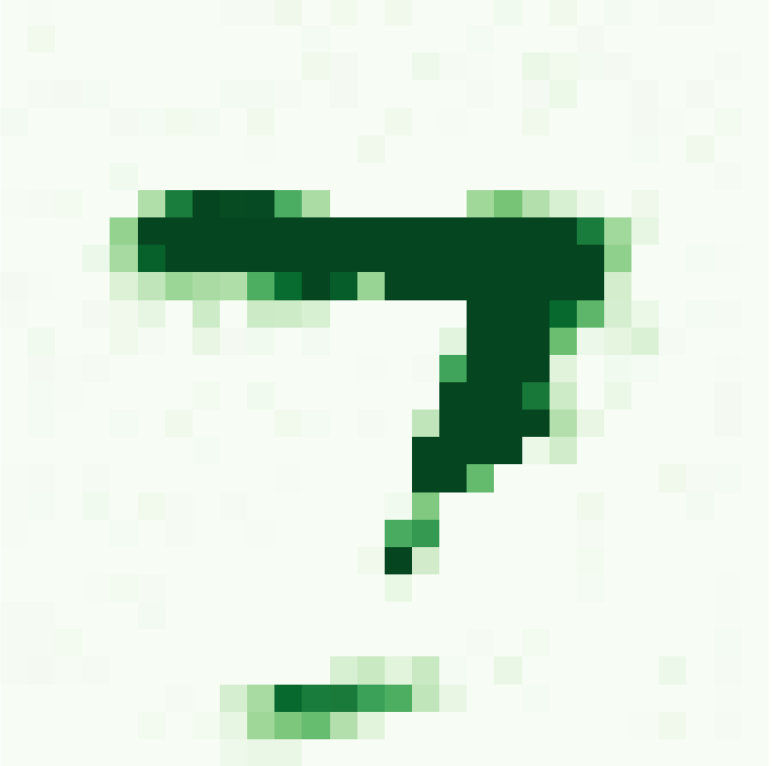} & 
\includegraphics[width=14.5mm]{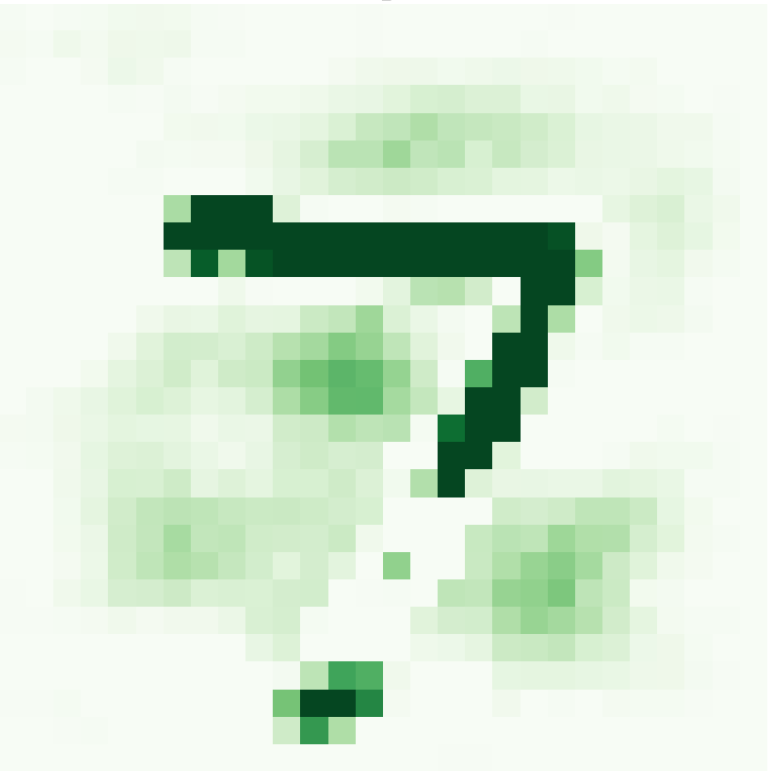} & 
\includegraphics[width=14.5mm]{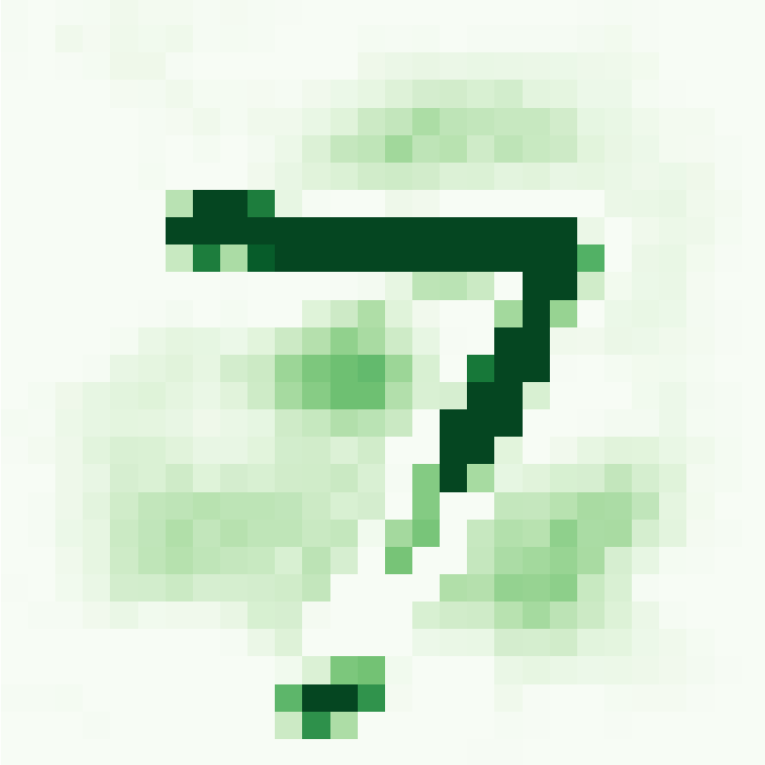} & 
\includegraphics[width=14.5mm]{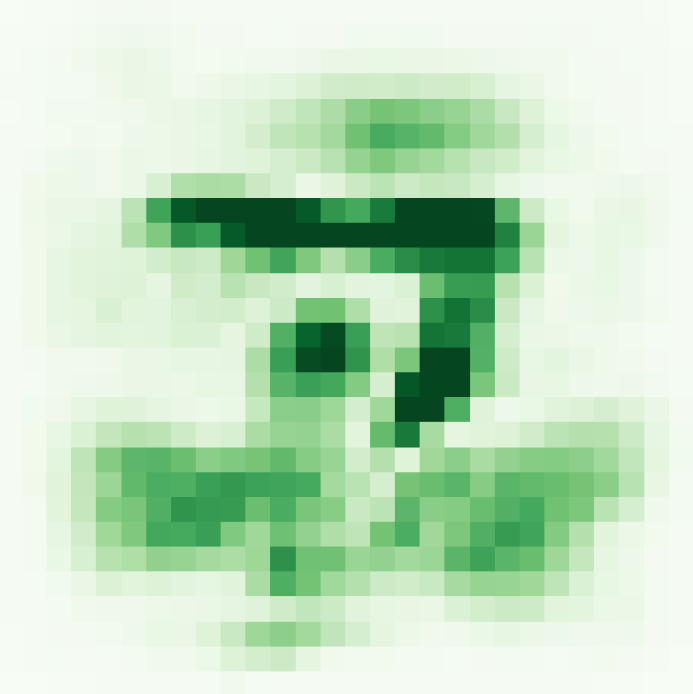} & 
\includegraphics[width=14.5mm]{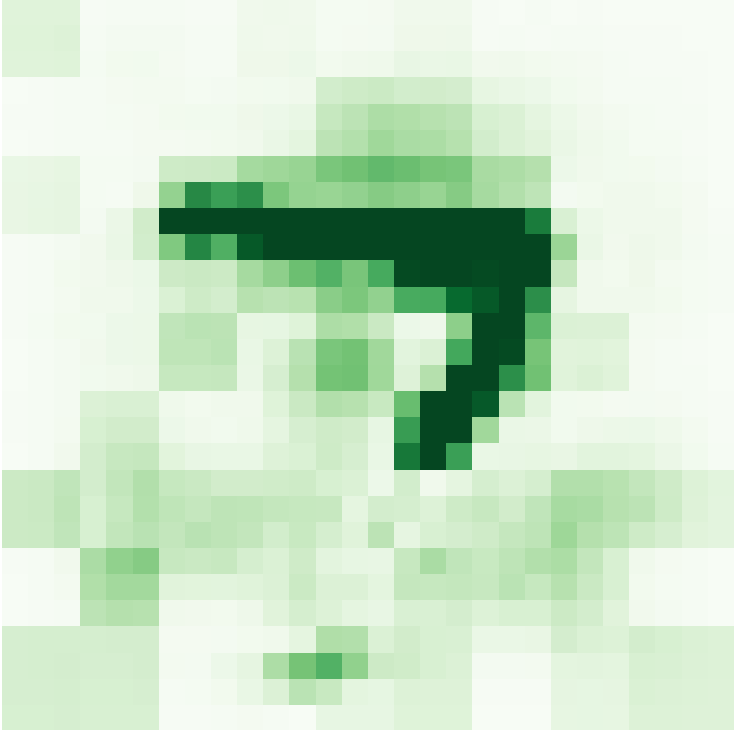} & 
\includegraphics[width=14.5mm]{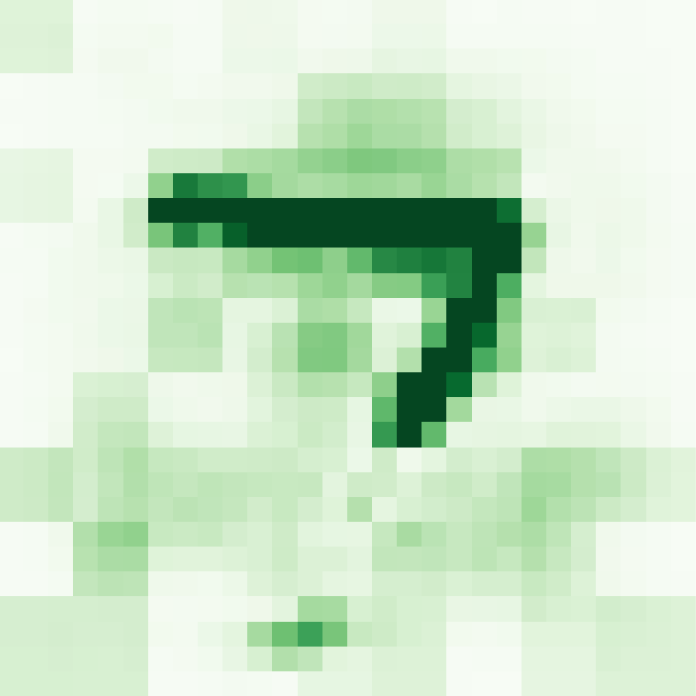} & 
\includegraphics[width=14.5mm]{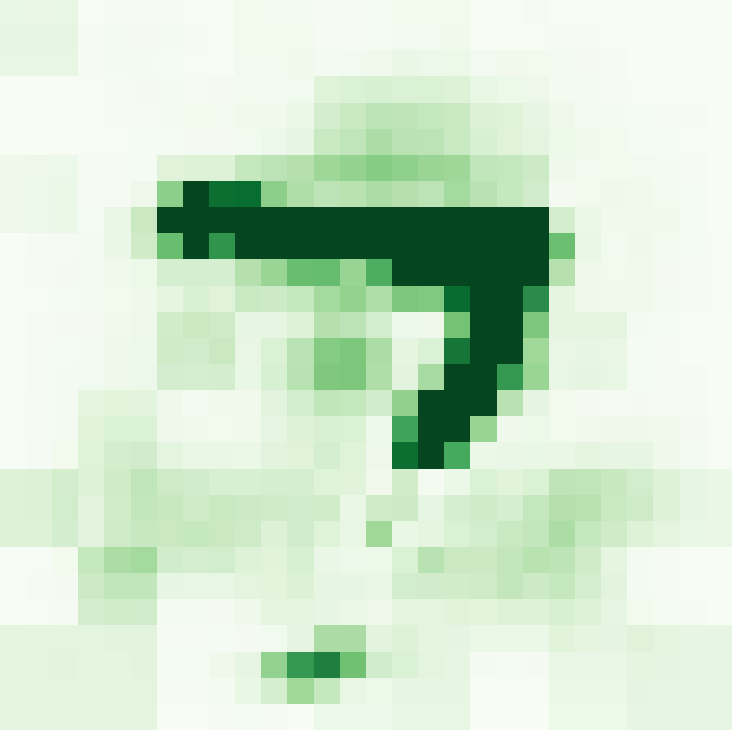} \\

\end{tabularx}
\caption{A visual comparison between base learners and ensemble methods on ImageNet \cite{deng2009imagenet} and MNIST \cite{lecun1998mnist} datasets.}
\label{tab:visual_experiment2}
\end{table*}

Table \ref{tab:visual_experiment2} presents extended experimental results for the compression with or without noisy feature attribution maps in the ensemble.

\end{document}